\documentclass[11pt]{article}
\usepackage[final]{acl}

\usepackage{times}
\usepackage{latexsym}
\usepackage[T1]{fontenc}
\usepackage[utf8]{inputenc}
\usepackage{microtype}
\usepackage{inconsolata}
\usepackage{graphicx}
\usepackage{tcolorbox}
\usepackage{amsmath}
\usepackage{amssymb}
\usepackage{booktabs}
\usepackage{multirow}
\usepackage{url}
\usepackage{enumitem}
\usepackage{placeins}
\usepackage{capt-of}
\usepackage{float}
\usepackage{perpage}
\MakeSorted{table}
\usepackage{tikz}
\usetikzlibrary{positioning,arrows.meta,fit,calc}
\usepackage{pgfplots}
\pgfplotsset{compat=1.17}

\title{HYPER: Hypothesis-Guided Planning for Evidence-Seeking Retrieval}

\author{
\textbf{Hangeol Chang}$^{1}$ \quad 
\textbf{Changsun Lee}$^{1}$ \quad
\textbf{Seungjoon Rho}$^{2}$ \quad
\textbf{Junho Yeo}$^{3}$ \quad
\textbf{Jong Chul Ye}$^{1}$ \\
$^{1}$Graduate School of AI, KAIST, Republic of Korea \\
$^{2}$School of Electrical Engineering, KAIST, Republic of Korea \\
$^{3}$Department of Industrial Engineering, Yonsei University, Republic of Korea \\
\texttt{\{hangeol,sunny17,ryzere\}@kaist.ac.kr} \\
\texttt{asap03153@yonsei.ac.kr \quad jong.ye@kaist.ac.kr}
}
\date{}
\hypersetup{pdftitle={HYPER: Hypothesis-Guided Planning for Evidence-Seeking Retrieval},pdfauthor={Hangeol Chang, Changsun Lee, Seungjoon Rho, Junho Yeo, Jong Chul Ye}}

\begin{document}
\maketitle

\begin{abstract}

Retrieval-Augmented Generation (RAG) improves Large Language Models (LLMs) by grounding generation in external, non-parametric knowledge. However, when a task requires choosing among competing possibilities, such as a diagnosis among plausible alternatives, topically relevant context is often insufficient to drive the final decision. To address this issue, we propose \textbf{HYPER}: \textbf{Hy}pothesis-Guided \textbf{P}lanning for \textbf{E}vidence-Seeking \textbf{R}etrieval, a training-free pre-retrieval planning framework that reorients RAG from topic-oriented retrieval to evidence-oriented retrieval. HYPER uses a provisional answer hypothesis as a planning anchor. It then decomposes retrieval into three evidence roles. \textsc{Support} seeks evidence for the provisional answer. \textsc{Distinction} retrieves criteria that separate competing alternatives. \textsc{Feature} grounds salient question features needed for the decision. This planning step steers retrieval toward evidence near the decision boundary, enabling the generator to answer from context that is more directly aligned with the required choice. On MedQA and MMLU-Med, where single-query retrieval falls below no-retrieval chain-of-thought prompting, HYPER achieves the best average accuracy among ten baselines, improving over Simple RAG by 5.9 and 3.6 points, respectively. Its gains transfer to the 14 subjects of MMLU-Pro, at the cost of two additional LLM calls per question.
Code is available at \url{https://github.com/hangeol/HypothesisRAG}.
\end{abstract}

\begin{figure*}[!t]
\centering
\includegraphics[width=\textwidth]{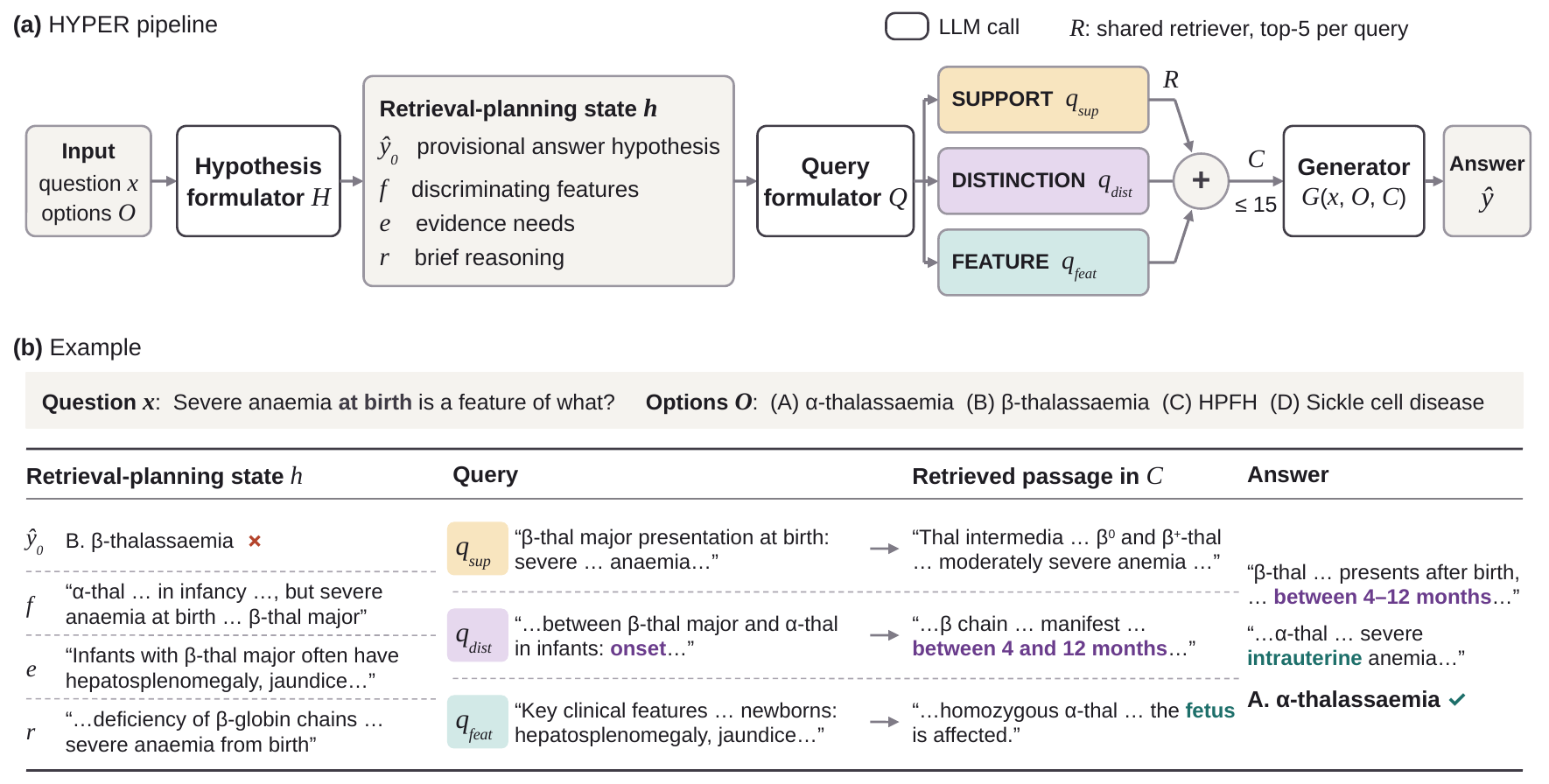}
\caption{\textbf{Overview of HYPER.} (a) $H$ forms a retrieval-planning state $h=(\hat{y}_0, f, e, r)$ from $x$ and $O$. $Q$ writes \textsc{Support}, \textsc{Distinction}, and \textsc{Feature} queries from $x$ and $h$. Each query retrieves its top-5 passages with a shared retriever $R$, and the results are fused into a context $C$ of at most 15 passages. $G$ answers from $(x, O, C)$ without seeing $h$. White boxes are LLM calls. (b) A real MMLU-Med example (Qwen3-4B), with quotes abridged with ``\ldots'' (thal = thalassaemia; full text in Appendix~\ref{app:example_corrected}). Although $\hat{y}_0$ is wrong, \textsc{Distinction} and \textsc{Feature} retrieve the onset evidence that leads $G$ to the correct answer.}
\label{fig:pipeline}
\end{figure*}

\section{Introduction}

Retrieval-Augmented Generation (RAG) grounds language-model generation
in external knowledge and is now a standard component of many question
answering systems \citep{lewis2020retrieval}. However, the benefit of retrieval varies. Retrieved documents can contain irrelevant or noisy
information that harms downstream generation
\citep{amiraz2025distracting,yoran2023making,du2025context}. This issue is especially visible in medical multiple-choice QA, where strong models can often answer from parametric knowledge alone. In MIRAGE's MedCorp/RRF-4 setting, retrieval lowered MedQA accuracy below no-retrieval chain-of-thought (CoT) prompting for five of six LLMs, including GPT-4, although it helped every model on literature-based questions \citep{xiong2024benchmarking}. In our experiments, simple single-query retrieval falls below CoT in all eight model--dataset settings.

The core difficulty is deciding what to retrieve.
In decision-oriented tasks, topically related passages may still be uninformative for choosing among competing answers.  For example, the question in Figure~\ref{fig:pipeline}b asks which disease causes severe anaemia at birth. A passage saying that a disease causes severe anaemia does not resolve this question unless it also establishes that the symptoms can be present at birth. In our analysis, 70\% of the contexts that single-query retrieval returns for MedQA are labeled \textsc{Not Useful} for reaching the correct answer (Section~\ref{sec:context-utility}). Under a bounded
context budget, such passages displace evidence that could narrow the
decision. We therefore focus on retrieval planning for decision-useful
evidence: evidence that supports one answer, distinguishes it from
plausible alternatives, or grounds salient clues in the question. This
framing builds on relevance to the user's information need
\citep{cooper1971definition} and recent RAG work that evaluates retrieved context
by downstream usefulness
\citep{salemi2024evaluating,joren2025sufficient,zhang2024large,zhang2024iterative}.

This motivates a pre-retrieval planning problem: the system should decide
what evidence would help resolve the answer choice before the candidate
pool is formed, rather than relying only on a generic first pass followed
by post-hoc selection.

To address this planning problem, we introduce \textbf{HYPER}:
\textbf{Hy}pothesis-Guided \textbf{P}lanning for
\textbf{E}vidence-Seeking \textbf{R}etrieval, a training-free
pre-retrieval planning method. HYPER uses a provisional answer hypothesis as a planning anchor for retrieval. From
the question and candidate answers, it constructs a compact
retrieval-planning state and converts it into three targeted queries:
\textsc{Support} for evidence supporting the provisional answer,
\textsc{Distinction} for criteria separating competing alternatives, and
\textsc{Feature} for evidence grounding salient question features. The provisional answer hypothesis steers retrieval and is withheld from the final generator by default.

We evaluate HYPER on MedQA and MMLU-Med, where questions often turn on the criterion that separates a few closely related diagnoses or treatments. Under the same final-context budget,
HYPER achieves the best average performance across both datasets and
outperforms strong retrieval baselines in nearly all dataset--model
settings. It also increases the fraction of retrieved context sets judged useful for the downstream decision, and its gains transfer to the non-medical subjects of MMLU-Pro.

Our contributions are:
\begin{enumerate}[leftmargin=*, itemsep=0pt, topsep=1pt]
    \item We introduce \textbf{HYPER}, a training-free pre-retrieval planning strategy that uses a provisional answer hypothesis to construct
    three complementary evidence-seeking queries.
    \item We analyze retrieved context with the labels
    \textsc{Entailed}, \textsc{Useful-but-Not-Entailed}, and
    \textsc{Not Useful}, measuring whether retrieval provides evidence useful for the downstream decision, and validate the labels with human annotators, a second judge, and a gold-blind protocol.
    \item We show that hypothesis-guided retrieval planning improves
    medical multiple-choice QA under a shared final-context budget, with
    gains in both answer accuracy and context utility. We further analyze accuracy when the provisional answer hypothesis is wrong and show transfer to MMLU-Pro.
\end{enumerate}

\section{Related Work}

\paragraph{Harmful context and retrieval utility.}
RAG systems are commonly organized into pre-retrieval, retrieval,
post-retrieval, and generation stages
\citep{fan2024survey,huang2024survey}. HYPER intervenes at the
pre-retrieval stage. This focus is motivated by evidence that additional context can hurt: irrelevant or weakly aligned passages can distract the model and reduce robustness, and longer contexts can impair reasoning even when relevant evidence is present
\citep{amiraz2025distracting,yoran2023making,du2025context}. Information retrieval defines relevance in relation to the user's information need \citep{cooper1971definition}. Recent RAG work
examines retrieval utility by evaluating retrieved context in terms of
sufficiency or downstream usefulness rather than surface-level relevance
alone
\citep{salemi2024evaluating,joren2025sufficient,zhang2024large,zhang2024iterative}.
HYPER follows this utility-oriented view, focusing on whether retrieved
context helps resolve the downstream decision rather than merely matching
the question topic.

\paragraph{Query transformation and multi-query retrieval.}
A direct way to improve retrieval is to transform the input query before
search. Training-based methods learn query rewriters or retrieval-aware
generators using supervised data, downstream reader feedback, or
process-level rewards
\citep{ma2023query,chan2024rq,ye2025q,wang2025maferw}. Training-free methods instead use prompting to rewrite, expand, decompose, or diversify queries \citep{gao2023precise,rackauckas2024rag,li2024dmqr}.  HYPER also generates multiple queries, each planned around one of three evidence roles:
supporting a provisional answer, distinguishing alternatives, and
verifying salient question features.

\paragraph{Reasoning-guided and agentic retrieval.}
Reasoning-guided RAG methods treat retrieval as part of an adaptive
reasoning process. IRCoT interleaves chain-of-thought reasoning with
retrieval for multi-hop questions \citep{trivedi2023interleaving};
Self-RAG trains a model to decide when to retrieve and to critique
retrieved evidence through self-reflection \citep{asai2024self}; and
broader agentic RAG systems use planning or tool-use policies to decide
when and how to retrieve \citep{liang2025reasoning}. These methods
mainly study adaptive retrieval control across one or more reasoning
steps. HYPER studies a complementary pre-retrieval planning problem:
before the first retrieval pass, how can a provisional reasoning state be
converted into targeted evidence-seeking queries while keeping the retriever and final generator fixed?

\paragraph{Post-retrieval selection.}
Post-retrieval methods start from a retrieved candidate pool and decide which passages should enter the final generator context. MAIN-RAG filters documents using LLM evaluations of question-document-answer triples \citep{chang2025main}, while ITEM-A iteratively alternates pseudo-answer generation with utility judgments to select useful passages \citep{zhang2024iterative}. Other approaches use MMR to balance relevance and novelty, reciprocal rank fusion to aggregate ranked lists \citep{carbonell1998use,cormack2009reciprocal}, or RankRAG to rerank context \citep{yu2024rankrag}. REPLUG uses a frozen language model to supervise retriever training and ensembles predictions across documents \citep{shi2024replug}; CRAG evaluates retrieval quality and can invoke additional web search \citep{yan2024corrective}. HYPER intervenes earlier: it constructs evidence-seeking queries before the candidate pool is formed.

\paragraph{Medical RAG.}
Prior work shows that medical RAG performance is sensitive to corpus choice,
retriever design, and retrieval strategy, and that reformulation or
iterative follow-up retrieval can improve performance
\citep{xiong2024benchmarking,xiong2024improving,kim2025rethinking}. We
use medical multiple-choice QA because it makes the gap between topical
relevance and decision usefulness especially visible, and the HYPER formulation applies beyond medicine (Section~\ref{sec:transfer}).

\section{Method}
\label{sec:method}

\subsection{Problem Setting}
We consider multiple-choice decision settings where, given a question $x$, a model selects an answer from candidates $O=\{o_1,\ldots,o_K\}$ and may use a context set $C_m(x,O)\subseteq\mathcal{D}$ that method $m$ retrieves from a corpus $\mathcal{D}$. We compare methods under a shared
final-context budget,
\[
|C_m(x,O)| \le B_{\max},
\]
which limits the evidence shown to the final generator. The budget is an upper bound: methods may inspect different intermediate
candidate pools, and the realized final context may contain fewer than
$B_{\max}$ documents because of deduplication or filtering.

In this setting, retrieval is useful only when the final context helps
resolve the answer choice.  Our scope is tasks with given candidates, and open-ended settings in which the candidates must first be generated are left for future work.

\subsection{HYPER}

HYPER (Figure~\ref{fig:pipeline}) first forms a provisional answer hypothesis, then infers what evidence should hold if that hypothesis were correct, and finally retrieves documents that test those expectations.

More specifically, the hypothesis formulator \(H\) first constructs the retrieval-planning state
\[
h = H(x, O) = (\hat{y}_0, f, e, r),
\]
where \(\hat{y}_0\) is the provisional answer hypothesis, \(f\) denotes discriminating features, \(e\) denotes evidence needs, and \(r\) denotes brief reasoning. Conditioned on the original question \(x\) and this state \(h\), the query formulator \(Q\) produces three queries,
\[
(q_{\mathrm{sup}}, q_{\mathrm{dist}}, q_{\mathrm{feat}}) = Q(x, h).
\]
The shared retriever \(R\) retrieves evidence for these queries, and the retrieved results are fused into a final context \(C\),
\[
C = \mathrm{Fuse}\big(R(q_{\mathrm{sup}}), R(q_{\mathrm{dist}}), R(q_{\mathrm{feat}})\big).
\]
Based on this context, the downstream generator \(G\) predicts the final answer as
\begin{align}\label{eq:answer}
\hat{y} = G(x, O, C).
\end{align}
Crucially, the hypothesis influences retrieval only through the rewritten queries and is withheld from the generator in the default setup. 

\paragraph{Stage 1: Hypothesis formulator.}

 This stage defines a concrete verification target for retrieval. The hypothesis formulator outputs a compact structured record containing the current best-guess option $\hat{y}_0$, question-specific clues $f$ that separate the leading candidates, facts $e$ that it expects would support the provisional answer and should be sought through retrieval, and a concise explanation $r$ for why this hypothesis is currently preferred.
We keep this state short and structured because HYPER uses it to steer retrieval rather than to expose a long reasoning trace to the generator. What matters is the verification agenda implied by the provisional answer hypothesis.

\paragraph{Stage 2: Query formulator.}

Conditioned on the original question and the structured retrieval-planning state $(\hat{y}_0, f, e, r)$, the query formulator generates exactly three high-precision search queries, which we name \textsc{Support}, \textsc{Distinction}, and \textsc{Feature}. The query formulator sees the candidate answers only through $\hat{y}_0$, $f$, and $r$, which name the leading options.
\begin{itemize}[leftmargin=*, itemsep=0pt, topsep=1pt]
    \item \textbf{\textsc{Support} $q_{\mathrm{sup}}$: supports the provisional answer hypothesis.} This query uses the hypothesized answer text, evidence needs and brief reasoning to retrieve passages that would directly support the current hypothesis.
    \item \textbf{\textsc{Distinction} $q_{\mathrm{dist}}$: distinguishes leading alternatives.} This query uses the discriminating features to retrieve criteria that separate the strongest competing options.
    \item \textbf{\textsc{Feature} $q_{\mathrm{feat}}$:} grounds salient question
    features. This query targets key clues in the question and retrieves
    evidence about their clinical meaning or decision relevance, independent
    of whether the current hypothesis is correct.
\end{itemize}
Together, these queries instantiate complementary retrieval roles.  This role separation is intended to make retrieval useful even when the provisional answer hypothesis is incomplete or partially wrong (Section~\ref{sec:hypothesis}).

\paragraph{Retrieval, fusion, and generation.}

The three rewritten queries are passed to a shared retriever, which retrieves the top-5 documents for each. We fuse these lists by deduplicating the documents and summing their retrieval scores across the queries, meaning documents retrieved multiple times receive a higher aggregate score. The final context $C$ is then formed by ranking the documents based on this score.

\section{Experimental Setup}
\label{sec:exp_setup}

\subsection{Benchmarks}
We focus on medical multiple-choice QA because it provides a clean setting
for studying decision-oriented retrieval: many questions are answerable
from parametric knowledge, but retrieval can help when it surfaces
diagnostic criteria, differential comparisons, or question-specific
clinical clues. Within this domain, we evaluate HYPER on two standard
benchmarks, \textbf{MedQA} \citep{jin2021medqa} and \textbf{MMLU-Med}
\citep{hendryckstest2021,xiong2024benchmarking}, which provide controlled multiple-choice
supervision for retrieval-augmented decision making. To test transfer beyond medicine, we also evaluate all 12,032 test questions of MMLU-Pro, which span 14 subjects with up to ten options \citep{wang2024mmlupro}.

\subsection{Models}

The main experiments use four generator configurations: Llama3.2-3B-instruct, Llama3.1-8B-instruct \citep{grattafiori2024llama}, Qwen3-4B-instruct-2507, and Qwen3-30B-A3B-instruct-2507 \citep{yang2025qwen3}, served with vLLM \citep{kwon2023vllm}. For each method, all LLM-based components use the same model.

\subsection{Compared Methods}
We compare HYPER with ten baselines from three families: no-retrieval or
simple retrieval, post-retrieval selection, and query transformation.
Detailed implementation choices, prompts, and hyperparameter settings are
provided in Appendix~\ref{app:experimental_details}.

\paragraph{No retrieval and simple retrieval.}
\textsc{CoT} answers directly without retrieval, while
\textsc{Simple RAG} retrieves the top-15 documents using the raw question
as a single query and passes them to the generator.

\paragraph{Post-retrieval selection.}
These methods first retrieve candidates with the raw question and then
select, filter, or reorder the final context.
\textsc{MMR-RAG} retrieves 150 candidates and applies maximal marginal
relevance as a diversity-aware selection criterion.
\textsc{Rerank-RAG} uses the same 150-candidate pool and reranks it with
the BGE reranker v2-m3 cross-encoder.
\textsc{MAIN-RAG} starts from the top-15 retrieved documents and filters
them using LLM evaluations of question-document-answer triples
\citep{chang2025main}.
\textsc{ITEM-A} also starts from the top-15 retrieved documents and
iteratively alternates pseudo-answer generation with LLM-based utility
judgments to select useful passages \citep{zhang2024iterative}.
Because \textsc{MAIN-RAG} and \textsc{ITEM-A} filter the initial context,
their final context may contain fewer than 15 documents.

\paragraph{Query transformation.}
These methods rewrite or expand the query before retrieval.
\textsc{RRR} rewrites the input query before retrieval \citep{ma2023query}.
\textsc{LC-MQR} is a LangChain-style MultiQueryRetriever baseline that
generates paraphrase-style query variants and fuses their retrieved results;
we follow the LC-MQR baseline setting used in prior work
\citep{kuo2025mmlf}.
\textsc{DMQR} generates diverse rewrites using information-level strategies,
including general rewriting, keyword rewriting, pseudo-answer rewriting,
and core-content extraction \citep{li2024dmqr}.
\textsc{HyDE} generates hypothetical documents, encodes them with the
retriever, averages their embeddings, and retrieves with the averaged query
representation \citep{gao2023precise}.
For \textsc{LC-MQR} and \textsc{DMQR}, we use the same retrieval-fusion
procedure as HYPER, so the final context may contain fewer than 15 unique
documents after deduplication.

\subsection{Retriever, Corpus, and Budget}

All retrieval-based methods use the same retriever and corpus so that
performance differences reflect retrieval strategy rather than backend
changes. We use MedCPT \citep{jin2023medcpt} over the medical textbook
corpus of \citet{xiong2024benchmarking}. The maximum final-context
budget is set to $B_{\max}=15$ documents. For broader-domain evaluation on MMLU-Pro, we use BGE-M3 \citep{multi2024m3} over the FineWeb-Edu 10BT sample \citep{penedo2024fineweb} with the same budget.

We normalize comparisons by the evidence shown to the final generator,
rather than by the number of intermediate candidates inspected. Methods
therefore retain their own candidate-pool rules: single-query methods
retrieve up to 15 documents directly; HYPER and LC-MQR retrieve top-5
per query, while DMQR divides 15 documents across up to five queries.
These results are fused with deduplication. Post-retrieval
selection methods may inspect larger or method-specific candidate pools
before producing the final context. Consequently, the realized final
context can contain fewer than 15 documents when fusion or filtering
removes candidates. Detailed candidate-pool choices for each baseline are
provided in Appendix~\ref{app:experimental_details}.

\begin{table*}[!t]
\centering
\fontsize{8}{8.5}\selectfont
\setlength{\tabcolsep}{3.0pt}
\caption{Accuracy (\%)  on MedQA and MMLU-Med for HYPER and baseline methods under the shared 15-document final-context budget. HYPER achieves the best average performance on both datasets. Significance tests for every cell are in Table~\ref{tab:sig_matrix}.}
\label{tab:main-results}
\begin{tabular}{@{}lccccc ccccc@{}}
\toprule
& \multicolumn{5}{c}{MedQA} & \multicolumn{5}{c}{MMLU-Med} \\
\cmidrule(lr){2-6} \cmidrule(l){7-11}
Method
& {\fontsize{7.2}{7.8}\selectfont Llama3.2 3B}
& {\fontsize{7.2}{7.8}\selectfont Llama3.1 8B}
& {\fontsize{7.2}{7.8}\selectfont Qwen3 4B}
& {\fontsize{7.2}{7.8}\selectfont Qwen3 30B}
& Average
& {\fontsize{7.2}{7.8}\selectfont Llama3.2 3B}
& {\fontsize{7.2}{7.8}\selectfont Llama3.1 8B}
& {\fontsize{7.2}{7.8}\selectfont Qwen3 4B}
& {\fontsize{7.2}{7.8}\selectfont Qwen3 30B}
& Average \\
\midrule
CoT                  & 58.0 & 69.0 & 72.8 & 84.1 & 71.0 & 66.3 & 75.0 & 82.9 & 90.2 & 78.6 \\
Simple RAG           & 55.6 & 66.8 & 71.3 & 82.7 & 69.1 & 64.5 & 74.7 & 81.8 & 88.9 & 77.5 \\
\midrule
\addlinespace[0.35em]
\multicolumn{11}{@{}l}{\textbf{Post-retrieval selection}} \\
\addlinespace[0.15em]
\quad MMR-RAG           & 54.4&	67.0 &71.8&	82.8& 69.0 & \textbf{69.0}	&75.8	&82.2&	89.3&	79.1 \\
\quad Rerank-RAG           & 56.8 & 67.3 & 72.1 & 83.3 & 69.9 & 67.1 & 75.8 & 82.3 & 90.1 & 78.8 \\
\quad MAIN-RAG             & 58.1 & 66.5 & 71.4 & 82.9 & 69.7 & 67.8 & 75.1 & 81.7 & 89.0 & 78.4   \\
\quad ITEM-A           & 55.5	&65.0	&69.8	&81.8	&68.0 & 63.7	&72.5	&82.0	&89.3	&76.9\\
\midrule
\addlinespace[0.35em]
\multicolumn{11}{@{}l}{\textbf{Query transformation}} \\
\addlinespace[0.15em]
\quad RRR          & 60.1	&72.0&	73.6	&81.9	&71.9 & 68.7&74.0&83.7&89.4&79.0 \\
\quad LC-MQR           & 58.9 & 67.5 & 73.1 & 83.3 & 70.7 & 67.0 & 74.7 & 82.5 & 88.8 & 78.2 \\
\quad DMQR     & 58.0	& 70.3	& 73.4	& 84.1	& 71.4& 66.1 & 76.5	& 84.5	&90.2&79.3\\
\quad HyDE     & 57.9 & 69.3 & 71.4 & 82.9 & 70.4 & 67.8 & 76.5 & 82.6 & 89.4 & 79.1 \\
\midrule
\textbf{HYPER (ours)} & \textbf{63.4} & \textbf{73.0} & \textbf{77.9} & \textbf{85.8} & \textbf{75.0}
                     & 68.6 & \textbf{79.7} & \textbf{85.4} & \textbf{90.6} & \textbf{81.1} \\
\bottomrule
\end{tabular}
\end{table*}

\begin{table*}[!t]
\centering
\fontsize{8}{8.5}\selectfont
\setlength{\tabcolsep}{3.0pt}
\caption{Decision-Useful Rate (\textsc{Entailed} + \textsc{Useful}) on MedQA and MMLU-Med for HYPER and baseline methods. HYPER yields the highest rate of decision-useful retrieved contexts across all reported dataset-model cells. Simple RAG, MMR-RAG, and Rerank-RAG retrieve the same context for every generator, so their values repeat across columns.}
\label{tab:dur-main}
\begin{tabular}{@{}lccccc ccccc@{}}
\toprule
& \multicolumn{5}{c}{MedQA} & \multicolumn{5}{c}{MMLU-Med} \\
\cmidrule(lr){2-6} \cmidrule(l){7-11}
Method
& {\fontsize{7.2}{7.8}\selectfont Llama3.2 3B}
& {\fontsize{7.2}{7.8}\selectfont Llama3.1 8B}
& {\fontsize{7.2}{7.8}\selectfont Qwen3 4B}
& {\fontsize{7.2}{7.8}\selectfont Qwen3 30B}
& Average
& {\fontsize{7.2}{7.8}\selectfont Llama3.2 3B}
& {\fontsize{7.2}{7.8}\selectfont Llama3.1 8B}
& {\fontsize{7.2}{7.8}\selectfont Qwen3 4B}
& {\fontsize{7.2}{7.8}\selectfont Qwen3 30B}
& Average \\
\midrule
Simple RAG           & 30.0 & 30.0 & 30.0 & 30.0 & 30.0 & 48.7 & 48.7 & 48.7 & 48.7 & 48.7 \\
\midrule
\addlinespace[0.35em]
\multicolumn{11}{@{}l}{\textbf{Post-retrieval selection}} \\
\addlinespace[0.15em]
\quad MMR-RAG           & 26.4 & 26.4 & 26.4 & 26.4 & 26.4 & 43.1 & 43.1 & 43.1 & 43.1 & 43.1 \\
\quad Rerank-RAG           & 44.1 & 44.1 & 44.1 & 44.1 & 44.1 & 56.8 & 56.8 & 56.8 & 56.8 & 56.8 \\
\quad MAIN-RAG             & 27.4 & 29.1 & 27.7 & 29.7 & 28.5 & 44.7 & 47.7 & 46.3 & 48.8 & 46.9 \\
\quad ITEM-A           &23.5&25.7&25.2&26.9&25.3&41.0&	43.4&43.6&45.5&43.4\\
\midrule
\addlinespace[0.35em]
\multicolumn{11}{@{}l}{\textbf{Query transformation}} \\
\addlinespace[0.15em]
\quad RRR          &62.0&68.4&65.4&66.8&65.7& 48.1&55.5&59.7&63.3&56.6 \\
\quad LC-MQR            & 45.2 & 51.2 & 62.2 & 65.6 & 56.1 & 51.4 & 53.7 & 56.7 & 59.1 & 55.3 \\
\quad DMQR     &52.4&58&62.2&72.1&61.2 &52.2&56.2&57.1&59.4&56.2 \\
\quad HyDE                 & 51.6 & 54.5 & 52.8 & 54.6 & 53.4 & 43.0 & 46.2 & 44.3 & 46.3 & 44.9 \\

\midrule
\textbf{HYPER (ours)} & \textbf{73.4} & \textbf{81.8} & \textbf{83.2} & \textbf{89.8} & \textbf{82.1} & \textbf{65.8} & \textbf{69.1} & \textbf{72.5} & \textbf{73.7} & \textbf{70.3} \\
\bottomrule
\end{tabular}
\end{table*}

\subsection{Prompting and Implementation Details}
For final answer generation on MedQA and MMLU-Med, all retrieval-based methods use the same MIRAGE prompt \citep{xiong2024benchmarking}. This keeps the generation interface fixed across these methods. Prompt templates and implementation details are provided in Appendix~\ref{app:experimental_details}.

\subsection{Context-Utility Evaluation}

Accuracy alone does not distinguish improvements due to better retrieved evidence from improvements caused by other changes in generation behavior. We therefore evaluate retrieved context sets directly using a separate LLM judge. For each question, method, and model setting, the judge first predicts the answer option most strongly supported by the retrieved context alone (\texttt{selected\_answer}) and then assigns a utility label describing how that context relates to the gold answer.

The judge uses three labels. For brevity, we abbreviate the prompt label \textsc{Useful-but-Not-Entailed} as \textsc{Useful} in tables and formulas.

\begin{itemize}[leftmargin=*, itemsep=2pt, topsep=2pt]
    \item \textbf{\textsc{Entailed}}: the gold answer follows decisively from the retrieved context alone, with no missing fact or outside knowledge, and the context-only \texttt{selected\_answer} matches the gold answer.
    \item \textbf{\textsc{Useful}}: the retrieved context provides strong decision-relevant support for the gold answer but lacks one final inferential bridge. This includes cases where the context contains a decisive clue, rules out plausible alternatives, or makes the gold answer reachable with minimal additional knowledge.
    \item \textbf{\textsc{Not Useful}}: the retrieved context does not provide strong decision-relevant support for the gold answer. This includes background-only passages, keyword overlap without meaningful narrowing power, weak associations, and cases where the context supports another answer more strongly than the gold answer.
\end{itemize}
Specifically, let $N_d$ denote the number of questions in dataset $d$, and let $N_{\textsc{E}}(d,m,g)$ and $N_{\textsc{U}}(d,m,g)$ be the numbers labeled \textsc{Entailed} and \textsc{Useful}, respectively, under method $m$ and generator $g$. Then, \textbf{Decision-Useful Rate} (DUR) is defined as
\[
\mathrm{DUR}(d,m,g)
=
\frac{N_{\textsc{E}}(d,m,g)+N_{\textsc{U}}(d,m,g)}{N_d}.
\]
DUR summarizes the fraction of questions for which retrieval yields context that is either decisive or meaningfully helpful for the correct answer.
We also report accuracy within each utility category. To test whether these labels correspond to downstream performance differences, we compare each method against no-retrieval CoT on the same subset of questions:
\[
\begin{aligned}
\Delta_{\mathrm{CoT}}(d,m,g \mid \ell)
&=
\mathrm{Acc}(d,m,g \mid \ell) \\
&\quad -
\mathrm{Acc}(d,\text{CoT},g \mid \ell).
\end{aligned}
\]
where the second term evaluates the same model's no-retrieval system on the subset of questions whose retrieved context is assigned label $\ell$ under method $m$. This same-subset comparison controls for differences in subset composition.

The context-utility evaluation is conducted using GPT-4.1 as the judge model, with the prompt in Figure~\ref{fig:prompt_gpt_judge}. Because the judge sees the gold answer, Section~\ref{sec:context-utility} validates its labels against human annotators, a second judge, and a gold-blind protocol.

\section{Results}

\begin{table}[!t]
\centering
\small
\caption{Average context-utility distribution across MedQA and MMLU-Med (\%). Values are averaged over the two datasets and four model configurations. \textsc{Useful} abbreviates \textsc{Useful-but-Not-Entailed}.}
\label{tab:utility-breakdown}
\begin{tabular}{lccc}
\toprule
Method & Entailed & Useful & Not Useful \\
\midrule
Simple RAG            & 14.3 & 25.1 & 60.7 \\
\midrule
MMR-RAG                 & 11.6 & 23.2 & 65.3  \\
Rerank-RAG            & 19.9 & 30.6 & 49.5 \\
MAIN-RAG              & 15.1 & 22.6 & 62.3 \\
ITEM-A                & 15.2 & 19.2 & 65.6 \\
\midrule
RRR                  & 24.9 & 36.3 & 38.9 \\
LC-MQR             & 19.6 & 36.1 & 44.3 \\
DMQR                 &22.9 & 35.8 & 41.3 \\
HyDE                  & 16.6 & 32.5 & 50.8 \\
\midrule
\textbf{HYPER (ours)}  & 34.7 & 41.5 & 23.8 \\
\bottomrule
\end{tabular}
\end{table}

\begin{table}[!t]
\centering
\small
\caption{Accuracy by utility label and same-subset gain over CoT, averaged over four models and ten methods. Each cell is \emph{accuracy / $\Delta_{\mathrm{CoT}}$}.}
\label{tab:subset-utility}
\begin{tabular}{lcc}
\toprule
Label & MedQA & MMLU-Med \\
\midrule
\textsc{Entailed} & 92.9 / +6.8 & 93.8 / +5.1 \\
\textsc{Useful} & 78.1 / +4.3 & 82.3 / +2.0 \\
\textsc{Not Useful} & 57.7 / -5.7 & 68.8 / -3.4 \\
\bottomrule
\end{tabular}
\end{table}

\subsection{Main accuracy results}

Table~\ref{tab:main-results} reports answer accuracy under the shared 15-document final-context budget. Simple retrieval with the raw question tends to hurt: \textsc{Simple RAG} underperforms no-retrieval \textsc{CoT} in every dataset-model cell. These results highlight the importance of evidence aligned with the decision.

\textsc{HYPER} gives the strongest overall performance. Averaged across models, it improves over \textsc{Simple RAG} by 5.9 points on MedQA and 3.6 points on MMLU-Med, and it achieves the best average accuracy on both datasets. The gains are especially clear on MedQA, where HYPER outperforms all baselines in every model setting. On MMLU-Med, HYPER also obtains the best average accuracy, although MMR-RAG and RRR are slightly higher than HYPER for Llama3.2-3B.

These gains are statistically reliable on MedQA and less uniform on MMLU-Med. Of the 80 exact McNemar comparisons with the ten baselines, 57 remain significant after Benjamini--Hochberg correction and 35 under Holm correction, with 36 of 40 on MedQA and 21 of 40 on MMLU-Med under Benjamini--Hochberg correction.  All eight paired bootstrap intervals for the difference from Simple RAG exclude zero, and five sampled runs of Qwen3-4B reproduce the ordering (Appendix~\ref{app:stats}).

The comparison across baseline families clarifies the source of HYPER's gains. Post-retrieval selection methods provide limited improvements: \textsc{Rerank-RAG} improves over \textsc{Simple RAG}, but remains below HYPER on average, while \textsc{MAIN-RAG} and \textsc{ITEM-A} stay close to \textsc{Simple RAG} on average despite filtering the retrieved context.  Query-transformation baselines such as \textsc{RRR}, \textsc{LC-MQR}, \textsc{DMQR}, and \textsc{HyDE} are more competitive, but still remain below HYPER on average. 

\subsection{Context-utility analysis}
\label{sec:context-utility}

The context-utility analysis shows that HYPER changes the quality of the retrieved context as well as the final answer. Table~\ref{tab:dur-main} shows that HYPER achieves the highest DUR in every dataset-model cell, with especially large gains on MedQA.  

Table~\ref{tab:utility-breakdown} summarizes the context-utility distribution across the two datasets and four generators. Compared with Simple RAG, HYPER increases both \textsc{Entailed} and \textsc{Useful} contexts while reducing \textsc{Not Useful} contexts. The per-setting distributions are in Table~\ref{tab:context-label-full}. Table~\ref{tab:subset-utility} validates these labels against downstream behavior: \textsc{Entailed} contexts yield the highest accuracy and largest gain over no-retrieval \textsc{CoT}, \textsc{Useful} contexts remain beneficial, and \textsc{Not Useful} contexts fall below the no-retrieval baseline. Together, these results show that higher accuracy accompanies
more decision-useful retrieved evidence. 

Because the judge sees the gold answer, we validated its labels in three ways (Appendix~\ref{app:judge}). Two of the authors independently labeled 60 Qwen3-4B MedQA contexts, 30 from HYPER and 30 from Simple RAG. On the DUR distinction, agreement was high between the authors (Cohen's $\kappa$=.93) and with GPT-4.1 ($\kappa$=.73), and both reproduced HYPER's advantage. Agreement between each author and GPT-4.1 on the exact three-way label was lower ($\kappa$=.43--.48), so we use DUR as the primary measure. A second judge, Qwen3.5-27B \citep{qwen2026qwen35}, also agreed closely with GPT-4.1 across 23,620 contexts ($\kappa$=.84). A gold-blind judge preserved the method ordering on all 1,273 Qwen3-4B MedQA questions (Table~\ref{tab:gold_blind}). Its labels measure usefulness for the option best supported by the context, which can differ from the gold answer.

\subsection{Controls and ablations}
\label{sec:controls-ablations}

\begin{table}[!t]
\centering
\small
\setlength{\tabcolsep}{4pt}
\caption{Diagnostic controls and query-role ablation (accuracy, \%, four-generator averages).}
\label{tab:option-control}\label{tab:query-ablation}
\begin{tabular}{lcc}
\toprule
Method & MedQA & MMLU-Med \\
\midrule
LC-MQR                & 70.7 & 78.2 \\
LC-MQR $+$ options    & 71.3 & 79.2 \\
\midrule
\textbf{HYPER}        & \textbf{75.0} & \textbf{81.1} \\
HYPER $-$ options     & 73.1 & 79.9 \\
HYPER $-$ $q_{\mathrm{sup}}$  & 72.4 & 79.0 \\
HYPER $-$ $q_{\mathrm{dist}}$ & 73.2 & 80.1 \\
HYPER $-$ $q_{\mathrm{feat}}$ & 73.0 & 80.5 \\
\bottomrule
\end{tabular}
\end{table}

We next examine which components of HYPER are responsible for the observed gains. Table~\ref{tab:option-control} evaluates whether improvements arise primarily from exposing answer options during retrieval planning. Adding answer options to a generic multi-query formulator yields only a modest gain over the original LC-MQR, whereas HYPER remains clearly stronger. This gap indicates that most of HYPER's benefit comes from organizing retrieval around a verification agenda induced by the provisional answer hypothesis, beyond what conditioning on the answer choices provides. 

Table~\ref{tab:query-ablation} evaluates the contribution of the three rewritten queries. Removing any one of the query roles reduces accuracy on both datasets. The largest decrease follows removal of $q_{\mathrm{sup}}$, the answer-support query, but removing $q_{\mathrm{dist}}$ or $q_{\mathrm{feat}}$ also degrades performance. These results indicate that the three query roles provide complementary contributions.

HYPER also improves over LC-MQR under the same three-query budget. LC-MQR issues as many queries as HYPER and MMR-RAG selects diverse documents from 150 candidates, yet both stay close to Simple RAG. DMQR expanded to five queries and 25 final documents remains below HYPER on both datasets (Table~\ref{tab:dmqr25}), and HYPER's own accuracy peaks at 15 documents. Across budgets from 3 to 30 documents, it has the highest accuracy and DUR on both datasets (Table~\ref{tab:budget_sensitivity_qwen}). HYPER also stays best under permuted option orders (Table~\ref{tab:option_order}) and with a PubMed corpus, and a stronger planner improves smaller generators (Appendix~\ref{app:robustness}).

\begin{table}[!t]
\centering
\small
\setlength{\tabcolsep}{3.2pt}
\caption{MMLU-Pro accuracy (\%) on all 12,032 test questions, retrieving from FineWeb-Edu under the 15-document budget. Per-subject results are in Table~\ref{tab:mmlupro-subject}.}
\label{tab:mmlupro}
\begin{tabular}{lccccc}
\toprule
Method & {\fontsize{7.2}{7.8}\selectfont Llama 3B} & {\fontsize{7.2}{7.8}\selectfont Llama 8B} & {\fontsize{7.2}{7.8}\selectfont Qwen 4B} & {\fontsize{7.2}{7.8}\selectfont Qwen 30B} & Avg. \\
\midrule
CoT        & 23.2 & 44.9 & 65.2 & \textbf{73.2} & 51.6 \\
Simple RAG & 27.0 & 46.9 & 66.7 & 72.5 & 53.3 \\
LC-MQR     & 23.3 & 46.8 & 66.4 & 73.0 & 52.4 \\
\textbf{HYPER} & \textbf{33.4} & \textbf{48.3} & \textbf{67.2} & 72.7 & \textbf{55.4} \\
\bottomrule
\end{tabular}
\end{table}

\begin{table}[!t]
\centering
\small
\setlength{\tabcolsep}{4pt}
\caption{Accuracy (\%) by correctness of HYPER's provisional answer hypothesis, pooled over all 9,448 question--generator pairs. ``Wrong'' includes unparsable hypotheses. Per-setting values are in Table~\ref{tab:wrong_by_run}.}
\label{tab:hypothesis-split}
\begin{tabular}{lcc}
\toprule
Method & Correct ($N$=6,888) & Wrong ($N$=2,560) \\
\midrule
CoT          & 88.9 & 35.7 \\
Simple RAG   & 85.5 & 39.3 \\
LC-MQR       & 86.2 & \textbf{41.9} \\
\midrule
\textbf{HYPER} & \textbf{91.9} & 39.8 \\
HYPER $-$ $q_{\mathrm{sup}}$  & 89.4 & 37.9 \\
HYPER $-$ $q_{\mathrm{dist}}$ & 91.2 & 36.6 \\
HYPER $-$ $q_{\mathrm{feat}}$ & 90.7 & 38.3 \\
\bottomrule
\end{tabular}
\end{table}

\subsection{Transfer beyond medicine}
\label{sec:transfer}
HYPER also transfers to MMLU-Pro with a general web corpus, achieving the best average accuracy across the four generators (Table~\ref{tab:mmlupro}; per-subject results in Table~\ref{tab:mmlupro-subject}). Gains are larger for the smaller generators, while Qwen3-30B performs better with CoT than with HYPER or the other retrieval methods. Gains also tend to narrow for stronger generators on the medical benchmarks, particularly GPT-4.1 (Table~\ref{tab:gpt41}). One possible explanation is that, as a model answers more questions from its own knowledge, fewer remaining errors concern questions for which the corpus provides useful evidence.

\subsection{When the provisional answer hypothesis is wrong}
\label{sec:hypothesis}
Because the support query follows the provisional answer hypothesis, a wrong hypothesis could steer retrieval toward the wrong option. Table~\ref{tab:hypothesis-split} splits results by whether the hypothesis is correct. When it is correct, HYPER outperforms CoT, Simple RAG, and LC-MQR. When it is wrong, HYPER remains above CoT and Simple RAG on average but falls 2.1 points below LC-MQR. Removing $q_{\mathrm{sup}}$ costs most when the hypothesis is correct, and removing $q_{\mathrm{dist}}$ costs most when it is wrong. On this subset HYPER trails Simple RAG in two of eight settings (Table~\ref{tab:wrong_by_run}).

The final answer repeats the wrong hypothesis in 41.3\% of these cases (Table~\ref{tab:correction_confirmation}). Removing $q_{\mathrm{dist}}$ lowers correction and increases this confirmation failure rate. For initially correct hypotheses, removing $q_{\mathrm{sup}}$ increases the rate of incorrect final answers. These ablations keep the same provisional hypotheses and remove one query's retrieved documents before rerunning the generator.

\subsection{Computational cost}
\label{sec:cost}
Compared with single-query RAG, HYPER adds two LLM calls and two retrieval queries per question (Table~\ref{tab:compute_budget}). Appendix~\ref{app:cost} reports end-to-end runtime and accuracy from the timing runs.

\section{Discussion}
\label{sec:discussion}

 The results suggest that HYPER's main advantage comes from steering
first-pass retrieval toward evidence needed for the decision, rather than
retrieving more broadly, diversifying the result set, or filtering after
retrieval. The increase in ENTAILED/USEFUL contexts supports this
mechanism. This design also mitigates a potential confirmation-bias
concern: only the support query directly follows the provisional answer hypothesis, while the distinction and feature queries can surface evidence
that challenges it by separating alternatives or grounding question-specific
clues. Section~\ref{sec:hypothesis} and Appendix~\ref{app:correction} quantify correction and confirmation of the hypothesis, and Appendix~\ref{app:examples} shows corrected and failed cases.

The filtering baselines illustrate a limitation of post-hoc selection in this setting. \textsc{MAIN-RAG} and \textsc{ITEM-A} reduce context size, but their accuracy remains close to \textsc{Simple RAG} on average. One possible explanation is that, in complex medical decision questions, utility is often distributed across partial clues, criteria, and contrastive evidence, making it difficult to separate removable noise from evidence that becomes decisive only in combination. Filtering can therefore discard documents that appear weak individually but are needed collectively to resolve the decision. Comparisons with the diversity and query-transformation baselines likewise point to gains from explicit evidence planning beyond coverage and query variation.

\section{Conclusion}
\label{sec:conclusion}

We introduced HYPER, a training-free retrieval planning method that converts a provisional answer hypothesis into evidence-seeking queries. Across MedQA and MMLU-Med, HYPER improves over single-query RAG in every setting and outperforms strong baselines on average under a shared final-context budget. Its gains also transfer to MMLU-Pro. These results suggest that, for bounded-context decision tasks, retrieval is most effective when it is planned around the evidence needed to resolve competing hypotheses.

\section*{Limitations}

This study has several limitations.

First, HYPER depends on the quality of the intermediate retrieval-planning state. If the provisional answer hypothesis, discriminating features, or evidence needs are inaccurate or incomplete, the resulting rewritten queries may be misaligned with the true decision boundary. The distinction and feature queries are intended to mitigate this risk and only partly offset it. When the hypothesis is wrong, HYPER is 2.1 points less accurate than LC-MQR on average (Section~\ref{sec:hypothesis}). HYPER also adds two LLM calls per question; Appendix~\ref{app:cost} reports the computational cost.

Second, our empirical evaluation is concentrated on medical multiple-choice QA. This is a deliberate choice because the distinction between topical relevance and decision-usefulness is particularly clear in this setting. However, it leaves open the extent to which the same design transfers to other domains or to open-ended tasks in which candidate answers must first be hypothesized. Our MMLU-Pro evaluation covers non-medical subjects against three baselines, and the size of the gain varies widely across generators.

Third, the context-utility analysis relies on an LLM judge. We use common label definitions across methods and use the resulting labels as a comparative measure that complements human annotation. Agreement with two human annotators is high for the binary DUR distinction but moderate for the three-way label, and the human sample covers 60 contexts from a single setting. Accordingly, the utility analysis is best understood as supporting evidence for the proposed mechanism rather than as a definitive causal characterization of retrieval quality.

Finally, our main comparisons use a fixed corpus, retriever, and final-context budget within each benchmark. The comparisons are normalized by the evidence shown to the final generator rather than by total retrieval-time computation, and some baselines inspect larger intermediate candidate pools by design. Transfer of the reported gains to open-web retrieval, substantially larger corpora, or fully agentic multi-turn retrieval settings remains to be tested.

\section*{Ethical Considerations}

Our primary experiments are conducted in the medical domain, where errors in retrieval or answer generation may have serious real-world consequences. This work studies retrieval behavior on benchmark QA tasks and is intended for research use only. Any real-world medical deployment would require substantially stronger validation, including expert review, provenance-aware interfaces, calibrated uncertainty handling, abstention mechanisms, and systematic monitoring for harmful retrieval failures.

More broadly, methods that condition retrieval on a provisional answer hypothesis may introduce additional risks if the intermediate hypothesis is inaccurate or systematically biased. Our default design partially mitigates this concern by restricting the role of the hypothesis to retrieval planning rather than exposing it directly to the final generator. However, this design choice only partly reduces the risk of biased or misdirected retrieval. For example, a planner that favors common diagnoses may under-retrieve evidence for rare diseases, and demographic biases in the hypothesis could carry over into the retrieved evidence. Careful evaluation remains necessary in any high-stakes application.

\section*{Acknowledgements}

This research was supported by a grant of the Korea Health Technology R\&D Project through the Korea Health Industry Development Institute (KHIDI), funded by the Ministry of Health \& Welfare, Republic of Korea (grant number : RS-2025-02213531). This work was supported by the National Research Foundation of Korea under Grant RS-2024-00336454. This work was supported by Artificial intelligence industrial convergence cluster development project funded by the Ministry of Science and ICT(MSIT, Korea)\&Gwangju Metropolitan City.

AI assistants were used for coding support and refinement of parts of the manuscript.

\bibliography{HYPER_eacl2027_final}

@inproceedings{fan2024survey,
author = {Fan, Wenqi and Ding, Yujuan and Ning, Liangbo and Wang, Shijie and Li, Hengyun and Yin, Dawei and Chua, Tat-Seng and Li, Qing},
title = {A Survey on RAG Meeting LLMs: Towards Retrieval-Augmented Large Language Models},
year = {2024},
isbn = {9798400704901},
publisher = {Association for Computing Machinery},
address = {New York, NY, USA},
url = {https://doi.org/10.1145/3637528.3671470},
doi = {10.1145/3637528.3671470},
booktitle = {Proceedings of the 30th ACM SIGKDD Conference on Knowledge Discovery and Data Mining},
pages = {6491–6501},
numpages = {11},
location = {Barcelona, Spain},
series = {KDD '24}
}

@misc{huang2024survey,
      title={A Survey on Retrieval-Augmented Text Generation for Large Language Models}, 
      author={Yizheng Huang and Jimmy Huang},
      year={2024},
      eprint={2404.10981},
      archivePrefix={arXiv},
      primaryClass={cs.IR},
      doi={https://doi.org/10.1145/3805774},
      url={https://arxiv.org/abs/2404.10981}, 
}

@inproceedings{lewis2020retrieval,
 author = {Lewis, Patrick and Perez, Ethan and Piktus, Aleksandra and Petroni, Fabio and Karpukhin, Vladimir and Goyal, Naman and K\"{u}ttler, Heinrich and Lewis, Mike and Yih, Wen-tau and Rockt\"{a}schel, Tim and Riedel, Sebastian and Kiela, Douwe},
 booktitle = {Advances in Neural Information Processing Systems},
 editor = {H. Larochelle and M. Ranzato and R. Hadsell and M.F. Balcan and H. Lin},
 pages = {9459--9474},
 publisher = {Curran Associates, Inc.},
 title = {Retrieval-Augmented Generation for Knowledge-Intensive NLP Tasks},
 url = {https://proceedings.neurips.cc/paper_files/paper/2020/file/6b493230205f780e1bc26945df7481e5-Paper.pdf},
 volume = {33},
 year = {2020}
}

@inproceedings{amiraz2025distracting,
    title = "The Distracting Effect: Understanding Irrelevant Passages in {RAG}",
    author = "Amiraz, Chen  and
      Cuconasu, Florin  and
      Filice, Simone  and
      Karnin, Zohar",
    editor = "Che, Wanxiang  and
      Nabende, Joyce  and
      Shutova, Ekaterina  and
      Pilehvar, Mohammad Taher",
    booktitle = "Proceedings of the 63rd Annual Meeting of the {A}ssociation for {C}omputational {L}inguistics (Volume 1: Long Papers)",
    month = jul,
    year = "2025",
    address = "Vienna, Austria",
    publisher = "Association for Computational Linguistics",
    url = "https://aclanthology.org/2025.acl-long.892/",
    doi = "10.18653/v1/2025.acl-long.892",
    pages = "18228--18258",
    ISBN = "979-8-89176-251-0"
}

@misc{yoran2023making,
      title={Making Retrieval-Augmented Language Models Robust to Irrelevant Context}, 
      author={Ori Yoran and Tomer Wolfson and Ori Ram and Jonathan Berant},
      year={2024},
      eprint={2310.01558},
      archivePrefix={arXiv},
      primaryClass={cs.CL},
      url={https://arxiv.org/abs/2310.01558}, 
}

@misc{du2025context,
      title={Context Length Alone Hurts LLM Performance Despite Perfect Retrieval}, 
      author={Yufeng Du and Minyang Tian and Srikanth Ronanki and Subendhu Rongali and Sravan Bodapati and Aram Galstyan and Azton Wells and Roy Schwartz and Eliu A Huerta and Hao Peng},
      year={2025},
      eprint={2510.05381},
      archivePrefix={arXiv},
      primaryClass={cs.CL},
      url={https://arxiv.org/abs/2510.05381}, 
}

@inproceedings{ma2023query,
    title = "Query Rewriting in Retrieval-Augmented Large Language Models",
    author = "Ma, Xinbei  and
      Gong, Yeyun  and
      He, Pengcheng  and
      Zhao, Hai  and
      Duan, Nan",
    editor = "Bouamor, Houda  and
      Pino, Juan  and
      Bali, Kalika",
    booktitle = "Proceedings of the 2023 Conference on Empirical Methods in Natural Language Processing",
    month = dec,
    year = "2023",
    address = "Singapore",
    publisher = "Association for Computational Linguistics",
    url = "https://aclanthology.org/2023.emnlp-main.322/",
    doi = "10.18653/v1/2023.emnlp-main.322",
    pages = "5303--5315"
}

@inproceedings{gao2023precise,
    title = "Precise Zero-Shot Dense Retrieval without Relevance Labels",
    author = "Gao, Luyu  and
      Ma, Xueguang  and
      Lin, Jimmy  and
      Callan, Jamie",
    editor = "Rogers, Anna  and
      Boyd-Graber, Jordan  and
      Okazaki, Naoaki",
    booktitle = "Proceedings of the 61st Annual Meeting of the {A}ssociation for {C}omputational {L}inguistics (Volume 1: Long Papers)",
    month = jul,
    year = "2023",
    address = "Toronto, Canada",
    publisher = "Association for Computational Linguistics",
    url = "https://aclanthology.org/2023.acl-long.99/",
    doi = "10.18653/v1/2023.acl-long.99",
    pages = "1762--1777"
}

@inproceedings{liang2025reasoning,
    title = "Reasoning {RAG} via System 1 or System 2: A Survey on Reasoning Agentic Retrieval-Augmented Generation for Industry Challenges",
    author = "Liang, Jintao  and
      Sugang  and
      Lin, Huifeng  and
      Wu, You  and
      Zhao, Rui  and
      Li, Ziyue",
    editor = "Inui, Kentaro  and
      Sakti, Sakriani  and
      Wang, Haofen  and
      Wong, Derek F.  and
      Bhattacharyya, Pushpak  and
      Banerjee, Biplab  and
      Ekbal, Asif  and
      Chakraborty, Tanmoy  and
      Singh, Dhirendra Pratap",
    booktitle = "Proceedings of the 14th International Joint Conference on Natural Language Processing and the 4th Conference of the {A}sia-Pacific Chapter of the {A}ssociation for {C}omputational {L}inguistics",
    month = dec,
    year = "2025",
    address = "Mumbai, India",
    publisher = "The Asian Federation of Natural Language Processing and The Association for Computational Linguistics",
    url = "https://aclanthology.org/2025.findings-ijcnlp.122/",
    doi = "10.18653/v1/2025.findings-ijcnlp.122",
    pages = "1954--1966",
    ISBN = "979-8-89176-303-6"
}

@inproceedings{trivedi2023interleaving,
    title = "Interleaving Retrieval with Chain-of-Thought Reasoning for Knowledge-Intensive Multi-Step Questions",
    author = "Trivedi, Harsh  and
      Balasubramanian, Niranjan  and
      Khot, Tushar  and
      Sabharwal, Ashish",
    editor = "Rogers, Anna  and
      Boyd-Graber, Jordan  and
      Okazaki, Naoaki",
    booktitle = "Proceedings of the 61st Annual Meeting of the {A}ssociation for {C}omputational {L}inguistics (Volume 1: Long Papers)",
    month = jul,
    year = "2023",
    address = "Toronto, Canada",
    publisher = "Association for Computational Linguistics",
    url = "https://aclanthology.org/2023.acl-long.557/",
    doi = "10.18653/v1/2023.acl-long.557",
    pages = "10014--10037"
}

@inproceedings{asai2024self,
 author = {Asai, Akari and Wu, Zeqiu and Wang, Yizhong and Sil, Avi and Hajishirzi, Hannaneh },
 booktitle = {International Conference on Learning Representations},
 editor = {B. Kim and Y. Yue and S. Chaudhuri and K. Fragkiadaki and M. Khan and Y. Sun},
 pages = {9112--9141},
 title = {Self-RAG: Learning to Retrieve, Generate, and Critique through Self-Reflection},
 url = {https://proceedings.iclr.cc/paper_files/paper/2024/file/25f7be9694d7b32d5cc670927b8091e1-Paper-Conference.pdf},
 volume = {2024},
 year = {2024}
}

@Article{jin2021medqa,
AUTHOR = {Jin, Di and Pan, Eileen and Oufattole, Nassim and Weng, Wei-Hung and Fang, Hanyi and Szolovits, Peter},
TITLE = {What Disease Does This Patient Have? A Large-Scale Open Domain Question Answering Dataset from Medical Exams},
JOURNAL = {Applied Sciences},
VOLUME = {11},
YEAR = {2021},
NUMBER = {14},
ARTICLE-NUMBER = {6421},
URL = {https://www.mdpi.com/2076-3417/11/14/6421},
ISSN = {2076-3417},
DOI = {10.3390/app11146421}
}

@misc{hendryckstest2021,
      title={Measuring Massive Multitask Language Understanding}, 
      author={Dan Hendrycks and Collin Burns and Steven Basart and Andy Zou and Mantas Mazeika and Dawn Song and Jacob Steinhardt},
      year={2021},
      eprint={2009.03300},
      archivePrefix={arXiv},
      primaryClass={cs.CY},
      url={https://arxiv.org/abs/2009.03300}, 
}

@article{jin2023medcpt,
    author = {Jin, Qiao and Kim, Won and Chen, Qingyu and Comeau, Donald C and Yeganova, Lana and Wilbur, W John and Lu, Zhiyong},
    title = {MedCPT: Contrastive Pre-trained Transformers with large-scale PubMed search logs for zero-shot biomedical information retrieval},
    journal = {Bioinformatics},
    volume = {39},
    number = {11},
    pages = {btad651},
    year = {2023},
    month = {11},
    issn = {1367-4811},
    doi = {10.1093/bioinformatics/btad651},
    url = {https://doi.org/10.1093/bioinformatics/btad651},
    eprint = {https://academic.oup.com/bioinformatics/article-pdf/39/11/btad651/52799559/btad651.pdf},
}

@inproceedings{xiong2024benchmarking,
    title = "Benchmarking Retrieval-Augmented Generation for Medicine",
    author = "Xiong, Guangzhi  and
      Jin, Qiao  and
      Lu, Zhiyong  and
      Zhang, Aidong",
    editor = "Ku, Lun-Wei  and
      Martins, Andre  and
      Srikumar, Vivek",
    booktitle = "Findings of the {A}ssociation for {C}omputational {L}inguistics: {ACL} 2024",
    month = aug,
    year = "2024",
    address = "Bangkok, Thailand",
    publisher = "Association for Computational Linguistics",
    url = "https://aclanthology.org/2024.findings-acl.372/",
    doi = "10.18653/v1/2024.findings-acl.372",
    pages = "6233--6251"
}

@inbook{xiong2024improving,
author = {Guangzhi Xiong and Qiao Jin and Xiao Wang and Minjia Zhang and Zhiyong Lu and Aidong Zhang},
title = {Improving Retrieval-Augmented Generation in Medicine with Iterative Follow-up Questions},
booktitle = {Biocomputing 2025},
chapter = {},
pages = {199-214},
doi = {10.1142/9789819807024_0015},
year = "2025",
URL = {https://www.worldscientific.com/doi/abs/10.1142/9789819807024_0015},
eprint = {https://www.worldscientific.com/doi/pdf/10.1142/9789819807024_0015}
}

@misc{kim2025rethinking,
      title={Rethinking Retrieval-Augmented Generation for Medicine: A Large-Scale, Systematic Expert Evaluation and Practical Insights}, 
      author={Hyunjae Kim and Jiwoong Sohn and Aidan Gilson and Nicholas Cochran-Caggiano and Serina Applebaum and Heeju Jin and Seihee Park and Yujin Park and Jiyeong Park and Seoyoung Choi and Brittany Alexandra Herrera Contreras and Thomas Huang and Jaehoon Yun and Ethan F. Wei and Roy Jiang and Leah Colucci and Eric Lai and Amisha Dave and Tuo Guo and Maxwell B. Singer and Yonghoe Koo and Ron A. Adelman and James Zou and Andrew Taylor and Arman Cohan and Hua Xu and Qingyu Chen},
      year={2025},
      eprint={2511.06738},
      archivePrefix={arXiv},
      primaryClass={cs.CL},
      url={https://arxiv.org/abs/2511.06738}, 
}

@inproceedings{yu2024rankrag,
 author = {Yu, Yue and Ping, Wei and Liu, Zihan and Wang, Boxin and You, Jiaxuan and Zhang, Chao and Shoeybi, Mohammad and Catanzaro, Bryan},
 booktitle = {Advances in Neural Information Processing Systems},
 doi = {10.52202/079017-3850},
 editor = {A. Globerson and L. Mackey and D. Belgrave and A. Fan and U. Paquet and J. Tomczak and C. Zhang},
 pages = {121156--121184},
 publisher = {Curran Associates, Inc.},
 title = {RankRAG: Unifying Context Ranking with Retrieval-Augmented Generation in LLMs},
 url = {https://proceedings.neurips.cc/paper_files/paper/2024/file/db93ccb6cf392f352570dd5af0a223d3-Paper-Conference.pdf},
 volume = {37},
 year = {2024}
}

@misc{yan2024corrective,
      title={Corrective Retrieval Augmented Generation}, 
      author={Shi-Qi Yan and Jia-Chen Gu and Yun Zhu and Zhen-Hua Ling},
      year={2024},
      eprint={2401.15884},
      archivePrefix={arXiv},
      primaryClass={cs.CL},
      url={https://arxiv.org/abs/2401.15884}, 
}

@inproceedings{chang2025main,
    title = "{MAIN}-{RAG}: Multi-Agent Filtering Retrieval-Augmented Generation",
    author = "Chang, Chia-Yuan  and
      Jiang, Zhimeng  and
      Rakesh, Vineeth  and
      Pan, Menghai  and
      Yeh, Chin-Chia Michael  and
      Wang, Guanchu  and
      Hu, Mingzhi  and
      Xu, Zhichao  and
      Zheng, Yan  and
      Das, Mahashweta  and
      Zou, Na",
    editor = "Che, Wanxiang  and
      Nabende, Joyce  and
      Shutova, Ekaterina  and
      Pilehvar, Mohammad Taher",
    booktitle = "Proceedings of the 63rd Annual Meeting of the {A}ssociation for {C}omputational {L}inguistics (Volume 1: Long Papers)",
    month = jul,
    year = "2025",
    address = "Vienna, Austria",
    publisher = "Association for Computational Linguistics",
    url = "https://aclanthology.org/2025.acl-long.131/",
    doi = "10.18653/v1/2025.acl-long.131",
    pages = "2607--2622",
    ISBN = "979-8-89176-251-0"
}

@inproceedings{salemi2024evaluating,
author = {Salemi, Alireza and Zamani, Hamed},
title = {Evaluating Retrieval Quality in Retrieval-Augmented Generation},
year = {2024},
isbn = {9798400704314},
publisher = {Association for Computing Machinery},
address = {New York, NY, USA},
url = {https://doi.org/10.1145/3626772.3657957},
doi = {10.1145/3626772.3657957},
booktitle = {Proceedings of the 47th International ACM SIGIR Conference on Research and Development in Information Retrieval},
pages = {2395–2400},
numpages = {6},
location = {Washington DC, USA},
series = {SIGIR '24}
}

@inproceedings{joren2025sufficient,
 author = {Joren, Hailey and Zhang, Jianyi and Ferng, Chun-Sung and Juan, Da-Cheng and Taly, Ankur and Rashtchian, Cyrus},
 booktitle = {International Conference on Learning Representations},
 editor = {Y. Yue and A. Garg and N. Peng and F. Sha and R. Yu},
 pages = {20310--20334},
 title = {Sufficient Context: A New Lens on Retrieval Augmented Generation Systems},
 url = {https://proceedings.iclr.cc/paper_files/paper/2025/file/33dffa2e3d2ab74a783d1a8c292f66d9-Paper-Conference.pdf},
 volume = {2025},
 year = {2025}
}

@inproceedings{kuo2025mmlf,
    title = "{MMLF}: Multi-query Multi-passage Late Fusion Retrieval",
    author = "Kuo, Yuan-Ching  and
      Yu, Yi  and
      Chen, Chih-Ming  and
      Wang, Chuan-Ju",
    editor = "Chiruzzo, Luis  and
      Ritter, Alan  and
      Wang, Lu",
    booktitle = "Findings of the {A}ssociation for {C}omputational {L}inguistics: {NAACL} 2025",
    month = apr,
    year = "2025",
    address = "Albuquerque, New Mexico",
    publisher = "Association for Computational Linguistics",
    url = "https://aclanthology.org/2025.findings-naacl.367/",
    doi = "10.18653/v1/2025.findings-naacl.367",
    pages = "6602--6613",
    ISBN = "979-8-89176-195-7"
}

@inproceedings{multi2024m3,
    title = "{M}3-Embedding: Multi-Linguality, Multi-Functionality, Multi-Granularity Text Embeddings Through Self-Knowledge Distillation",
    author = "Chen, Jianlyu  and
      Xiao, Shitao  and
      Zhang, Peitian  and
      Luo, Kun  and
      Lian, Defu  and
      Liu, Zheng",
    editor = "Ku, Lun-Wei  and
      Martins, Andre  and
      Srikumar, Vivek",
    booktitle = "Findings of the {A}ssociation for {C}omputational {L}inguistics: {ACL} 2024",
    month = aug,
    year = "2024",
    address = "Bangkok, Thailand",
    publisher = "Association for Computational Linguistics",
    url = "https://aclanthology.org/2024.findings-acl.137/",
    doi = "10.18653/v1/2024.findings-acl.137",
    pages = "2318--2335"
}

@inproceedings{zhang2024large,
author = {Zhang, Hengran and Zhang, Ruqing and Guo, Jiafeng and de Rijke, Maarten and Fan, Yixing and Cheng, Xueqi},
title = {Are Large Language Models Good at Utility Judgments?},
year = {2024},
isbn = {9798400704314},
publisher = {Association for Computing Machinery},
address = {New York, NY, USA},
url = {https://doi.org/10.1145/3626772.3657784},
doi = {10.1145/3626772.3657784},
booktitle = {Proceedings of the 47th International ACM SIGIR Conference on Research and Development in Information Retrieval},
pages = {1941–1951},
numpages = {11},
location = {Washington DC, USA},
series = {SIGIR '24}
}

@article{cooper1971definition,
title = {A definition of relevance for information retrieval},
journal = {Information Storage and Retrieval},
volume = {7},
number = {1},
pages = {19-37},
year = {1971},
issn = {0020-0271},
doi = {https://doi.org/10.1016/0020-0271(71)90024-6},
url = {https://www.sciencedirect.com/science/article/pii/0020027171900246},
author = {W.S. Cooper}
}

@misc{zhang2024iterative,
      title={An Iterative Utility Judgment Framework Inspired by Philosophical Relevance via LLMs}, 
      author={Hengran Zhang and Keping Bi and Jiafeng Guo and Xueqi Cheng},
      year={2026},
      eprint={2406.11290},
      archivePrefix={arXiv},
      primaryClass={cs.IR},
      url={https://arxiv.org/abs/2406.11290}, 
}

@misc{jin2025search,
      title={Search-R1: Training LLMs to Reason and Leverage Search Engines with Reinforcement Learning}, 
      author={Bowen Jin and Hansi Zeng and Zhenrui Yue and Jinsung Yoon and Sercan Arik and Dong Wang and Hamed Zamani and Jiawei Han},
      year={2025},
      eprint={2503.09516},
      archivePrefix={arXiv},
      primaryClass={cs.CL},
      url={https://arxiv.org/abs/2503.09516}, 
}

@misc{li2024dmqr,
      title={DMQR-RAG: Diverse Multi-Query Rewriting for RAG}, 
      author={Zhicong Li and Jiahao Wang and Zhishu Jiang and Hangyu Mao and Zhongxia Chen and Jiazhen Du and Yuanxing Zhang and Fuzheng Zhang and Di Zhang and Yong Liu},
      year={2024},
      eprint={2411.13154},
      archivePrefix={arXiv},
      primaryClass={cs.IR},
      url={https://arxiv.org/abs/2411.13154}, 
}

@misc{chan2024rq,
      title={RQ-RAG: Learning to Refine Queries for Retrieval Augmented Generation}, 
      author={Chi-Min Chan and Chunpu Xu and Ruibin Yuan and Hongyin Luo and Wei Xue and Yike Guo and Jie Fu},
      year={2024},
      eprint={2404.00610},
      archivePrefix={arXiv},
      primaryClass={cs.CL},
      url={https://arxiv.org/abs/2404.00610}, 
}

@inproceedings{ye2025q,
    title = "{Q}-{PRM}: Adaptive Query Rewriting for Retrieval-Augmented Generation via Step-level Process Supervision",
    author = "Ye, Xiaopeng  and
      Xu, Chen  and
      Zhang, Chaoliang  and
      Du, Zhaocheng  and
      Xu, Jun  and
      Wang, Gang  and
      Dong, Zhenhua",
    editor = "Christodoulopoulos, Christos  and
      Chakraborty, Tanmoy  and
      Rose, Carolyn  and
      Peng, Violet",
    booktitle = "Findings of the {A}ssociation for {C}omputational {L}inguistics: {EMNLP} 2025",
    month = nov,
    year = "2025",
    address = "Suzhou, China",
    publisher = "Association for Computational Linguistics",
    url = "https://aclanthology.org/2025.findings-emnlp.817/",
    doi = "10.18653/v1/2025.findings-emnlp.817",
    pages = "15113--15128",
    ISBN = "979-8-89176-335-7"
}

@article{wang2025maferw, title={MaFeRw: Query Rewriting with Multi-Aspect Feedbacks for Retrieval-Augmented Large Language Models}, volume={39}, url={https://ojs.aaai.org/index.php/AAAI/article/view/34732}, DOI={10.1609/aaai.v39i24.34732}, abstractNote={In a real-world RAG system, the current query often involves spoken ellipses and ambiguous references from dialogue contexts, necessitating query rewriting to better describe user’s information needs. However, traditional context-based rewriting has minimal enhancement on downstream generation tasks due to the lengthy process from query rewriting to response generation. Some researchers try to utilize reinforcement learning with generation feedback to assist the rewriter, but this sparse rewards provide little guidance in most cases, leading to unstable training and generation results.We find that user’s needs are also reflected in the gold documents, retrieved documents and ground-truth. Therefore, by feeding back these multi-aspect dense rewards to query rewriting, more stable and satisfactory responses can be achieved. In this paper, we propose a novel query rewriting method MaFeRw, which improves RAG performance by integrating multi-aspect feedback from both the retrieval process and generated results. Specifically, we first use manual data to train a T5 model for the rewriter initialization. Next, we design three metrics as reinforcement learning feedback: the similarity between the rewritten query and the gold document, the ranking metrics, and ROUGE between the generation and the ground truth. Inspired by RLAIF, we train three kinds of reward models for the above metrics to achieve more efficient training. Finally, we combine the scores of these reward models as feedback, and use PPO algorithm to explore the optimal query rewriting strategy.Experimental results on two conversational RAG datasets demonstrate that MaFeRw achieves superior generation metrics and more stable training compared to baselines.}, number={24}, journal={Proceedings of the AAAI Conference on Artificial Intelligence}, author={Wang, Yujing and Zhang, Hainan and Pang, Liang and Guo, Binghui and Zheng, Hongwei and Zheng, Zhiming}, year={2025}, month={Apr.}, pages={25434–25442} }

@inproceedings{shi2024replug,
    title = "{REPLUG}: Retrieval-Augmented Black-Box Language Models",
    author = "Shi, Weijia  and
      Min, Sewon  and
      Yasunaga, Michihiro  and
      Seo, Minjoon  and
      James, Richard  and
      Lewis, Mike  and
      Zettlemoyer, Luke  and
      Yih, Wen-tau",
    editor = "Duh, Kevin  and
      Gomez, Helena  and
      Bethard, Steven",
    booktitle = "Proceedings of the 2024 Conference of the {N}orth {A}merican Chapter of the {A}ssociation for {C}omputational {L}inguistics: Human Language Technologies (Volume 1: Long Papers)",
    month = jun,
    year = "2024",
    address = "Mexico City, Mexico",
    publisher = "Association for Computational Linguistics",
    url = "https://aclanthology.org/2024.naacl-long.463/",
    doi = "10.18653/v1/2024.naacl-long.463",
    pages = "8371--8384"
}

@inproceedings{carbonell1998use,
author = {Carbonell, Jaime and Goldstein, Jade},
title = {The use of MMR, diversity-based reranking for reordering documents and producing summaries},
year = {1998},
isbn = {1581130155},
publisher = {Association for Computing Machinery},
address = {New York, NY, USA},
url = {https://doi.org/10.1145/290941.291025},
doi = {10.1145/290941.291025},
booktitle = {Proceedings of the 21st Annual International ACM SIGIR Conference on Research and Development in Information Retrieval},
pages = {335–336},
numpages = {2},
location = {Melbourne, Australia},
series = {SIGIR '98}
}

@inproceedings{cormack2009reciprocal,
author = {Cormack, Gordon V. and Clarke, Charles L A and Buettcher, Stefan},
title = {Reciprocal rank fusion outperforms condorcet and individual rank learning methods},
year = {2009},
isbn = {9781605584836},
publisher = {Association for Computing Machinery},
address = {New York, NY, USA},
url = {https://doi.org/10.1145/1571941.1572114},
doi = {10.1145/1571941.1572114},
booktitle = {Proceedings of the 32nd International ACM SIGIR Conference on Research and Development in Information Retrieval},
pages = {758–759},
numpages = {2},
location = {Boston, MA, USA},
series = {SIGIR '09}
}

@misc{rackauckas2024rag,
      title={RAG-Fusion: a New Take on Retrieval-Augmented Generation}, 
      author={Zackary Rackauckas},
      year={2024},
      eprint={2402.03367},
      archivePrefix={arXiv},
      primaryClass={cs.IR},
      doi={https://doi.org/10.5121/ijnlc.2024.13103},
      url={https://arxiv.org/abs/2402.03367}, 
}

@misc{yang2025qwen3,
      title={Qwen3 Technical Report}, 
      author={An Yang and Anfeng Li and Baosong Yang and Beichen Zhang and Binyuan Hui and Bo Zheng and Bowen Yu and Chang Gao and Chengen Huang and Chenxu Lv and Chujie Zheng and Dayiheng Liu and Fan Zhou and Fei Huang and Feng Hu and Hao Ge and Haoran Wei and Huan Lin and Jialong Tang and Jian Yang and Jianhong Tu and Jianwei Zhang and Jianxin Yang and Jiaxi Yang and Jing Zhou and Jingren Zhou and Junyang Lin and Kai Dang and Keqin Bao and Kexin Yang and Le Yu and Lianghao Deng and Mei Li and Mingfeng Xue and Mingze Li and Pei Zhang and Peng Wang and Qin Zhu and Rui Men and Ruize Gao and Shixuan Liu and Shuang Luo and Tianhao Li and Tianyi Tang and Wenbiao Yin and Xingzhang Ren and Xinyu Wang and Xinyu Zhang and Xuancheng Ren and Yang Fan and Yang Su and Yichang Zhang and Yinger Zhang and Yu Wan and Yuqiong Liu and Zekun Wang and Zeyu Cui and Zhenru Zhang and Zhipeng Zhou and Zihan Qiu},
      year={2025},
      eprint={2505.09388},
      archivePrefix={arXiv},
      primaryClass={cs.CL},
      url={https://arxiv.org/abs/2505.09388}, 
}

@misc{grattafiori2024llama,
      title={The Llama 3 Herd of Models}, 
      author={Aaron Grattafiori and Abhimanyu Dubey and Abhinav Jauhri and Abhinav Pandey and Abhishek Kadian and Ahmad Al-Dahle and Aiesha Letman and Akhil Mathur and Alan Schelten and Alex Vaughan and Amy Yang and Angela Fan and Anirudh Goyal and Anthony Hartshorn and Aobo Yang and Archi Mitra and Archie Sravankumar and Artem Korenev and Arthur Hinsvark and Arun Rao and Aston Zhang and Aurelien Rodriguez and Austen Gregerson and Ava Spataru and Baptiste Roziere and Bethany Biron and Binh Tang and Bobbie Chern and Charlotte Caucheteux and Chaya Nayak and Chloe Bi and Chris Marra and Chris McConnell and Christian Keller and Christophe Touret and Chunyang Wu and Corinne Wong and Cristian Canton Ferrer and Cyrus Nikolaidis and Damien Allonsius and Daniel Song and Danielle Pintz and Danny Livshits and Danny Wyatt and David Esiobu and Dhruv Choudhary and Dhruv Mahajan and Diego Garcia-Olano and Diego Perino and Dieuwke Hupkes and Egor Lakomkin and Ehab AlBadawy and Elina Lobanova and Emily Dinan and Eric Michael Smith and Filip Radenovic and Francisco Guzmán and Frank Zhang and Gabriel Synnaeve and Gabrielle Lee and Georgia Lewis Anderson and Govind Thattai and Graeme Nail and Gregoire Mialon and Guan Pang and Guillem Cucurell and Hailey Nguyen and Hannah Korevaar and Hu Xu and Hugo Touvron and Iliyan Zarov and Imanol Arrieta Ibarra and Isabel Kloumann and Ishan Misra and Ivan Evtimov and Jack Zhang and Jade Copet and Jaewon Lee and Jan Geffert and Jana Vranes and Jason Park and Jay Mahadeokar and Jeet Shah and Jelmer van der Linde and Jennifer Billock and Jenny Hong and Jenya Lee and Jeremy Fu and Jianfeng Chi and Jianyu Huang and Jiawen Liu and Jie Wang and Jiecao Yu and Joanna Bitton and Joe Spisak and Jongsoo Park and Joseph Rocca and Joshua Johnstun and Joshua Saxe and Junteng Jia and Kalyan Vasuden Alwala and Karthik Prasad and Kartikeya Upasani and Kate Plawiak and Ke Li and Kenneth Heafield and Kevin Stone and Khalid El-Arini and Krithika Iyer and Kshitiz Malik and Kuenley Chiu and Kunal Bhalla and Kushal Lakhotia and Lauren Rantala-Yeary and Laurens van der Maaten and Lawrence Chen and Liang Tan and Liz Jenkins and Louis Martin and Lovish Madaan and Lubo Malo and Lukas Blecher and Lukas Landzaat and Luke de Oliveira and Madeline Muzzi and Mahesh Pasupuleti and Mannat Singh and Manohar Paluri and Marcin Kardas and Maria Tsimpoukelli and Mathew Oldham and Mathieu Rita and Maya Pavlova and Melanie Kambadur and Mike Lewis and Min Si and Mitesh Kumar Singh and Mona Hassan and Naman Goyal and Narjes Torabi and Nikolay Bashlykov and Nikolay Bogoychev and Niladri Chatterji and Ning Zhang and Olivier Duchenne and Onur Çelebi and Patrick Alrassy and Pengchuan Zhang and Pengwei Li and Petar Vasic and Peter Weng and Prajjwal Bhargava and Pratik Dubal and Praveen Krishnan and Punit Singh Koura and Puxin Xu and Qing He and Qingxiao Dong and Ragavan Srinivasan and Raj Ganapathy and Ramon Calderer and Ricardo Silveira Cabral and Robert Stojnic and Roberta Raileanu and Rohan Maheswari and Rohit Girdhar and Rohit Patel and Romain Sauvestre and Ronnie Polidoro and Roshan Sumbaly and Ross Taylor and Ruan Silva and Rui Hou and Rui Wang and Saghar Hosseini and Sahana Chennabasappa and Sanjay Singh and Sean Bell and Seohyun Sonia Kim and Sergey Edunov and Shaoliang Nie and Sharan Narang and Sharath Raparthy and Sheng Shen and Shengye Wan and Shruti Bhosale and Shun Zhang and Simon Vandenhende and Soumya Batra and Spencer Whitman and Sten Sootla and Stephane Collot and Suchin Gururangan and Sydney Borodinsky and Tamar Herman and Tara Fowler and Tarek Sheasha and Thomas Georgiou and Thomas Scialom and Tobias Speckbacher and Todor Mihaylov and Tong Xiao and Ujjwal Karn and Vedanuj Goswami and Vibhor Gupta and Vignesh Ramanathan and Viktor Kerkez and Vincent Gonguet and Virginie Do and Vish Vogeti and Vítor Albiero and Vladan Petrovic and Weiwei Chu and Wenhan Xiong and Wenyin Fu and Whitney Meers and Xavier Martinet and Xiaodong Wang and Xiaofang Wang and Xiaoqing Ellen Tan and Xide Xia and Xinfeng Xie and Xuchao Jia and Xuewei Wang and Yaelle Goldschlag and Yashesh Gaur and Yasmine Babaei and Yi Wen and Yiwen Song and Yuchen Zhang and Yue Li and Yuning Mao and Zacharie Delpierre Coudert and Zheng Yan and Zhengxing Chen and Zoe Papakipos and Aaditya Singh and Aayushi Srivastava and Abha Jain and Adam Kelsey and Adam Shajnfeld and Adithya Gangidi and Adolfo Victoria and Ahuva Goldstand and Ajay Menon and Ajay Sharma and Alex Boesenberg and Alexei Baevski and Allie Feinstein and Amanda Kallet and Amit Sangani and Amos Teo and Anam Yunus and Andrei Lupu and Andres Alvarado and Andrew Caples and Andrew Gu and Andrew Ho and Andrew Poulton and Andrew Ryan and Ankit Ramchandani and Annie Dong and Annie Franco and Anuj Goyal and Aparajita Saraf and Arkabandhu Chowdhury and Ashley Gabriel and Ashwin Bharambe and Assaf Eisenman and Azadeh Yazdan and Beau James and Ben Maurer and Benjamin Leonhardi and Bernie Huang and Beth Loyd and Beto De Paola and Bhargavi Paranjape and Bing Liu and Bo Wu and Boyu Ni and Braden Hancock and Bram Wasti and Brandon Spence and Brani Stojkovic and Brian Gamido and Britt Montalvo and Carl Parker and Carly Burton and Catalina Mejia and Ce Liu and Changhan Wang and Changkyu Kim and Chao Zhou and Chester Hu and Ching-Hsiang Chu and Chris Cai and Chris Tindal and Christoph Feichtenhofer and Cynthia Gao and Damon Civin and Dana Beaty and Daniel Kreymer and Daniel Li and David Adkins and David Xu and Davide Testuggine and Delia David and Devi Parikh and Diana Liskovich and Didem Foss and Dingkang Wang and Duc Le and Dustin Holland and Edward Dowling and Eissa Jamil and Elaine Montgomery and Eleonora Presani and Emily Hahn and Emily Wood and Eric-Tuan Le and Erik Brinkman and Esteban Arcaute and Evan Dunbar and Evan Smothers and Fei Sun and Felix Kreuk and Feng Tian and Filippos Kokkinos and Firat Ozgenel and Francesco Caggioni and Frank Kanayet and Frank Seide and Gabriela Medina Florez and Gabriella Schwarz and Gada Badeer and Georgia Swee and Gil Halpern and Grant Herman and Grigory Sizov and Guangyi and Zhang and Guna Lakshminarayanan and Hakan Inan and Hamid Shojanazeri and Han Zou and Hannah Wang and Hanwen Zha and Haroun Habeeb and Harrison Rudolph and Helen Suk and Henry Aspegren and Hunter Goldman and Hongyuan Zhan and Ibrahim Damlaj and Igor Molybog and Igor Tufanov and Ilias Leontiadis and Irina-Elena Veliche and Itai Gat and Jake Weissman and James Geboski and James Kohli and Janice Lam and Japhet Asher and Jean-Baptiste Gaya and Jeff Marcus and Jeff Tang and Jennifer Chan and Jenny Zhen and Jeremy Reizenstein and Jeremy Teboul and Jessica Zhong and Jian Jin and Jingyi Yang and Joe Cummings and Jon Carvill and Jon Shepard and Jonathan McPhie and Jonathan Torres and Josh Ginsburg and Junjie Wang and Kai Wu and Kam Hou U and Karan Saxena and Kartikay Khandelwal and Katayoun Zand and Kathy Matosich and Kaushik Veeraraghavan and Kelly Michelena and Keqian Li and Kiran Jagadeesh and Kun Huang and Kunal Chawla and Kyle Huang and Lailin Chen and Lakshya Garg and Lavender A and Leandro Silva and Lee Bell and Lei Zhang and Liangpeng Guo and Licheng Yu and Liron Moshkovich and Luca Wehrstedt and Madian Khabsa and Manav Avalani and Manish Bhatt and Martynas Mankus and Matan Hasson and Matthew Lennie and Matthias Reso and Maxim Groshev and Maxim Naumov and Maya Lathi and Meghan Keneally and Miao Liu and Michael L. Seltzer and Michal Valko and Michelle Restrepo and Mihir Patel and Mik Vyatskov and Mikayel Samvelyan and Mike Clark and Mike Macey and Mike Wang and Miquel Jubert Hermoso and Mo Metanat and Mohammad Rastegari and Munish Bansal and Nandhini Santhanam and Natascha Parks and Natasha White and Navyata Bawa and Nayan Singhal and Nick Egebo and Nicolas Usunier and Nikhil Mehta and Nikolay Pavlovich Laptev and Ning Dong and Norman Cheng and Oleg Chernoguz and Olivia Hart and Omkar Salpekar and Ozlem Kalinli and Parkin Kent and Parth Parekh and Paul Saab and Pavan Balaji and Pedro Rittner and Philip Bontrager and Pierre Roux and Piotr Dollar and Polina Zvyagina and Prashant Ratanchandani and Pritish Yuvraj and Qian Liang and Rachad Alao and Rachel Rodriguez and Rafi Ayub and Raghotham Murthy and Raghu Nayani and Rahul Mitra and Rangaprabhu Parthasarathy and Raymond Li and Rebekkah Hogan and Robin Battey and Rocky Wang and Russ Howes and Ruty Rinott and Sachin Mehta and Sachin Siby and Sai Jayesh Bondu and Samyak Datta and Sara Chugh and Sara Hunt and Sargun Dhillon and Sasha Sidorov and Satadru Pan and Saurabh Mahajan and Saurabh Verma and Seiji Yamamoto and Sharadh Ramaswamy and Shaun Lindsay and Shaun Lindsay and Sheng Feng and Shenghao Lin and Shengxin Cindy Zha and Shishir Patil and Shiva Shankar and Shuqiang Zhang and Shuqiang Zhang and Sinong Wang and Sneha Agarwal and Soji Sajuyigbe and Soumith Chintala and Stephanie Max and Stephen Chen and Steve Kehoe and Steve Satterfield and Sudarshan Govindaprasad and Sumit Gupta and Summer Deng and Sungmin Cho and Sunny Virk and Suraj Subramanian and Sy Choudhury and Sydney Goldman and Tal Remez and Tamar Glaser and Tamara Best and Thilo Koehler and Thomas Robinson and Tianhe Li and Tianjun Zhang and Tim Matthews and Timothy Chou and Tzook Shaked and Varun Vontimitta and Victoria Ajayi and Victoria Montanez and Vijai Mohan and Vinay Satish Kumar and Vishal Mangla and Vlad Ionescu and Vlad Poenaru and Vlad Tiberiu Mihailescu and Vladimir Ivanov and Wei Li and Wenchen Wang and Wenwen Jiang and Wes Bouaziz and Will Constable and Xiaocheng Tang and Xiaojian Wu and Xiaolan Wang and Xilun Wu and Xinbo Gao and Yaniv Kleinman and Yanjun Chen and Ye Hu and Ye Jia and Ye Qi and Yenda Li and Yilin Zhang and Ying Zhang and Yossi Adi and Youngjin Nam and Yu and Wang and Yu Zhao and Yuchen Hao and Yundi Qian and Yunlu Li and Yuzi He and Zach Rait and Zachary DeVito and Zef Rosnbrick and Zhaoduo Wen and Zhenyu Yang and Zhiwei Zhao and Zhiyu Ma},
      year={2024},
      eprint={2407.21783},
      archivePrefix={arXiv},
      primaryClass={cs.AI},
      url={https://arxiv.org/abs/2407.21783}, 
}

@inproceedings{kwon2023vllm,
author = {Kwon, Woosuk and Li, Zhuohan and Zhuang, Siyuan and Sheng, Ying and Zheng, Lianmin and Yu, Cody Hao and Gonzalez, Joseph and Zhang, Hao and Stoica, Ion},
title = {Efficient Memory Management for Large Language Model Serving with PagedAttention},
year = {2023},
isbn = {9798400702297},
publisher = {Association for Computing Machinery},
address = {New York, NY, USA},
url = {https://doi.org/10.1145/3600006.3613165},
doi = {10.1145/3600006.3613165},
booktitle = {Proceedings of the 29th Symposium on Operating Systems Principles},
pages = {611–626},
numpages = {16},
location = {Koblenz, Germany},
series = {SOSP '23}
}

@inproceedings{wang2024mmlupro,
 author = {Wang, Yubo and Ma, Xueguang and Zhang, Ge and Ni, Yuansheng and Chandra, Abhranil and Guo, Shiguang and Ren, Weiming and Arulraj, Aaran and He, Xuan and Jiang, Ziyan and Li, Tianle and Ku, Max and Wang, Kai and Zhuang, Alex and Fan, Rongqi and Yue, Xiang and Chen, Wenhu},
 booktitle = {Advances in Neural Information Processing Systems},
 doi = {10.52202/079017-3018},
 editor = {A. Globerson and L. Mackey and D. Belgrave and A. Fan and U. Paquet and J. Tomczak and C. Zhang},
 pages = {95266--95290},
 publisher = {Curran Associates, Inc.},
 title = {MMLU-Pro: A More Robust and Challenging Multi-Task Language Understanding Benchmark},
 url = {https://proceedings.neurips.cc/paper_files/paper/2024/file/ad236edc564f3e3156e1b2feafb99a24-Paper-Datasets_and_Benchmarks_Track.pdf},
 volume = {37},
 year = {2024}
}

@inproceedings{penedo2024fineweb,
 author = {Penedo, Guilherme and Kydl\'{\i}\v{c}ek, Hynek and allal, Loubna Ben and Lozhkov, Anton and Mitchell, Margaret and Raffel, Colin and Von Werra, Leandro and Wolf, Thomas},
 booktitle = {Advances in Neural Information Processing Systems},
 doi = {10.52202/079017-0970},
 editor = {A. Globerson and L. Mackey and D. Belgrave and A. Fan and U. Paquet and J. Tomczak and C. Zhang},
 pages = {30811--30849},
 publisher = {Curran Associates, Inc.},
 title = {The FineWeb Datasets: Decanting the Web for the Finest Text Data at Scale},
 url = {https://proceedings.neurips.cc/paper_files/paper/2024/file/370df50ccfdf8bde18f8f9c2d9151bda-Paper-Datasets_and_Benchmarks_Track.pdf},
 volume = {37},
 year = {2024}
}

@inproceedings{zheng2024large,
 author = {Zheng, Chujie and Zhou, Hao and Meng, Fandong and Zhou, Jie and Huang, Minlie},
 booktitle = {International Conference on Learning Representations},
 editor = {B. Kim and Y. Yue and S. Chaudhuri and K. Fragkiadaki and M. Khan and Y. Sun},
 pages = {19426--19454},
 title = {Large Language Models Are Not Robust Multiple Choice Selectors},
 url = {https://proceedings.iclr.cc/paper_files/paper/2024/file/54dd9e0cff6d9214e20d97eb2a3bae49-Paper-Conference.pdf},
 volume = {2024},
 year = {2024}
}

@misc{qwen2026qwen35,
    title = {Qwen3.5: Towards Native Multimodal Agents},
    url = {https://qwen.ai/blog?id=qwen3.5},
    author = {{Qwen Team}},
    month = {February},
    year = {2026}
}

\clearpage
\appendix
\renewcommand{\dbltopfraction}{0.9}
\renewcommand{\dblfloatpagefraction}{0.7}
\setcounter{dbltopnumber}{4}

\section*{Appendix}
\label{sec:appendix}

The appendix is organized as follows. Appendix~\ref{app:method_comparison} compares HYPER with related retrieval methods. Appendix~\ref{app:experimental_details} gives implementation details and the cost accounting. Appendix~\ref{app:stats} reports all significance tests. Appendix~\ref{app:utility} gives per-generator context-utility results and the validation of the judge. Appendix~\ref{app:robustness} collects the controls and robustness analyses. Appendix~\ref{app:cost} reports runtime, and Appendix~\ref{app:mmlupro} gives MMLU-Pro details. Appendix~\ref{app:examples} analyzes the provisional answer hypothesis and gives two corrected examples and a failed one, and Appendix~\ref{app:prompts} lists all prompts.

\section{Comparison with Related Methods}
\label{app:method_comparison}

Table~\ref{tab:method_comparison} summarizes the methods closest to HYPER by the intermediate signal they use and the way that signal changes retrieval. Query-transformation baselines such as LC-MQR \citep{kuo2025mmlf}, DMQR \citep{li2024dmqr}, and HyDE \citep{gao2023precise} change the wording or the representation of the question, whereas HYPER assigns a separate evidential function to each query. IRCoT \citep{trivedi2023interleaving} and Self-RAG \citep{asai2024self} use an intermediate reasoning or generation state to decide the next retrieval step, and MAIN-RAG and ITEM-A use intermediate judgments to filter a pool that has already been retrieved. HYPER uses its intermediate state before the first retrieval to specify which kinds of evidence would settle the choice.

\begin{table}[H]
\centering
\footnotesize
\setlength{\tabcolsep}{3pt}
\caption{Intermediate signals used by related methods and how they shape retrieval.}
\label{tab:method_comparison}
\begin{tabular}{@{}p{1.45cm}p{2.35cm}p{3.6cm}@{}}
\toprule
Method & Intermediate signal & How the signal shapes retrieval \\
\midrule
LC-MQR & Original question & Paraphrased queries whose results are fused \\
DMQR & Original question & Information-level rewrites, including a pseudo-answer \\
HyDE & Original question & Hypothetical answer documents as the query representation \\
IRCoT & Evolving reasoning chain & Next query formed during interleaved reasoning \\
Self-RAG & Generation state & Decides when to retrieve and critiques retrieved passages \\
MAIN-RAG, ITEM-A & Question, document, and answer judgments & Filters a pool retrieved with the original question \\
\midrule
HYPER & Provisional answer hypothesis, discriminating features, evidence needs & Three role-specific queries written before retrieval \\
\bottomrule
\end{tabular}
\end{table}

\section{Experimental Details}
\label{app:experimental_details}
This section records how HYPER and the baselines are implemented. All LLM components within a method use the same model as the final generator unless stated otherwise.

\subsection{Shared Inference Settings}
\label{app:shared_inference}
All non-judge inference runs with vLLM \citep{kwon2023vllm}. Unless stated otherwise, the context length is 8,192 tokens, the maximum output length is 2,048 tokens, and decoding is greedy. HyDE generates its hypothetical documents at temperature 0.7 to keep them diverse. For the $B_{\max}=30$ budget setting and for MMLU-Pro, the context length is 16,384 tokens. For the thinking-mode and Search-R1 comparisons, the context length and maximum output length are 24,576 and 6,144 tokens.

All retrieval-based methods on MedQA and MMLU-Med use the same MedCPT retriever over the same textbook corpus and the same final-context budget of 15 documents. The comparison is therefore normalized by the evidence shown to the generator, although methods differ in how many intermediate candidates they inspect.

\subsection{Computational Budget Accounting}
\label{app:compute_budget}
Table~\ref{tab:compute_budget} counts the LLM calls and retrieval queries of each method before the shared final answer call. HYPER adds two fixed LLM calls, one for the hypothesis and one for the queries. This is more than single-query retrieval but fewer than MAIN-RAG and ITEM-A, which judge candidate documents with the LLM. Measured runtimes are in Appendix~\ref{app:cost}.

\begin{table}[H]
\centering
\small
\begin{tabular}{lcc}
\toprule
Method & LLM calls & Retrieval queries \\
\midrule
\textsc{CoT} & 0 & 0 \\
\textsc{Simple RAG} & 0 & 1 \\
\midrule
\addlinespace[0.35em]
\multicolumn{3}{@{}l}{\textbf{Post-retrieval selection}} \\
\addlinespace[0.15em]
\textsc{MMR-RAG} & 0 & 1 \\
\textsc{Rerank-RAG} & 0 & 1 \\
\textsc{MAIN-RAG} & 30 & 1 \\
\textsc{ITEM-A} & 4-6 & 1 \\
\midrule
\addlinespace[0.35em]
\multicolumn{3}{@{}l}{\textbf{Query transformation}}\\
\addlinespace[0.15em]
\textsc{RRR} & 1 & 1 \\
\textsc{LC-MQR} & 1 & 3 \\
\textsc{DMQR} & 1 & 5 \\
\textsc{HyDE} & 8 & 1 \\
\midrule
\textsc{HYPER} & 2 & 3 \\
\bottomrule
\end{tabular}
\caption{
Analytical per-question computational budget. Counts exclude the shared
final answer-generation call.
}
\label{tab:compute_budget}
\end{table}

\subsection{No-retrieval CoT and Simple RAG}
\label{app:cot_simple_rag}
The no-retrieval baseline answers directly from the question and options using the CoT prompt in Figure~\ref{fig:prompt_cot}. \textsc{Simple RAG} uses the raw question as a single retrieval query, retrieves the top 15 documents, and passes them directly to the shared final generator prompt in Figure~\ref{fig:prompt_generator_v1}. These two baselines define the lower and upper ends of the minimal retrieval setup used throughout the paper: the former removes retrieval entirely, while the latter introduces retrieval without any additional planning, transformation, reranking, or filtering.

\subsection{HYPER}
\label{app:HYPER_details}

HYPER is implemented as the two-stage pipeline of Section~\ref{sec:method}. The hypothesis formulator (Figure~\ref{fig:prompt_HYPER_hypothesis}) reads the question and answer options and returns a retrieval-planning state with discriminating features, a brief rationale, evidence needs, and the provisional answer hypothesis. The query formulator (Figure~\ref{fig:prompt_HYPER_formulator}) receives the question text and this state and writes exactly three queries, one for each of the \textsc{Support}, \textsc{Distinction}, and \textsc{Feature} roles.

Each of the three queries retrieves the top 5 documents. The retrieved sets are merged, deduplicated, and truncated only as needed to satisfy the shared final-context budget. By default, the provisional answer hypothesis is used only for retrieval planning and is withheld from the final generator. The \textsc{HYPER} $-$ options control removes the answer options from the hypothesis-formulation prompt while leaving the rest of the pipeline unchanged.

\subsection{LC-MQR}
\label{app:rewriting_details}

Our LC-MQR baseline adapts the three-query prompt of \citet{kuo2025mmlf}, prompting the model to produce exactly three paraphrase-style queries from the original question. We retrieve with these three queries, without appending the original question as an additional query, and use HYPER's score-sum fusion. The option-aware control in Figure~\ref{fig:prompt_rewriting_options} additionally exposes the answer options and otherwise follows the LC-MQR setup. In both cases, we retrieve the top 5 documents per query, deduplicate the retrieved sets, rank documents by their summed retrieval scores, and pass the resulting context to the shared final generator prompt. These controls match HYPER in the number of queries and in the final evidence budget and differ only in how the queries are constructed.

\subsection{HyDE}
\label{app:hyde}

HyDE uses the prompt in Figure~\ref{fig:prompt_hyde} to generate hypothetical documents from the input question. In our implementation, we generate 8 hypothetical passages per question. The original query embedding and the 8 hypothetical-passage embeddings are then averaged to form the retrieval representation used by MedCPT. As noted above, HyDE is the only method that samples its intermediate generation step, at temperature 0.7, to keep the synthetic passages diverse.

\subsection{Rerank-RAG}
\label{app:reranking_impl}

\textsc{Rerank-RAG} first retrieves 150 candidate documents with the raw question using MedCPT and then reranks those candidates with the \texttt{BAAI/bge-reranker-v2-m3} cross-encoder \citep{multi2024m3}.\footnote{\url{https://huggingface.co/BAAI/bge-reranker-v2-m3}} Reranking is performed on the question--document pair, with a batch size of 16 and a maximum sequence length of 512 tokens. The top 15 reranked documents are then passed to the shared final generator. This baseline therefore tests whether stronger post-retrieval selection can compensate for starting from a raw-question first pass.

\subsection{MAIN-RAG}
\label{app:main_rag}

\textsc{MAIN-RAG} follows the retrieve-then-filter design described in the main text. It begins with 15 documents retrieved from the raw question. For each candidate document, Agent-1 (Figure~\ref{fig:prompt_mainrag_agent1}) produces a document-conditioned answer choice, and Agent-2 (Figure~\ref{fig:prompt_mainrag_agent2}) judges whether the document contains sufficient information to support that answer. We quantify the Agent-2 judgment using the difference between the next-token log probabilities of \texttt{Yes} and \texttt{No} and retain documents whose scores are at least the mean score of the candidate set. The surviving documents, ordered by descending judge score, constitute the final context shown to the shared generator. The predictor and judge output limits are 32 and 1 tokens, respectively.

\subsection{MMR-RAG}
\label{app:mmr_rag}

\textsc{MMR-RAG} is a post-retrieval diversity selection baseline based on
maximal marginal relevance \citep{carbonell1998use}. We first retrieve
150 candidate documents using the raw question and the shared MedCPT
retriever. From this pool, we greedily select 15 documents by balancing
similarity to the question and dissimilarity to the documents already
selected. We use inner product similarity over the stored MedCPT embeddings
and set the diversity weight to $\lambda=0.5$. The selected documents are
passed to the final generator in their MMR selection order.

\subsection{RRR}
\label{app:rrr}

\textsc{RRR} follows the training-free query-rewriting setting of
Rewrite-Retrieve-Read \citep{ma2023query}. We use an LLM as a frozen
rewriter that receives the input question and answer options and produces a
single rewritten search query before retrieval. We use zero-shot prompting
for this step to avoid adding task-specific demonstrations. The rewritten
query is then used to retrieve the top 15 documents with MedCPT, and the
retrieved documents are passed to the shared final generator. Thus, this
baseline differs from \textsc{Simple RAG} only in replacing the raw question
with a single rewritten query. The exact prompt used for query rewriting is
provided in Figure~\ref{fig:prompt_rrr}.

\subsection{DMQR}
\label{app:dmqr}

\textsc{DMQR} is adapted from DMQR-RAG \citep{li2024dmqr} as an
information-based multi-query rewriting baseline. We use a fixed strategy
set rather than adaptive strategy selection. Given the input question, the
rewriter generates up to four rewrites corresponding to general query
rewriting, keyword rewriting, pseudo-answer rewriting, and core-content
extraction. Together with the original question, this yields up to five
retrieval queries after removing duplicate rewrites. The exact prompt used
to generate the strategy-specific rewrites is provided in
Figure~\ref{fig:prompt_dmqr}.

To keep the final-context budget matched to the rest of our experiments, we
allocate a total retrieval budget of 15 documents across the available
queries. This corresponds to retrieving the top 3 documents per query when
all five queries are available. The resulting lists are
fused using the same score-fusion rule as HYPER: duplicate documents are
merged, retrieval scores are summed across lists, and the final context is
ranked by the aggregate score. We use the shared retriever, corpus, and generator interface in place of the original DMQR-RAG web-search retriever, BGE reranking stage, and adaptive strategy selector, so that the comparison isolates the effect of information-diverse query rewriting.

\subsection{ITEM-A}
\label{app:item_a}

\textsc{ITEM-A} is a post-retrieval utility-selection baseline adapted from
the ITEM framework \citep{zhang2024iterative}. We use the answer-in-the-loop
variant without the additional relevance-reranking loop, so that the baseline
tests utility-based filtering rather than reranking. For each question, we
first retrieve the top 15 documents using the raw question. The LLM then
alternates between generating a pseudo-answer from the currently selected
evidence and selecting documents from the original candidate set that are
useful for answering or verifying the question. The prompts used for
pseudo-answer generation and utility selection are provided in
Figures~\ref{fig:prompt_ITEM-A1} and~\ref{fig:prompt_ITEM-A2},
respectively.

We run at most three iterations and stop early when the selected document set stays the same. The final selected documents are passed to the same
final generator prompt used by the other retrieval-based methods. Since
\textsc{ITEM-A} filters the initial raw-question context, its final context
may contain fewer than 15 documents. In our runs, \textsc{ITEM-A} returned an empty selected set for a small fraction of examples, with an average rate of 2.4\% across datasets. These cases are passed to the generator with an empty retrieved context and are counted as \textsc{Not Useful} in the context-utility evaluation.

\subsection{Context-utility judge}
\label{app:judge_details}

The context-utility analysis uses the judge prompt in Figure~\ref{fig:prompt_gpt_judge}. Given the question, options, gold answer, and the retrieved context set produced by a method, the judge first identifies the answer option most strongly supported by the context alone (\texttt{selected\_answer}) and then assigns one of the three labels used in the paper: \textsc{Entailed}, \textsc{Useful-but-Not-Entailed}, or \textsc{Not Useful}. The prompt is intentionally conservative about \textsc{Entailed} and \textsc{Useful-but-Not-Entailed}; background-only passages, weak topical relevance, and contexts that support a competing option are labeled \textsc{Not Useful}. The label definitions and output schema are fixed across methods. In the second-judge corpus, the archived prompts for MMR-RAG, ITEM-A, RRR, and DMQR prefix the context-only rule with ``Closed-world constraint:''; the other methods use the wording in Figure~\ref{fig:prompt_gpt_judge}.

\subsection{Gold-Blind Judge and Second Judge}
\label{app:judge_protocols}
The gold-blind protocol uses the same three labels with the gold answer hidden. The judge first selects the option best supported by the retrieved context alone, or \texttt{UNSURE}, and then labels the context relative to that selection. It is applied to all 1,273 Qwen3-4B MedQA questions for the methods whose retrieval runs exactly match the main experiments. The second judge, Qwen3.5-27B \citep{qwen2026qwen35}, reuses each context's archived gold-aware GPT input for all ten methods in the Qwen3-4B experiments on both datasets. Empty contexts are labeled \textsc{Not Useful} by rule.

\subsection{MMLU-Pro Setup}
\label{app:mmlupro_setup}
For MMLU-Pro we split FineWeb-Edu sample-10BT \citep{penedo2024fineweb} into 28,352,435 chunks of 480 BGE-M3 tokens with a 32-token overlap and search them by exact dense retrieval with BGE-M3. Every method retrieves at most 15 documents, and the generator context is 16,384 tokens. CoT uses the category-specific few-shot prompt of MMLU-Pro, and the retrieval methods add the retrieved documents to the same prompt. For compatibility with the existing pipeline the prompt introduces the documents as Wikipedia documents, although they come from FineWeb-Edu. HYPER uses the same hypothesis and query prompts as in the medical experiments, adapted to ten options.

\section{Statistical Analysis}
\label{app:stats}

Table~\ref{tab:sig_matrix} lists the accuracy difference between HYPER and each baseline for every generator and dataset, together with the exact McNemar test after Benjamini--Hochberg and Holm correction over all 80 comparisons. HYPER is ahead in 78 of the 80 comparisons. The two exceptions are MMR-RAG and RRR with Llama3.2-3B on MMLU-Med, where the differences are 0.4 and 0.1 points. Table~\ref{tab:bootstrap_simple} gives paired bootstrap intervals against Simple RAG, all of which exclude zero. Table~\ref{tab:seeds} repeats the comparison with three primary baselines over five sampled runs of Qwen3-4B. All six paired $t$-tests remain significant after Holm correction.

\begin{table*}[!t]
\centering
\fontsize{8}{8.5}\selectfont
\setlength{\tabcolsep}{3.4pt}
\caption{Accuracy difference HYPER $-$ baseline (percentage points) for every generator and dataset. $^{*}$: significant by exact McNemar test after Benjamini--Hochberg correction over all 80 comparisons ($q<.05$). $^{\dagger}$: also significant after Holm correction ($p<.05$). Significant: 57/80 (BH), 35/80 (Holm).}
\label{tab:sig_matrix}
\begin{tabular}{@{}lcccc cccc@{}}
\toprule
& \multicolumn{4}{c}{MedQA} & \multicolumn{4}{c}{MMLU-Med} \\
\cmidrule(lr){2-5} \cmidrule(l){6-9}
Baseline & Llama3.2 3B & Llama3.1 8B & Qwen3 4B & Qwen3 30B & Llama3.2 3B & Llama3.1 8B & Qwen3 4B & Qwen3 30B \\
\midrule
CoT          & +5.4$^{*\dagger}$ & +3.9$^{*}$ & +5.1$^{*\dagger}$ & +1.6 & +2.3 & +4.7$^{*\dagger}$ & +2.5$^{*}$ & +0.5 \\
Simple RAG   & +7.8$^{*\dagger}$ & +6.2$^{*\dagger}$ & +6.6$^{*\dagger}$ & +3.1$^{*\dagger}$ & +4.1$^{*}$ & +5.0$^{*\dagger}$ & +3.6$^{*\dagger}$ & +1.7 \\
MMR-RAG      & +9.0$^{*\dagger}$ & +6.0$^{*\dagger}$ & +6.1$^{*\dagger}$ & +3.0$^{*}$ & $-$0.4 & +3.9$^{*}$ & +3.2$^{*}$ & +1.3 \\
Rerank-RAG   & +6.6$^{*\dagger}$ & +5.7$^{*\dagger}$ & +5.8$^{*\dagger}$ & +2.5$^{*}$ & +1.5 & +3.9$^{*}$ & +3.1$^{*}$ & +0.6 \\
MAIN-RAG     & +5.3$^{*\dagger}$ & +6.4$^{*\dagger}$ & +6.5$^{*\dagger}$ & +2.9$^{*}$ & +0.8 & +4.6$^{*\dagger}$ & +3.7$^{*\dagger}$ & +1.7 \\
ITEM-A       & +7.9$^{*\dagger}$ & +8.0$^{*\dagger}$ & +8.2$^{*\dagger}$ & +4.0$^{*\dagger}$ & +4.9$^{*}$ & +7.2$^{*\dagger}$ & +3.4$^{*}$ & +1.3 \\
RRR          & +3.3$^{*}$ & +1.0 & +4.3$^{*\dagger}$ & +3.9$^{*\dagger}$ & $-$0.1 & +5.7$^{*\dagger}$ & +1.7 & +1.2 \\
LC-MQR       & +4.5$^{*}$ & +5.5$^{*\dagger}$ & +4.9$^{*\dagger}$ & +2.4$^{*}$ & +1.6 & +5.1$^{*\dagger}$ & +2.9$^{*}$ & +1.8$^{*}$ \\
DMQR         & +5.4$^{*\dagger}$ & +2.7 & +4.6$^{*\dagger}$ & +1.7 & +2.5 & +3.2$^{*}$ & +0.9 & +0.5 \\
HyDE         & +5.5$^{*\dagger}$ & +3.7$^{*}$ & +6.5$^{*\dagger}$ & +2.9$^{*}$ & +0.8 & +3.2$^{*}$ & +2.8$^{*}$ & +1.2 \\
\bottomrule
\end{tabular}
\end{table*}

\begin{table}[H]
\centering
\small
\caption{HYPER $-$ Simple RAG accuracy difference with 95\% paired bootstrap intervals (percentage points, 20,000 resamples).}
\label{tab:bootstrap_simple}
\begin{tabular}{lcc}
\toprule
Generator & MedQA & MMLU-Med \\
\midrule
Llama3.2 3B & +7.8 [4.8, 10.8] & +4.1 [1.2, 7.1] \\
Llama3.1 8B & +6.2 [3.5, 9.0] & +5.0 [2.4, 7.5] \\
Qwen3 4B    & +6.6 [4.3, 8.9]  & +3.6 [1.5, 5.7] \\
Qwen3 30B   & +3.1 [1.3, 4.9]  & +1.7 [0.1, 3.4] \\
\bottomrule
\end{tabular}
\end{table}

\begin{table}[H]
\centering
\fontsize{8}{8.5}\selectfont
\setlength{\tabcolsep}{2.6pt}
\caption{Qwen3-4B at temperature 0.6 over five seeds (mean $\pm$ SD, \%). Two-sided paired $t$-tests over seedwise differences ($df$=4).}
\label{tab:seeds}
\begin{tabular}{llcccc}
\toprule
Data & Baseline & HYPER & Baseline & $\Delta$ & $p$ \\
\midrule
MedQA & CoT        & 76.6 $\pm$ 1.0 & 73.0 $\pm$ 0.7 & +3.7 & .0010 \\
MedQA & Simple RAG & 76.6 $\pm$ 1.0 & 71.8 $\pm$ 0.8 & +4.9 & .0032 \\
MedQA & LC-MQR     & 76.6 $\pm$ 1.0 & 73.2 $\pm$ 0.7 & +3.4 & .0032 \\
MMLU  & CoT        & 84.3 $\pm$ 0.7 & 83.3 $\pm$ 0.5 & +1.1 & .0111 \\
MMLU  & Simple RAG & 84.3 $\pm$ 0.7 & 82.3 $\pm$ 0.6 & +2.1 & .0160 \\
MMLU  & LC-MQR     & 84.3 $\pm$ 0.7 & 82.8 $\pm$ 1.0 & +1.6 & .0042 \\
\bottomrule
\end{tabular}
\end{table}

\section{Context Utility}
\label{app:utility}

\subsection{Per-Generator Results}
Table~\ref{tab:dur-main} in the main text gives DUR for every generator, and Table~\ref{tab:context-label-full} gives the full label distribution. HYPER has the highest DUR in every cell and raises both \textsc{Entailed} and \textsc{Useful} rates. Table~\ref{tab:subset-utility-full} gives accuracy and same-subset gains over CoT by label for every method.

\begin{table*}[!t]
\centering
\fontsize{7.5}{8.2}\selectfont
\setlength{\tabcolsep}{2.8pt}
\caption{Full context-utility breakdown underlying Section~\ref{sec:context-utility}. For each method, we report the rates (\%) of \textsc{Entailed}, \textsc{Useful-but-Not-Entailed} (\textsc{Useful}), and \textsc{Not Useful} contexts across all four model configurations and their average.}
\label{tab:context-label-full}
\begin{tabular}{@{}llccccc ccccc@{}}
\toprule
& & \multicolumn{5}{c}{MedQA} & \multicolumn{5}{c}{MMLU-Med} \\
\cmidrule(lr){3-7} \cmidrule(l){8-12}
Method & Label & {\fontsize{6.8}{7.2}\selectfont Llama3.2 3B} & {\fontsize{6.8}{7.2}\selectfont Llama3.1 8B} & {\fontsize{6.8}{7.2}\selectfont Qwen3 4B} & {\fontsize{6.8}{7.2}\selectfont Qwen3 30B} & Avg. & {\fontsize{6.8}{7.2}\selectfont Llama3.2 3B} & {\fontsize{6.8}{7.2}\selectfont Llama3.1 8B} & {\fontsize{6.8}{7.2}\selectfont Qwen3 4B} & {\fontsize{6.8}{7.2}\selectfont Qwen3 30B} & Avg. \\
\midrule

\multirow{3}{*}{Simple RAG} & Entailed & 6.0 & 6.0 & 6.0 & 6.0 & 6.0 & 22.5 & 22.5 & 22.5 & 22.5 & 22.5 \\
 & Useful & 24.0 & 24.0 & 24.0 & 24.0 & 24.0 & 26.2 & 26.2 & 26.2 & 26.2 & 26.2 \\
 & Not Useful & 70.0 & 70.0 & 70.0 & 70.0 & 70.0 & 51.3 & 51.3 & 51.3 & 51.3 & 51.3 \\
\midrule

\multirow{3}{*}{MMR-RAG} & Entailed & 4.2 & 4.2 & 4.2 & 4.2 & 4.2 & 18.9 & 18.9 & 18.9 & 18.9 & 18.9 \\
 & Useful & 22.2 & 22.2 & 22.2 & 22.2 & 22.2 & 24.2 & 24.2 & 24.2 & 24.2 & 24.2 \\
 & Not Useful & 73.6 & 73.6 & 73.6 & 73.6 & 73.6 & 56.9 & 56.9 & 56.9 & 56.9 & 56.9 \\
\midrule

\multirow{3}{*}{Rerank-RAG} & Entailed & 10.7 & 10.7 & 10.7 & 10.7 & 10.7 & 29.1 & 29.1 & 29.1 & 29.1 & 29.1 \\
 & Useful & 33.4 & 33.4 & 33.4 & 33.4 & 33.4 & 27.7 & 27.7 & 27.7 & 27.7 & 27.7 \\
 & Not Useful & 55.9 & 55.9 & 55.9 & 55.9 & 55.9 & 43.2 & 43.2 & 43.2 & 43.2 & 43.2 \\
\midrule

\multirow{3}{*}{MAIN-RAG} & Entailed & 6.2 & 6.5 & 6.4 & 7.0 & 6.5 & 21.8 & 24.6 & 24.0 & 24.2 & 23.6 \\
 & Useful & 21.2 & 22.6 & 21.2 & 22.7 & 21.9 & 23.0 & 23.0 & 22.3 & 24.5 & 23.2 \\
 & Not Useful & 72.6 & 70.9 & 72.3 & 70.3 & 71.5 & 55.3 & 52.3 & 53.7 & 51.2 & 53.1 \\
\midrule

\multirow{3}{*}{ITEM-A} & Entailed & 6.4 & 7.1 & 7.9 & 7.9 & 7.3 & 20.6 & 22.9 & 24.0 & 24.6 & 23.0 \\
 & Useful & 17.0 & 18.5 & 17.4 & 18.9 & 18.0 & 20.5 & 20.6 & 19.7 & 20.9 & 20.4 \\
 & Not Useful & 76.5 & 74.3 & 74.8 & 73.1 & 74.7 & 59.0 & 56.6 & 56.4 & 54.5 & 56.6 \\
\midrule

\multirow{3}{*}{RRR} & Entailed & 20.6 & 25.5 & 21.9 & 23.9 & 23.0 & 19.9 & 26.4 & 29.2 & 31.8 & 26.8 \\
 & Useful & 41.4 & 43.0 & 43.4 & 43.0 & 42.7 & 28.2 & 29.1 & 30.5 & 31.5 & 29.8 \\
 & Not Useful & 38.0 & 31.6 & 34.6 & 33.2 & 34.3 & 51.9 & 44.5 & 40.3 & 36.7 & 43.4 \\
\midrule

\multirow{3}{*}{LC-MQR} & Entailed & 11.0 & 12.7 & 15.5 & 18.4 & 14.4 & 23.0 & 23.8 & 24.5 & 27.8 & 24.8 \\
 & Useful & 34.2 & 38.5 & 46.7 & 47.2 & 41.7 & 28.4 & 29.9 & 32.2 & 31.3 & 30.5 \\
 & Not Useful & 54.8 & 48.8 & 37.8 & 34.4 & 43.9 & 48.6 & 46.3 & 43.3 & 40.9 & 44.7 \\
\midrule

\multirow{3}{*}{LC-MQR $+$ options} & Entailed & 13.4 & 18.1 & 20.8 & 24.3 & 19.1 & 26.4 & 27.4 & 28.9 & 30.7 & 28.3 \\
 & Useful & 37.8 & 44.9 & 51.0 & 52.1 & 46.4 & 30.5 & 34.0 & 35.9 & 35.6 & 34.0 \\
 & Not Useful & 48.9 & 37.1 & 28.2 & 23.6 & 34.4 & 43.2 & 38.7 & 35.2 & 33.7 & 37.7 \\
\midrule

\multirow{3}{*}{DMQR} & Entailed & 16.0 & 16.7 & 18.2 & 24.7 & 18.9 & 25.4 & 26.1 & 27.3 & 28.7 & 26.9 \\
 & Useful & 36.4 & 41.2 & 44.0 & 47.4 & 42.3 & 26.7 & 30.1 & 29.8 & 30.8 & 29.4 \\
 & Not Useful & 47.6 & 42.0 & 37.8 & 27.9 & 38.8 & 47.8 & 43.8 & 42.9 & 40.6 & 43.8 \\
\midrule

\multirow{3}{*}{HyDE} & Entailed & 13.0 & 14.0 & 13.6 & 14.5 & 13.7 & 18.9 & 19.9 & 19.5 & 19.8 & 19.5 \\
 & Useful & 38.6 & 40.5 & 39.2 & 40.1 & 39.6 & 24.1 & 26.3 & 24.8 & 26.4 & 25.4 \\
 & Not Useful & 48.4 & 45.5 & 47.2 & 45.4 & 46.6 & 57.0 & 53.8 & 55.7 & 53.7 & 55.1 \\
\midrule

\multirow{3}{*}{\textbf{HYPER (ours)}} & Entailed & 29.6 & 32.0 & 35.3 & 41.9 & 34.7 & 31.9 & 34.5 & 35.5 & 36.7 & 34.7 \\
 & Useful & 43.8 & 49.8 & 47.8 & 47.9 & 47.3 & 34.0 & 34.5 & 37.0 & 37.0 & 35.6 \\
 & Not Useful & 26.6 & 18.2 & 16.8 & 10.2 & 17.9 & 34.2 & 30.9 & 27.5 & 26.3 & 29.7 \\
\midrule

\multirow{3}{*}{HYPER $-$ options} & Entailed & 18.5 & 24.9 & 25.1 & 31.6 & 25.0 & 25.2 & 28.5 & 29.1 & 32.4 & 28.8 \\
 & Useful & 42.5 & 45.6 & 45.4 & 48.9 & 45.6 & 31.9 & 32.6 & 35.1 & 35.1 & 33.7 \\
 & Not Useful & 39.0 & 29.5 & 29.5 & 19.5 & 29.4 & 43.0 & 38.9 & 35.8 & 32.5 & 37.6 \\
\midrule

\multirow{3}{*}{HYPER - $q_{\mathrm{sup}}$} & Entailed & 12.5 & 12.6 & 15.5 & 14.0 & 13.6 & 16.4 & 18.6 & 19.7 & 17.9 & 18.1 \\
 & Useful & 42.5 & 49.6 & 51.0 & 56.9 & 50.0 & 32.2 & 33.8 & 35.6 & 37.9 & 34.8 \\
 & Not Useful & 45.0 & 37.9 & 33.5 & 29.1 & 36.4 & 51.5 & 47.7 & 44.8 & 44.2 & 47.0 \\
\midrule

\multirow{3}{*}{HYPER - $q_{\mathrm{dist}}$} & Entailed & 18.2 & 24.0 & 25.2 & 31.7 & 24.8 & 23.2 & 25.5 & 27.4 & 27.2 & 25.8 \\
 & Useful & 42.5 & 47.7 & 44.9 & 48.7 & 46.0 & 30.3 & 31.9 & 34.4 & 36.9 & 33.4 \\
 & Not Useful & 39.2 & 28.3 & 29.9 & 19.6 & 29.2 & 46.5 & 42.6 & 38.2 & 35.8 & 40.8 \\
\midrule

\multirow{3}{*}{HYPER - $q_{\mathrm{feat}}$} & Entailed & 20.0 & 22.9 & 25.1 & 31.7 & 24.9 & 20.1 & 23.5 & 22.5 & 25.1 & 22.8 \\
 & Useful & 43.1 & 50.0 & 49.1 & 50.9 & 48.3 & 31.4 & 33.6 & 38.2 & 37.9 & 35.3 \\
 & Not Useful & 36.9 & 27.0 & 25.8 & 17.4 & 26.8 & 48.4 & 43.0 & 39.2 & 37.0 & 41.9 \\
\bottomrule
\end{tabular}
\end{table*}

\begin{table*}[!t]
\centering
\fontsize{7.5}{8.2}\selectfont
\setlength{\tabcolsep}{1.25pt}
\caption{Full model-by-model values underlying Table~\ref{tab:subset-utility}. Each cell is reported as \emph{accuracy / $\Delta_{\mathrm{CoT}}$}.}
\label{tab:subset-utility-full}
\begin{tabular}{@{}llccccc ccccc@{}}
\toprule
& & \multicolumn{5}{c}{MedQA} & \multicolumn{5}{c}{MMLU-Med} \\
\cmidrule(lr){3-7} \cmidrule(l){8-12}
Method & Label & {\fontsize{6.8}{7.2}\selectfont Llama3.2 3B} & {\fontsize{6.8}{7.2}\selectfont Llama3.1 8B} & {\fontsize{6.8}{7.2}\selectfont Qwen3 4B} & {\fontsize{6.8}{7.2}\selectfont Qwen3 30B} & Avg. & {\fontsize{6.8}{7.2}\selectfont Llama3.2 3B} & {\fontsize{6.8}{7.2}\selectfont Llama3.1 8B} & {\fontsize{6.8}{7.2}\selectfont Qwen3 4B} & {\fontsize{6.8}{7.2}\selectfont Qwen3 30B} & Avg. \\
\midrule

\multirow{3}{*}{Simple RAG} & Entailed & 81.8 / +13.0 & 85.7 / +2.6 & 93.5 / +9.1 & 96.1 / -1.3 & 89.3 / +5.8 & 84.9 / +4.5 & 93.9 / +6.5 & 97.1 / +6.1 & 98.4 / +0.4 & 93.6 / +4.4 \\
 & Useful & 63.3 / +3.3 & 79.7 / +3.3 & 85.6 / +5.9 & 91.8 / +3.9 & 80.1 / +4.1 & 67.7 / +0.7 & 80.4 / +1.8 & 88.1 / +0.7 & 90.2 / +0.0 & 81.6 / +0.8 \\
 & Not Useful & 50.7 / -5.6 & 60.7 / -4.6 & 64.5 / -4.9 & 78.5 / -3.3 & 63.6 / -4.6 & 53.8 / -5.9 & 63.5 / -4.3 & 71.9 / -5.2 & 84.1 / -2.7 & 68.3 / -4.5 \\
\midrule

\multirow{3}{*}{MMR-RAG} & Entailed & 87.0 / +9.3 & 90.7 / +1.9 & 98.1 / +13.0 & 100.0 / +1.9 & 94.0 / +6.5 & 87.4 / +6.3 & 92.7 / +6.3 & 95.1 / +4.9 & 98.1 / +1.0 & 93.3 / +4.6 \\
 & Useful & 61.0 / -4.3 & 79.1 / +0.4 & 84.8 / +5.0 & 91.8 / +0.4 & 79.2 / +0.4 & 77.9 / +9.9 & 81.0 / +1.9 & 86.3 / -1.1 & 92.4 / +1.9 & 84.4 / +3.1 \\
 & Not Useful & 50.5 / -4.2 & 62.0 / -3.0 & 66.4 / -3.6 & 79.1 / -2.0 & 64.5 / -3.2 & 59.0 / -1.6 & 68.1 / -1.5 & 76.1 / -2.4 & 85.2 / -2.6 & 72.1 / -2.0 \\
\midrule

\multirow{3}{*}{Rerank-RAG} & Entailed & 81.6 / +8.1 & 92.6 / +4.4 & 97.8 / +8.1 & 99.3 / -0.7 & 92.8 / +5.0 & 86.1 / +6.3 & 93.4 / +7.9 & 95.6 / +6.0 & 98.1 / +1.9 & 93.3 / +5.5 \\
 & Useful & 64.0 / +5.6 & 77.9 / +5.9 & 81.6 / +5.2 & 89.6 / +2.8 & 78.3 / +4.9 & 66.9 / +2.3 & 76.5 / +0.3 & 82.1 / -2.0 & 90.4 / +2.3 & 79.0 / +0.7 \\
 & Not Useful & 47.8 / -7.0 & 56.2 / -7.4 & 61.5 / -5.9 & 76.4 / -3.1 & 60.5 / -5.9 & 54.5 / -3.8 & 63.4 / -3.8 & 73.4 / -4.3 & 84.5 / -3.0 & 68.9 / -3.7 \\
\midrule

\multirow{3}{*}{MAIN-RAG} & Entailed & 87.3 / +11.4 & 91.6 / +6.0 & 97.6 / +8.5 & 97.8 / -1.1 & 93.6 / +6.2 & 90.3 / +8.4 & 94.4 / +7.8 & 97.3 / +5.7 & 98.5 / +0.4 & 95.1 / +5.6 \\
 & Useful & 73.3 / +12.6 & 79.2 / +3.8 & 88.9 / +8.9 & 90.7 / +3.1 & 83.0 / +7.1 & 74.4 / +5.6 & 77.3 / -0.4 & 87.7 / -0.4 & 90.6 / +1.9 & 82.5 / +1.7 \\
 & Not Useful & 51.1 / -4.5 & 60.2 / -5.3 & 64.0 / -5.3 & 78.9 / -2.7 & 63.5 / -4.5 & 56.1 / -3.0 & 65.1 / -3.3 & 72.3 / -4.6 & 83.7 / -3.4 & 69.3 / -3.6 \\
\midrule

\multirow{3}{*}{ITEM-A} & Entailed & 84.1 / +13.4 & 94.5 / +11.0 & 98.0 / +13.0 & 99.0 / +2.0 & 93.9 / +9.8 & 87.1 / +7.6 & 93.2 / +6.4 & 96.6 / +5.7 & 99.6 / +2.6 & 94.1 / +5.6 \\
 & Useful & 68.2 / +9.2 & 80.5 / +4.2 & 87.8 / +7.7 & 93.8 / +3.3 & 82.6 / +6.1 & 67.7 / -1.3 & 80.8 / +1.8 & 87.9 / +1.9 & 92.5 / +2.2 & 82.2 / +1.1 \\
 & Not Useful & 50.3 / -6.4 & 58.2 / -7.6 & 62.6 / -7.2 & 76.8 / -4.3 & 62.0 / -6.4 & 54.2 / -6.5 & 61.2 / -7.6 & 73.8 / -4.7 & 83.5 / -3.5 & 68.2 / -5.6 \\
\midrule

\multirow{3}{*}{RRR} & Entailed & 82.1 / +7.6 & 93.5 / +7.4 & 94.3 / +6.5 & 97.7 / +2.3 & 91.9 / +5.9 & 85.3 / +7.8 & 93.4 / +6.6 & 95.3 / +7.2 & 96.8 / +0.3 & 92.7 / +5.5 \\
 & Useful & 64.5 / +6.3 & 73.5 / +5.5 & 77.4 / +4.3 & 83.4 / -0.2 & 74.7 / +4.0 & 73.6 / +5.5 & 78.2 / +3.2 & 86.1 / +0.6 & 90.1 / +0.3 & 82.0 / +2.4 \\
 & Not Useful & 43.4 / -5.4 & 52.5 / -4.2 & 55.8 / -7.3 & 68.5 / -8.3 & 55.0 / -6.3 & 59.6 / -1.4 & 59.8 / -8.2 & 73.6 / -3.6 & 82.5 / -2.5 & 68.9 / -4.0 \\
\midrule

\multirow{3}{*}{LC-MQR} & Entailed & 86.4 / +13.6 & 94.4 / +11.7 & 98.5 / +8.6 & 97.4 / +0.4 & 94.2 / +8.6 & 84.9 / +9.6 & 91.5 / +6.9 & 95.5 / +6.0 & 98.7 / +1.7 & 92.6 / +6.0 \\
 & Useful & 66.5 / +8.9 & 74.5 / +2.2 & 78.5 / +4.2 & 88.0 / +0.7 & 76.9 / +4.0 & 70.6 / +1.6 & 77.6 / +1.5 & 88.0 / +0.0 & 90.0 / +0.6 & 81.6 / +0.9 \\
 & Not Useful & 48.6 / -6.6 & 54.9 / -8.1 & 55.9 / -8.1 & 69.4 / -3.4 & 57.2 / -6.5 & 56.5 / -4.0 & 64.1 / -5.4 & 70.9 / -4.5 & 81.1 / -4.9 & 68.2 / -4.7 \\
\midrule

\multirow{3}{*}{DMQR} & Entailed & 79.4 / +8.3 & 93.9 / +7.0 & 97.0 / +6.5 & 98.4 / +1.3 & 92.2 / +5.8 & 83.8 / +5.1 & 93.7 / +7.4 & 96.0 / +5.7 & 97.1 / +1.0 & 92.6 / +4.8 \\
 & Useful & 63.9 / +4.8 & 80.6 / +7.4 & 80.0 / +5.4 & 87.6 / +0.3 & 78.0 / +4.5 & 71.8 / +4.5 & 79.3 / +3.4 & 87.7 / +2.2 & 91.9 / +0.6 & 82.7 / +2.6 \\
 & Not Useful & 46.2 / -6.4 & 50.8 / -7.1 & 54.3 / -7.9 & 65.4 / -2.0 & 54.2 / -5.9 & 53.6 / -5.6 & 64.4 / -3.4 & 74.9 / -1.5 & 83.9 / -1.1 & 69.2 / -2.9 \\
\midrule

\multirow{3}{*}{HyDE} & Entailed & 81.8 / +6.7 & 93.8 / +10.7 & 94.8 / +4.6 & 99.5 / +1.6 & 92.5 / +5.9 & 91.7 / +9.7 & 94.5 / +4.6 & 98.1 / +2.8 & 98.6 / -0.5 & 95.7 / +4.2 \\
 & Useful & 60.6 / +4.9 & 72.3 / +1.7 & 76.2 / +3.4 & 83.8 / +0.0 & 73.2 / +2.5 & 71.0 / +3.8 & 81.8 / +2.1 & 88.5 / +1.9 & 95.1 / +2.8 & 84.1 / +2.6 \\
 & Not Useful & 49.4 / -5.8 & 59.1 / -4.3 & 60.7 / -7.2 & 76.8 / -3.3 & 61.5 / -5.2 & 58.5 / -2.3 & 67.2 / +0.0 & 74.5 / -2.5 & 83.2 / -2.6 & 70.9 / -1.8 \\
\midrule

\multirow{3}{*}{\textbf{HYPER (ours)}} & Entailed & 89.1 / +14.3 & 94.3 / +8.8 & 97.1 / +6.9 & 98.5 / +3.2 & 94.8 / +8.3 & 90.2 / +8.1 & 92.3 / +5.1 & 98.2 / +3.4 & 97.5 / +1.2 & 94.5 / +4.4 \\
 & Useful & 64.5 / +6.3 & 74.6 / +6.8 & 77.3 / +7.4 & 84.9 / +2.8 & 75.3 / +5.8 & 73.0 / +4.6 & 81.9 / +7.7 & 84.6 / +3.7 & 91.6 / +1.0 & 82.8 / +4.3 \\
 & Not Useful & 32.8 / -5.9 & 31.0 / -12.5 & 39.3 / -5.1 & 37.7 / -10.0 & 35.2 / -8.4 & 44.1 / -5.4 & 63.2 / +0.9 & 69.9 / -0.3 & 79.7 / -1.4 & 64.2 / -1.6 \\
\bottomrule
\vspace{1cm}
\end{tabular}
\end{table*}

\subsection{Judge Validation}
\label{app:judge}
\paragraph{Human annotation.} Two of the authors independently labeled 60 Qwen3-4B MedQA contexts, 30 from HYPER and 30 from Simple RAG, using the label definitions of the judge prompt (Figure~\ref{fig:prompt_gpt_judge}). Table~\ref{tab:human_agreement} gives agreement on the exact three-way label and on the binary DUR distinction, which merges \textsc{Entailed} and \textsc{Useful}. Agreement on DUR is high both between the annotators and between each annotator and GPT-4.1. Most three-way disagreements concern the boundary between \textsc{Entailed} and \textsc{Useful}. Table~\ref{tab:human_dur} shows that both annotators reproduce the DUR gap between HYPER and Simple RAG.

\begin{table}[H]
\centering
\small
\setlength{\tabcolsep}{3.2pt}
\caption{Agreement on 60 Qwen3-4B MedQA contexts. Agreement in \%, with Cohen's $\kappa$.}
\label{tab:human_agreement}
\begin{tabular}{lcccc}
\toprule
& \multicolumn{2}{c}{Three-way} & \multicolumn{2}{c}{DUR} \\
\cmidrule(lr){2-3} \cmidrule(l){4-5}
Comparison & Agr. & $\kappa$ & Agr. & $\kappa$ \\
\midrule
Human 1 vs.\ Human 2  & 88.3 & .805 & 96.7 & .933 \\
Human 1 vs.\ GPT-4.1  & 63.3 & .432 & 86.7 & .734 \\
Human 2 vs.\ GPT-4.1  & 66.7 & .481 & 86.7 & .733 \\
\bottomrule
\end{tabular}
\end{table}

\begin{table}[H]
\centering
\small
\caption{DUR (\%) on the 60 human-validation contexts.}
\label{tab:human_dur}
\begin{tabular}{lccc}
\toprule
Labeler & HYPER & Simple RAG & Gap \\
\midrule
Human 1 & 76.7 & 20.0 & +56.7 \\
Human 2 & 76.7 & 26.7 & +50.0 \\
GPT-4.1 & 76.7 & 26.7 & +50.0 \\
\bottomrule
\end{tabular}
\end{table}

\paragraph{Second judge.} We compare Qwen3.5-27B and GPT-4.1 labels for 23,620 contexts from all ten methods in the Qwen3-4B experiments on both datasets, including 83 empty contexts labeled \textsc{Not Useful} by rule. Its agreement with GPT-4.1 is 84.5\% on the three-way label ($\kappa$=.749) and 91.9\% on DUR ($\kappa$=.838). Table~\ref{tab:qwen_dur} shows that the ranking of methods is preserved.

\begin{table}[H]
\centering
\small
\caption{Qwen3-4B DUR averaged over MedQA and MMLU-Med under two judges (\%), with the rank among all ten methods.}
\label{tab:qwen_dur}
\begin{tabular}{lcccc}
\toprule
Method & GPT-4.1 & Rank & Qwen3.5-27B & Rank \\
\midrule
HYPER      & 77.9 & 1 & 75.6 & 1 \\
RRR        & 62.5 & 2 & 60.6 & 2 \\
LC-MQR     & 59.5 & 4 & 58.2 & 4 \\
Simple RAG & 39.3 & 7 & 39.6 & 7 \\
\bottomrule
\end{tabular}
\end{table}

\paragraph{Gold-blind judge.} Table~\ref{tab:gold_blind} compares the gold-aware and gold-blind protocols on the same Qwen3-4B MedQA contexts. Gold-blind DUR is higher for every method because a context can be useful for the option it supports even when that option is wrong. The ordering of methods is unchanged. When the gold-blind judge labels a HYPER context \textsc{Entailed}, the option it selects is correct in 98.8\% of cases, and for \textsc{Useful} contexts in 82.3\%. The corresponding values are 96.4\% and 87.2\% for Simple RAG and 98.1\% and 87.3\% for LC-MQR. For HYPER, $\kappa$ between the protocols is lower (.505) at a similar raw agreement, likely because most of its contexts fall into the useful categories under both protocols, which depresses $\kappa$.

\begin{table}[H]
\centering
\small
\setlength{\tabcolsep}{3.2pt}
\caption{Gold-aware and gold-blind DUR on Qwen3-4B MedQA (\%), with agreement between the protocols on the DUR label.}
\label{tab:gold_blind}
\begin{tabular}{lcccc}
\toprule
Method & Aware & Blind & Agr. & $\kappa$ \\
\midrule
Simple RAG & 30.0 & 39.9 & 87.9 & .737 \\
LC-MQR     & 62.2 & 73.8 & 85.3 & .668 \\
HYPER      & 83.2 & 91.1 & 88.8 & .505 \\
\bottomrule
\end{tabular}
\end{table}

\section{Controls and Robustness}
\label{app:robustness}

\paragraph{Option exposure and query roles.} Tables~\ref{tab:option_exposure_accuracy_full} and~\ref{tab:option_exposure_dur_full} give per-generator accuracy and DUR for the option-exposure control. Adding the options to LC-MQR raises DUR by about 8 points on average, but HYPER stays well above it on every generator. Tables~\ref{tab:ablation_queries_accuracy_full} and~\ref{tab:ablation_queries_dur_full} give the query-role ablation. Removing a role occasionally raises accuracy for one generator, for example \textsc{Distinction} for Qwen3-30B on MedQA, but it lowers DUR in every cell.

\begin{table*}[!t]
\centering
\fontsize{8}{8.5}\selectfont
\setlength{\tabcolsep}{3.0pt}
\caption{Full model-by-model accuracy values underlying the dataset-level averages reported in Table~\ref{tab:option-control}.}
\label{tab:option_exposure_accuracy_full}
\begin{tabular}{@{}lccccc ccccc@{}}
\toprule
& \multicolumn{5}{c}{MedQA} & \multicolumn{5}{c}{MMLU-Med} \\
\cmidrule(lr){2-6} \cmidrule(l){7-11}
Method
& {\fontsize{7.2}{7.8}\selectfont Llama3.2 3B}
& {\fontsize{7.2}{7.8}\selectfont Llama3.1 8B}
& {\fontsize{7.2}{7.8}\selectfont Qwen3 4B}
& {\fontsize{7.2}{7.8}\selectfont Qwen3 30B}
& Average
& {\fontsize{7.2}{7.8}\selectfont Llama3.2 3B}
& {\fontsize{7.2}{7.8}\selectfont Llama3.1 8B}
& {\fontsize{7.2}{7.8}\selectfont Qwen3 4B}
& {\fontsize{7.2}{7.8}\selectfont Qwen3 30B}
& Average \\
\midrule
LC-MQR            & 58.9 & 67.5 & 73.1 & 83.3 & 70.7 & 67.0 & 74.7 & 82.5 & 88.8 & 78.2 \\
LC-MQR $+$ options  & 57.0 & 68.3 & 75.8 & 84.1 & 71.3 & 67.5 & 77.2 & 83.7 & 88.4 & 79.2 \\
HYPER $-$ options       & 61.1 & 71.6 & 75.3 & 84.5 & 73.1 & \textbf{69.1} & 77.3 & 83.7 & 89.4 & 79.9 \\
\textbf{HYPER (ours)} & \textbf{63.4} & \textbf{73.0} & \textbf{77.9} & \textbf{85.8} & \textbf{75.0}
                     & 68.6 & \textbf{79.7} & \textbf{85.4} & \textbf{90.6} & \textbf{81.1} \\
\bottomrule
\end{tabular}
\end{table*}

\begin{table*}[!t]
\centering
\fontsize{8}{8.5}\selectfont
\setlength{\tabcolsep}{3.0pt}
\caption{Additional DUR results for the option-exposure control.  DUR denotes \textsc{Entailed} + \textsc{Useful}.} 
\label{tab:option_exposure_dur_full}
\begin{tabular}{@{}lccccc ccccc@{}}
\toprule
& \multicolumn{5}{c}{MedQA} & \multicolumn{5}{c}{MMLU-Med} \\
\cmidrule(lr){2-6} \cmidrule(l){7-11}
Method
& {\fontsize{7.2}{7.8}\selectfont Llama3.2 3B}
& {\fontsize{7.2}{7.8}\selectfont Llama3.1 8B}
& {\fontsize{7.2}{7.8}\selectfont Qwen3 4B}
& {\fontsize{7.2}{7.8}\selectfont Qwen3 30B}
& Average
& {\fontsize{7.2}{7.8}\selectfont Llama3.2 3B}
& {\fontsize{7.2}{7.8}\selectfont Llama3.1 8B}
& {\fontsize{7.2}{7.8}\selectfont Qwen3 4B}
& {\fontsize{7.2}{7.8}\selectfont Qwen3 30B}
& Average \\
\midrule
LC-MQR            & 45.2 & 51.2 & 62.2 & 65.6 & 56.1 & 51.4 & 53.7 & 56.7 & 59.1 & 55.3 \\
LC-MQR $+$ options  & 51.1 & 62.9 & 71.8 & 76.4 & 65.6 & 56.8 & 61.3 & 64.8 & 66.3 & 62.3 \\
HYPER $-$ options       & 61.0 & 70.5 & 70.5 & 80.5 & 70.6 & 57.0 & 61.1 & 64.2 & 67.5 & 62.4 \\
\textbf{HYPER (ours)} & \textbf{73.4} & \textbf{81.8} & \textbf{83.2} & \textbf{89.8} & \textbf{82.1} & \textbf{65.8} & \textbf{69.1} & \textbf{72.5} & \textbf{73.7} & \textbf{70.3} \\
\bottomrule
\end{tabular}
\end{table*}

\begin{table*}[!t]
\centering
\fontsize{8}{8.5}\selectfont
\setlength{\tabcolsep}{3.0pt}
\caption{Full model-by-model accuracy values underlying the dataset-level averages reported in Table~\ref{tab:query-ablation}. }
\label{tab:ablation_queries_accuracy_full}
\begin{tabular}{@{}lccccc ccccc@{}}
\toprule
& \multicolumn{5}{c}{MedQA} & \multicolumn{5}{c}{MMLU-Med} \\
\cmidrule(lr){2-6} \cmidrule(l){7-11}
Method
& {\fontsize{7.2}{7.8}\selectfont Llama3.2 3B}
& {\fontsize{7.2}{7.8}\selectfont Llama3.1 8B}
& {\fontsize{7.2}{7.8}\selectfont Qwen3 4B}
& {\fontsize{7.2}{7.8}\selectfont Qwen3 30B}
& Average
& {\fontsize{7.2}{7.8}\selectfont Llama3.2 3B}
& {\fontsize{7.2}{7.8}\selectfont Llama3.1 8B}
& {\fontsize{7.2}{7.8}\selectfont Qwen3 4B}
& {\fontsize{7.2}{7.8}\selectfont Qwen3 30B}
& Average \\
\midrule
\textbf{HYPER (ours)} & \textbf{63.4} & \textbf{73.0} & \textbf{77.9} & 85.8 & \textbf{75.0}& 68.6 & \textbf{79.7} & 85.4 & \textbf{90.6} & \textbf{81.1} \\
HYPER $-$ $q_{\mathrm{sup}}$ & 62.1 & 70.1 & 73.8 & 83.5 & 72.4 & 66.9 & 77.1 & 83.0 & 88.9 & 79.0\\
HYPER $-$ $q_{\mathrm{dist}}$ & 61.2 & 71.5 & 74.0 & \textbf{86.3} & 73.2 & \textbf{69.2} & 78.7 & 83.4 & 89.0 & 80.1 \\
HYPER $-$ $q_{\mathrm{feat}}$ & 61.3 & 70.9 & 74.3 & 85.6 & 73.0 & 68.2 & 77.8 & \textbf{86.0} & 90.1 & 80.5 \\
\bottomrule
\end{tabular}
\end{table*}

\begin{table*}[!t]
\centering
\fontsize{8}{8.5}\selectfont
\setlength{\tabcolsep}{3.0pt}
\caption{Additional DUR results for the HYPER query-role ablation. DUR denotes \textsc{Entailed} + \textsc{Useful}.}
\label{tab:ablation_queries_dur_full}
\begin{tabular}{@{}lccccc ccccc@{}}
\toprule
& \multicolumn{5}{c}{MedQA} & \multicolumn{5}{c}{MMLU-Med} \\
\cmidrule(lr){2-6} \cmidrule(l){7-11}
Method
& {\fontsize{7.2}{7.8}\selectfont Llama3.2 3B}
& {\fontsize{7.2}{7.8}\selectfont Llama3.1 8B}
& {\fontsize{7.2}{7.8}\selectfont Qwen3 4B}
& {\fontsize{7.2}{7.8}\selectfont Qwen3 30B}
& Average
& {\fontsize{7.2}{7.8}\selectfont Llama3.2 3B}
& {\fontsize{7.2}{7.8}\selectfont Llama3.1 8B}
& {\fontsize{7.2}{7.8}\selectfont Qwen3 4B}
& {\fontsize{7.2}{7.8}\selectfont Qwen3 30B}
& Average \\
\midrule
\textbf{HYPER (ours)} & \textbf{73.4} & \textbf{81.8} & \textbf{83.2} & \textbf{89.8} & \textbf{82.1} & \textbf{65.8} & \textbf{69.1} & \textbf{72.5} & \textbf{73.7} & \textbf{70.3} \\
HYPER $-$ $q_{\mathrm{sup}}$ & 55.0 & 62.1 & 66.5 & 70.9 & 63.6 & 48.5 & 52.3 & 55.2 & 55.8 & 53.0 \\
HYPER $-$ $q_{\mathrm{dist}}$ & 60.8 & 71.7 & 70.1 & 80.4 & 70.8 & 53.5 & 57.4 & 61.8 & 64.2 & 59.2 \\
HYPER $-$ $q_{\mathrm{feat}}$ & 63.1 & 73.0 & 74.2 & 82.6 & 73.2 & 51.6 & 57.0 & 60.8 & 63.0 & 58.1 \\
\bottomrule
\end{tabular}
\end{table*}

\paragraph{Final-context budget.} Table~\ref{tab:budget_sensitivity_qwen} varies the budget for Qwen3-4B from 3 to 30 documents. HYPER has the highest accuracy and DUR at every budget. Its accuracy peaks at 15 documents on both datasets, so its advantage holds at a matched context size.

\begin{table}[H]
\centering
\fontsize{8}{8.5}\selectfont
\setlength{\tabcolsep}{3.2pt}
\caption{Qwen3-4B budget sensitivity. Each method cell is reported as \emph{accuracy / DUR}.}
\label{tab:budget_sensitivity_qwen}
\begin{tabular}{llccc}
\toprule
Dataset & $B$ & Simple RAG & LC-MQR & HYPER \\
\midrule
MedQA & 3  & 70.3 / 13.1 & 71.4 / 31.2 & \textbf{73.1} / \textbf{62.8} \\
MedQA & 6  & 71.3 / 19.6 & 71.7 / 44.4 & \textbf{74.5} / \textbf{73.5} \\
MedQA & 9  & 71.5 / 22.7 & 73.1 / 50.6 & \textbf{75.6} / \textbf{77.8} \\
MedQA & 15 & 71.3 / 30.0 & 73.1 / 62.2 & \textbf{77.9} / \textbf{83.2} \\
MedQA & 30 & 72.3 / 36.9 & 73.3 / 68.9 & \textbf{77.4} / \textbf{86.9} \\
\midrule
MMLU-Med & 3  & 81.5 / 31.2 & 81.7 / 35.2 & \textbf{84.3} / \textbf{52.5} \\
MMLU-Med & 6  & 82.3 / 37.8 & 83.3 / 46.5 & \textbf{84.7} / \textbf{62.9} \\
MMLU-Med & 9  & 81.9 / 41.5 & 82.6 / 50.4 & \textbf{85.1} / \textbf{67.1} \\
MMLU-Med & 15 & 81.8 / 48.7 & 82.5 / 56.7 & \textbf{85.4} / \textbf{72.5} \\
MMLU-Med & 30 & 81.9 / 51.8 & 83.1 / 61.3 & \textbf{84.7} / \textbf{73.3} \\
\bottomrule
\end{tabular}
\end{table}

\paragraph{Fusion order.} Table~\ref{tab:fusion_ordering} compares score fusion with appending documents in query order and removing later duplicates. HYPER stays above LC-MQR under both rules, and score fusion is slightly better on average.

\begin{table*}[!t]
\centering
\fontsize{8}{8.5}\selectfont
\setlength{\tabcolsep}{2.3pt}
\caption{Fusion-order sensitivity. Values are accuracy (\%).}
\label{tab:fusion_ordering}
\begin{tabular}{@{}lcccccc ccccc@{}}
\toprule
& & \multicolumn{5}{c}{MedQA} & \multicolumn{5}{c}{MMLU-Med} \\
\cmidrule(lr){3-7} \cmidrule(l){8-12}
Method & Ordering
& {\fontsize{7.2}{7.8}\selectfont Llama3.2 3B}
& {\fontsize{7.2}{7.8}\selectfont Llama3.1 8B}
& {\fontsize{7.2}{7.8}\selectfont Qwen3 4B}
& {\fontsize{7.2}{7.8}\selectfont Qwen3 30B}
& Average
& {\fontsize{7.2}{7.8}\selectfont Llama3.2 3B}
& {\fontsize{7.2}{7.8}\selectfont Llama3.1 8B}
& {\fontsize{7.2}{7.8}\selectfont Qwen3 4B}
& {\fontsize{7.2}{7.8}\selectfont Qwen3 30B}
& Average \\
\midrule

LC-MQR          & score fusion & 58.9 & 67.5 & 73.1 & 83.3 & 70.7 & 67.0 & 74.7 & 82.5 & 88.8 & 78.2 \\
LC-MQR          & query order &58.2 &	67.9 &	73.1 &	84.1 &	70.8 & 68.4&	75.1&	81.3&	89.7&	78.6 \\
HYPER & score fusion & 63.4 & 73.0 & 77.9 & 85.8 & 75.0 & 68.6 & 79.7 & 85.4 & 90.6 & 81.1 \\
HYPER &query order  &62.5&73.1&76.1&85.3&74.3 &69.6&	78.6&	84.2&	89.7&	80.5 \\
\bottomrule
\end{tabular}
\end{table*}

\paragraph{Retrieval breadth.} Table~\ref{tab:dmqr25} expands DMQR to retrieve five documents for each of its five queries with a cap of 25 final documents. Because this needs a longer input, DMQR-25 uses a 16,384-token context, while HYPER keeps 15 documents and 8,192 tokens. The setting therefore favors DMQR, and HYPER is still ahead for every generator.

\begin{table}[!htbp]
\centering
\small
\setlength{\tabcolsep}{3.0pt}
\caption{DMQR with 25 final documents against HYPER with 15 (accuracy, \%).}
\label{tab:dmqr25}
\begin{tabular}{lcccc}
\toprule
& \multicolumn{2}{c}{MedQA} & \multicolumn{2}{c}{MMLU-Med} \\
\cmidrule(lr){2-3} \cmidrule(l){4-5}
Generator & DMQR-25 & HYPER & DMQR-25 & HYPER \\
\midrule
Llama3.2 3B & 56.6 & \textbf{63.4} & 62.5 & \textbf{68.6} \\
Llama3.1 8B & 69.1 & \textbf{73.0} & 71.7 & \textbf{79.7} \\
Qwen3 4B    & 74.6 & \textbf{77.9} & 83.3 & \textbf{85.4} \\
Qwen3 30B   & 84.3 & \textbf{85.8} & 90.1 & \textbf{90.6} \\
\midrule
Average     & 71.2 & \textbf{75.0} & 76.9 & \textbf{81.1} \\
\bottomrule
\end{tabular}
\end{table}

\paragraph{Option order.} Multiple-choice accuracy can depend on the order of the options \citep{zheng2024large}. Table~\ref{tab:option_order} rotates the options of all 1,273 MedQA questions cyclically for Qwen3-4B, applies the same order to the planner and the generator, and remaps the gold label. HYPER is best under every order, and its lowest accuracy is 2.5 points above the best baseline result under any order. Over the four runs, the provisional answer hypothesis falls on positions A, B, C, and D in 25.9\%, 24.8\%, 26.3\%, and 23.0\% of the questions with a valid hypothesis.

\begin{table}[H]
\centering
\fontsize{8}{8.5}\selectfont
\setlength{\tabcolsep}{2.8pt}
\caption{Qwen3-4B MedQA accuracy (\%) under cyclic option orders.}
\label{tab:option_order}
\begin{tabular}{lccccc}
\toprule
Method & ABCD & DABC & CDAB & BCDA & Mean $\pm$ SD \\
\midrule
CoT        & 72.8 & 72.3 & 72.6 & 73.1 & 72.7 $\pm$ 0.3 \\
Simple RAG & 71.3 & 70.0 & 70.4 & 70.3 & 70.5 $\pm$ 0.6 \\
LC-MQR     & 73.1 & 72.9 & 72.5 & 73.6 & 73.0 $\pm$ 0.5 \\
HYPER      & \textbf{77.9} & \textbf{76.1} & \textbf{77.0} & \textbf{76.5} & \textbf{76.9 $\pm$ 0.8} \\
\bottomrule
\end{tabular}
\end{table}

\paragraph{Closed-source model.} Table~\ref{tab:gpt41} uses GPT-4.1 as both the planner and the generator on MedQA with the same retrieval setup. HYPER is the most accurate method, but only 0.5 points above CoT, because GPT-4.1 already answers 93\% of the questions without retrieval. The DUR values are judged by GPT-4.1 itself, which may favor contexts planned by the same model. Simple RAG retrieves the same context as in the main experiments, so its DUR is unchanged.

\begin{table}[H]
\centering
\small
\caption{GPT-4.1 as planner and generator on MedQA (\%).}
\label{tab:gpt41}
\begin{tabular}{lcc}
\toprule
Method & Accuracy & DUR \\
\midrule
CoT        & 93.0 & -- \\
Simple RAG & 90.8 & 30.0 \\
LC-MQR     & 89.9 & 66.9 \\
HYPER      & \textbf{93.5} & \textbf{93.0} \\
\bottomrule
\end{tabular}
\end{table}

\paragraph{Corpus.} Table~\ref{tab:pubmed_corpus_analysis} repeats the Qwen3-4B comparison with PubMed as the corpus, keeping the retriever and budget fixed. HYPER again has the highest accuracy and DUR.

\begin{table}[H]
\centering
\small
\caption{Qwen3-4B PubMed corpus analysis.}
\label{tab:pubmed_corpus_analysis}
\begin{tabular}{llcc}
\toprule
Dataset & Method & Acc. & DUR \\
\midrule
MedQA & Simple RAG & 68.7 & 21.8 \\
MedQA & LC-MQR & 71.6 & 51.1 \\
MedQA & HYPER & \textbf{73.1} & \textbf{77.0} \\
\midrule
MMLU-Med & Simple RAG & 80.8 & 43.6 \\
MMLU-Med & LC-MQR & 81.5 & 52.6 \\
MMLU-Med & HYPER & \textbf{82.6} & \textbf{72.9} \\
\bottomrule
\end{tabular}
\end{table}

\paragraph{Planner and generator.} Table~\ref{tab:cross_model_planner} treats the hypothesis and query formulators as a planning module and pairs it with a different generator. A stronger planner (Qwen3-30B) raises the accuracy of both the Qwen3-4B and the Llama3.2-3B generators, so the planner and generator can be different models.

\begin{table}[H]
\centering
\small
\caption{Cross-model planner/generator analysis.}
\label{tab:cross_model_planner}
\begin{tabular}{lccc}
\toprule
Planner & Generator & MedQA & MMLU-Med \\
\midrule
\addlinespace[0.35em]
\multicolumn{4}{@{}l}{\textbf{Single model}} \\
\addlinespace[0.15em]
Llama3.2 3B & Llama3.2 3B & 63.4 & 68.6 \\
Qwen3 4B & Qwen3 4B & 77.9 & 85.4 \\
\midrule
\addlinespace[0.35em]
\multicolumn{4}{@{}l}{\textbf{Cross model}} \\
\addlinespace[0.15em]
Llama3.2 3B & Qwen3 4B & 74.0 & 84.2 \\
Qwen3 4B & Llama3.2 3B & 63.2 & 71.5 \\
\midrule
\addlinespace[0.35em]
\multicolumn{4}{@{}l}{\textbf{Strong planner}} \\
\addlinespace[0.15em]
Qwen3 30B & Llama3.2 3B & 68.7 & 72.8 \\
Qwen3 30B & Qwen3 4B & 81.9 & 86.4 \\
\bottomrule
\end{tabular}
\end{table}

\paragraph{Reasoning models.}
\label{app:reasoning_comparisons}
Table~\ref{tab:reasoning_comparison} compares HYPER with thinking-mode
and search-trained models, including Search-R1 \citep{jin2025search}.
These comparisons are diagnostic rather than directly matched baselines,
because they change model training or inference behavior, whereas HYPER
keeps the base generator fixed and changes the retrieval-planning
interface.

For these runs, we increase the vLLM context length and maximum output
length by a factor of three, to 24,576 and 6,144 tokens. We use
Qwen/Qwen3-4B-Thinking-2507 and Qwen/Qwen3-30B-A3B-Thinking-2507 for the
thinking-mode variants, and
PeterJinGo/SearchR1-nq\_hotpotqa\_train-llama3.2-3b-it-em-ppo for the
Search-R1 comparison. For Search-R1, we also increase the maximum number
of search turns from the default 4 to 12, since the default setting
often spent the available turns issuing queries before producing a final answer in our MedQA setting.

The thinking-mode results show that stronger test-time reasoning can
improve performance for sufficiently strong models, as in Qwen3-30B on
MedQA. However, this improvement is uneven: Qwen3-4B Thinking is
weaker than the corresponding non-thinking CoT and HYPER settings, and
Qwen3-30B HYPER remains comparable to Qwen3-30B Thinking overall. These
results suggest that longer reasoning is complementary to, rather than a
replacement for, retrieval-side evidence planning.

Search-R1 tests a different question: whether search-oriented RL
training transfers to this medical QA setting. Despite increasing the
turn budget, Search-R1 performs substantially worse than the
corresponding Llama3.2-3B CoT and HYPER settings. The evaluated Search-R1 model therefore does not show a transfer advantage in this setting.

\begin{table}[H]
\centering
\small
\caption{Reasoning-enhanced model comparisons.}
\label{tab:reasoning_comparison}
\begin{tabular}{lcc}
\toprule
Method & MedQA & MMLU-Med \\
\midrule
\addlinespace[0.35em]
\multicolumn{3}{@{}l}{\textbf{Llama3.2 model}} \\
\addlinespace[0.15em]
Llama3.2 3B CoT & 58.0 & 66.3 \\
Llama3.2 3B HYPER & 63.4 & 68.6 \\
Search-R1 3B PPO & 48.0 & 57.5 \\
\midrule
\addlinespace[0.35em]
\multicolumn{3}{@{}l}{\textbf{Qwen3 model}} \\
\addlinespace[0.15em]
Qwen3 4B CoT & 72.8 & 82.9 \\
Qwen3 4B Thinking CoT & 65.4 & 73.9 \\
Qwen3 4B HYPER & 77.9 & 85.4 \\
Qwen3 30B CoT & 84.1 & 90.2 \\
Qwen3 30B Thinking CoT & 86.7 & 90.4 \\
Qwen3 30B HYPER & 85.8 & 90.6 \\
\bottomrule
\end{tabular}
\end{table}

\section{Runtime and Accuracy}
\label{app:cost}
We measure end-to-end wall-clock time on all 1,273 MedQA questions with Qwen3-4B using vLLM 0.10.2. All methods run sequentially on the same node and NVIDIA B200 GPU, with access to all 72 physical CPU cores. Model files are cached locally, and compilation caches are reused where applicable. Each measurement spans a complete evaluation process, including model initialization, CPU retrieval, all generation stages, and result saving. Model downloads and environment setup are excluded.

LLM requests are batched by pipeline stage, with batched query encoding and FAISS search where supported. The implementations retain their own caching and model lifecycles: HYPER starts three model processes, while HyDE reads retrieved document files without using the text cache. Table~\ref{tab:runtime_measurement} reports the runtime and accuracy of one run per method. Accuracy is computed from the saved responses using the shared answer parser, with unparsable responses counted as incorrect. These timing runs are separate from the main accuracy experiments.

\begin{table}[H]
\centering
\small
\setlength{\tabcolsep}{4pt}
\caption{End-to-end runtime and accuracy from the same MedQA runs with Qwen3-4B.}
\label{tab:runtime_measurement}
\begin{tabular}{lrr}
\toprule
Method & Runtime (s) & Accuracy (\%) \\
\midrule
CoT & 99.8 & 72.6 \\
Simple RAG & 356.3 & 72.2 \\
MMR-RAG & 364.1 & 71.8 \\
Rerank-RAG & 1,567.8 & 73.0 \\
MAIN-RAG & 411.5 & 72.3 \\
ITEM-A & 356.2 & 69.7 \\
RRR & 387.2 & 73.6 \\
LC-MQR & 355.2 & 74.3 \\
DMQR & 366.6 & 74.5 \\
HyDE & 1,322.5 & 74.9 \\
HYPER & 529.3 & 76.7 \\
\bottomrule
\end{tabular}
\end{table}

\section{MMLU-Pro Details}
\label{app:mmlupro}
Table~\ref{tab:mmlupro-subject} gives accuracy for each of the 14 MMLU-Pro subjects. Among the four methods, HYPER is the best or tied for best in 34 of the 56 subject and generator cells (32 strictly). Its advantage is most consistent with Llama3.2-3B, where it leads in 12 of 14 subjects. With Qwen3-30B, CoT is best in 7 of 14 subjects. Unparsable responses are counted as wrong. Such responses are at most 1.2\% for every method and generator, and for Llama3.2-3B they are 0.0\% for CoT and 0.5\% for HYPER, so output-format differences account for little of the gain.

\begin{table*}[!t]
\centering
\fontsize{8.5}{9.2}\selectfont
\setlength{\tabcolsep}{2.6pt}
\caption{MMLU-Pro accuracy (\%) by subject. C: CoT, S: Simple RAG, L: LC-MQR, H: HYPER. Bold marks the best method for each generator and subject.}
\label{tab:mmlupro-subject}
\begin{tabular}{@{}lr cccc cccc cccc cccc@{}}
\toprule
& & \multicolumn{4}{c}{Llama3.2 3B} & \multicolumn{4}{c}{Llama3.1 8B} & \multicolumn{4}{c}{Qwen3 4B} & \multicolumn{4}{c}{Qwen3 30B} \\
\cmidrule(lr){3-6} \cmidrule(lr){7-10} \cmidrule(lr){11-14} \cmidrule(l){15-18}
Subject & $N$ & C & S & L & H & C & S & L & H & C & S & L & H & C & S & L & H \\
\midrule
Biology & 717 & 25.1 & 34.3 & 29.3 & \textbf{44.9} & \textbf{67.4} & 65.0 & 65.6 & 67.1 & 82.3 & 83.0 & 82.0 & \textbf{84.8} & 86.3 & 86.3 & 85.6 & \textbf{86.6} \\
Business & 789 & 29.4 & 31.9 & 22.7 & \textbf{38.4} & 47.8 & \textbf{51.5} & 51.2 & 50.8 & 69.5 & 70.3 & 69.3 & \textbf{71.1} & \textbf{77.6} & 76.6 & 76.4 & 76.4 \\
Chemistry & 1,132 & 25.2 & 23.1 & 19.1 & \textbf{29.0} & 38.1 & 40.2 & 40.8 & \textbf{41.2} & 71.8 & 72.0 & 72.5 & \textbf{73.9} & \textbf{78.1} & 76.5 & 77.4 & 75.6 \\
Computer science & 410 & 19.0 & 21.2 & 19.8 & \textbf{29.8} & 44.9 & \textbf{46.6} & 46.1 & 46.3 & 67.8 & \textbf{68.5} & 67.6 & \textbf{68.5} & \textbf{79.5} & 77.1 & 78.5 & 79.0 \\
Economics & 844 & 29.6 & 21.0 & 17.3 & \textbf{30.9} & 52.3 & 56.9 & 57.2 & \textbf{59.4} & 75.0 & 75.5 & 76.1 & \textbf{79.5} & \textbf{81.6} & 79.5 & 81.2 & 81.5 \\
Engineering & 969 & 17.4 & 18.2 & 16.2 & \textbf{22.3} & 27.2 & 32.2 & 29.8 & \textbf{32.3} & 42.6 & 44.3 & \textbf{45.1} & 43.7 & \textbf{52.4} & 52.3 & 50.5 & 51.9 \\
Health & 818 & 20.8 & 37.3 & 35.7 & \textbf{45.5} & 54.0 & 56.7 & 54.2 & \textbf{58.7} & 65.4 & 66.4 & 66.3 & \textbf{67.1} & 72.2 & 73.0 & \textbf{73.3} & 70.5 \\
History & 381 & 13.4 & 15.7 & 16.5 & \textbf{27.0} & 44.9 & 48.6 & 44.9 & \textbf{49.9} & 53.0 & \textbf{59.6} & 58.3 & 54.3 & \textbf{68.2} & 66.4 & 66.7 & 68.0 \\
Law & 1,101 & 9.7 & 10.6 & 10.9 & \textbf{18.4} & 30.4 & 26.6 & 29.9 & \textbf{30.8} & 35.2 & 36.2 & 37.1 & \textbf{39.5} & 49.4 & 49.9 & 50.2 & \textbf{51.6} \\
Math & 1,351 & \textbf{37.7} & 30.9 & 27.5 & 36.4 & 50.0 & 49.7 & \textbf{51.4} & 50.6 & 82.8 & \textbf{83.6} & 83.0 & 83.0 & 84.5 & \textbf{86.1} & 85.2 & 83.0 \\
Other & 924 & 25.9 & \textbf{41.0} & 36.9 & 40.9 & 44.9 & 51.6 & 50.4 & \textbf{52.8} & 59.0 & 64.6 & \textbf{65.0} & 64.5 & 71.0 & 69.0 & 71.2 & \textbf{71.9} \\
Philosophy & 499 & 13.6 & 30.9 & 29.3 & \textbf{35.1} & 43.5 & \textbf{45.7} & 43.5 & 44.9 & 57.5 & \textbf{62.3} & 61.7 & 61.5 & 67.5 & 66.1 & 68.9 & \textbf{69.1} \\
Physics & 1,299 & 23.3 & 28.3 & 21.7 & \textbf{29.5} & 38.8 & 41.9 & 42.5 & \textbf{44.5} & 71.8 & \textbf{72.4} & 72.0 & 71.7 & \textbf{79.1} & 77.8 & 78.8 & 77.4 \\
Psychology & 798 & 19.5 & 30.7 & 24.4 & \textbf{44.9} & 57.8 & 58.5 & 57.0 & \textbf{60.2} & 69.9 & \textbf{70.4} & 68.3 & \textbf{70.4} & 76.9 & 75.6 & 76.9 & \textbf{77.1} \\
\bottomrule
\end{tabular}
\end{table*}

\section{Provisional Answer Hypothesis and Examples}
\label{app:examples}

\subsection{Accuracy by Hypothesis Correctness}
Table~\ref{tab:wrong_by_run} breaks the wrong or unparsable hypothesis subset of Table~\ref{tab:hypothesis-split} down by generator and dataset. HYPER is ahead of Simple RAG in six of the eight settings. The two exceptions are Llama3.1-8B on MedQA ($-$7.8 points) and on MMLU-Med ($-$1.5 points).

\begin{table}[H]
\centering
\small
\setlength{\tabcolsep}{3.0pt}
\caption{Accuracy (\%) on questions whose provisional answer hypothesis is wrong or unparsable, by setting.}
\label{tab:wrong_by_run}
\begin{tabular}{llcccc}
\toprule
Generator & Dataset & $N$ & HYPER & Simple & $\Delta$ \\
\midrule
Llama3.2 3B & MedQA    & 525 & 36.0 & 35.8 & +0.2 \\
Llama3.2 3B & MMLU-Med & 351 & 39.3 & 36.5 & +2.8 \\
Llama3.1 8B & MedQA    & 435 & 37.9 & 45.7 & $-$7.8 \\
Llama3.1 8B & MMLU-Med & 270 & 45.6 & 47.0 & $-$1.5 \\
Qwen3 4B    & MedQA    & 407 & 41.5 & 36.4 & +5.2 \\
Qwen3 4B    & MMLU-Med & 203 & 42.4 & 36.9 & +5.4 \\
Qwen3 30B   & MedQA    & 237 & 40.5 & 37.6 & +3.0 \\
Qwen3 30B   & MMLU-Med & 132 & 40.2 & 39.4 & +0.8 \\
\bottomrule
\end{tabular}
\end{table}

\subsection{Correction and Confirmation}
\label{app:correction}
\begin{table}[H]
\centering
\small
\setlength{\tabcolsep}{4.0pt}
\caption{Correction, confirmation failure and damage rates (\%) of HYPER with each query removed, pooled over the four generators. Correction and confirmation are computed on wrong or unparsable hypotheses ($N$=2,560; MedQA 1,604; MMLU-Med 956), damage on correct hypotheses ($N$=6,888; MedQA 3,488; MMLU-Med 3,400).}
\label{tab:correction_confirmation}
\begin{tabular}{lccc}
\toprule
Condition & Correction $\uparrow$ & Confirm. $\downarrow$ & Damage $\downarrow$ \\
\midrule
\multicolumn{4}{l}{\textit{Pooled}} \\
HYPER                       & \textbf{39.8} & \textbf{41.3} & \textbf{8.1} \\
$-$ $q_{\mathrm{sup}}$      & 37.9 & \textbf{41.3} & 10.6 \\
$-$ $q_{\mathrm{dist}}$     & 36.6 & 44.4 & 8.8 \\
$-$ $q_{\mathrm{feat}}$     & 38.3 & 43.2 & 9.3 \\
\midrule
\multicolumn{4}{l}{\textit{MedQA}} \\
HYPER                       & \textbf{38.6} & 43.8 & \textbf{8.2} \\
$-$ $q_{\mathrm{sup}}$      & 38.1 & \textbf{42.6} & 11.8 \\
$-$ $q_{\mathrm{dist}}$     & 35.2 & 46.6 & 9.3 \\
$-$ $q_{\mathrm{feat}}$     & 35.7 & 46.1 & 9.8 \\
\midrule
\multicolumn{4}{l}{\textit{MMLU-Med}} \\
HYPER                       & 41.8 & \textbf{37.0} & \textbf{7.9} \\
$-$ $q_{\mathrm{sup}}$      & 37.7 & 39.0 & 9.4 \\
$-$ $q_{\mathrm{dist}}$     & 39.0 & 40.8 & 8.4 \\
$-$ $q_{\mathrm{feat}}$     & \textbf{42.7} & 38.3 & 8.9 \\
\bottomrule
\end{tabular}
\end{table}

Table~\ref{tab:correction_confirmation} describes how the final answer relates to the provisional hypothesis $\hat{y}_0$. The \emph{correction rate} is the share of wrong-hypothesis questions answered correctly, $P(\hat{y}=y^\ast \mid \hat{y}_0\neq y^\ast)$, and equals the wrong-hypothesis accuracy in Table~\ref{tab:hypothesis-split}. The \emph{confirmation failure rate} is the share of wrong-hypothesis questions whose final answer repeats the wrong hypothesis, $P(\hat{y}=\hat{y}_0 \mid \hat{y}_0\neq y^\ast)$. The \emph{damage rate} is the share of correct-hypothesis questions answered incorrectly, $P(\hat{y}\neq y^\ast \mid \hat{y}_0=y^\ast)$. Unparsable hypotheses count as wrong and are never counted as a confirmation match. Each ablation removes the documents retrieved by one query from the fused context and reruns the generator, so $\hat{y}_0$ and the partition into correct and wrong hypotheses are the same in every row.

With all three queries, the generator overturns 39.8\% of wrong hypotheses and keeps 41.3\%. Removing $q_{\mathrm{dist}}$ causes the largest loss in correction ($-$3.2 points) and the largest rise in confirmation failure ($+$3.2 points), consistent with its role of retrieving evidence that separates the hypothesis from its closest competitor. Removing $q_{\mathrm{sup}}$ leaves pooled confirmation failure unchanged at 41.3\% and causes the largest rise in damage to correct hypotheses (8.1\% to 10.6\%), so the support query mainly protects answers that are already right. Removing $q_{\mathrm{feat}}$ raises confirmation failure by 2.0 points. The pattern differs by dataset. On MedQA, removing either $q_{\mathrm{dist}}$ or $q_{\mathrm{feat}}$ raises confirmation failure by more than two points. On MMLU-Med, removing $q_{\mathrm{sup}}$ lowers correction the most, and removing $q_{\mathrm{feat}}$ leaves correction slightly higher (42.7\% against 41.8\%). Across all rows, four in ten wrong hypotheses survive to the final answer, the risk noted in the Limitations.

\subsection{Corrected Hypothesis: The Example in Figure~\ref{fig:pipeline}}
\label{app:example_corrected}
Figures~\ref{fig:ex757_state}--\ref{fig:ex757_output} give the full text of the example in Figure~\ref{fig:pipeline}(b), an MMLU-Med question answered by Qwen3-4B. The hypothesis formulator proposes $\beta$-thalassaemia, reasoning that $\beta$-thalassaemia major causes severe anaemia from birth. The three queries retrieve different evidence. The \textsc{Support} query returns a table of thalassaemia phenotypes and a passage stating that deleting all four $\alpha$-globin genes causes severe intrauterine anaemia. The \textsc{Distinction} query contrasts $\beta$- and $\alpha$-thalassaemia by onset and returns a passage stating that $\beta$-chain disorders become clinically apparent between 4 and 12 months of age. The \textsc{Feature} query returns passages on $\alpha$-thalassaemia genotypes, including the statement that the fetus is affected because fetal haemoglobin contains $\alpha$ chains. The generator, which sees only the question, the options, and the retrieved passages, uses the onset evidence to rule out $\beta$-thalassaemia and selects $\alpha$-thalassaemia, the correct answer. Simple RAG, which retrieves with the question alone, instead cites a passage stating that $\beta$-thalassemia major ``is readily made during childhood on the basis of severe anemia'' and answers $\beta$-thalassaemia. That passage is on topic, yet it is silent on whether the anaemia is present at birth.

\subsection{Corrected Hypothesis: A MedQA Example}
\label{app:example_medqa}
Figures~\ref{fig:HYPER_example_question}--\ref{fig:HYPER_example_generator_output} follow a MedQA question through HYPER with Qwen3-4B. The hypothesis formulator reads the deformity as clubfoot and proposes a foot abduction brace. All three queries are framed around clubfoot, but the \textsc{Distinction} query also names the other management options, and the \textsc{Support} and \textsc{Distinction} queries both retrieve a passage recommending reassurance and follow-up for in-toeing. The generator, which sees only the question, the options, and the retrieved passages, identifies metatarsus adductus and selects reassurance, the correct answer.

\subsection{Failed Hypothesis: A MedQA Example}
Figures~\ref{fig:HYPER_fail_question}--\ref{fig:HYPER_fail_output} show the opposite case for Qwen3-4B. A patient develops acute kidney injury with eosinophilia and livedo after cardiac catheterization, and the biopsy shows cholesterol clefts, described as spindle-shaped vacuoles. The planner reads these findings as eosinophilic vasculitis and proposes eosinophilic granulomatosis with polyangiitis (EGPA). All three queries inherit this reading. The \textsc{Distinction} query names cholesterol embolization but asks how it differs from EGPA in terms of eosinophilia and biopsy findings framed by the wrong interpretation, so retrieval returns vasculitis passages. The generator then rejects cholesterol embolization and selects EGPA. In this case the misreading of the key finding entered the retrieval-planning state itself, so all three queries carried it.

\begin{figure*}[!t]
\centering
\begin{tcolorbox}[colback=gray!5!white, colframe=gray!75!black, title={Example (MMLU-Med, Qwen3-4B): Question and Retrieval-Planning State}, fonttitle=\bfseries]
\small
\textbf{Question:} Severe anaemia at birth is a feature of what?

\textbf{Options:} A: Alpha-thalassaemia \quad B: Beta-thalassaemia \quad C: Hereditary persistence of fetal haemoglobin \quad D: Sickle cell disease \qquad \textbf{Gold answer:} A

\vspace{2mm}
\textbf{Provisional answer hypothesis $\hat{y}_0$:} B (Beta-thalassaemia)

\vspace{1mm}
\textbf{Discriminating features $f$:}
(1) Severe anaemia at birth is a hallmark of certain inherited haemoglobinopathies, particularly those involving defective globin chain synthesis.
(2) Alpha-thalassaemia typically presents with haemolytic anaemia in infancy or early childhood, but severe anaemia at birth is more characteristic of beta-thalassaemia major (coombs-negative, severe transfusion-dependent anaemia from birth).
(3) Hereditary persistence of fetal haemoglobin (HPFH) is generally asymptomatic or presents with mild features, not severe anaemia at birth.
(4) Sickle cell disease typically presents with pain crises, haemolytic anaemia, and complications later in life, not with severe anaemia at birth.

\vspace{1mm}
\textbf{Evidence needs $e$:}
(1) Beta-thalassaemia major presents with severe microcytic hypochromic anaemia at birth or within the first few months of life, requiring lifelong transfusions.
(2) Infants with beta-thalassaemia major often have hepatosplenomegaly, jaundice, and growth retardation due to ineffective erythropoiesis.
(3) In contrast, alpha-thalassaemia may present with mild to moderate anaemia, and severe cases are usually diagnosed later or in infancy with variable severity.

\vspace{1mm}
\textbf{Brief reasoning $r$:} Severe anaemia at birth is most classically associated with beta-thalassaemia major (also known as Cooley's anaemia), where there is a complete deficiency of beta-globin chains leading to ineffective erythropoiesis and severe anaemia from birth. In contrast, alpha-thalassaemia may cause anaemia, but it often presents later or with variable severity, and severe anaemia at birth is less typical. HPFH and sickle cell disease do not typically present with severe anaemia at birth. Therefore, beta-thalassaemia is the best match.
\end{tcolorbox}
\vspace{-2mm}
\caption{Question and retrieval-planning state for the example in Figure~\ref{fig:pipeline}(b). The provisional answer hypothesis is wrong.}
\label{fig:ex757_state}
\end{figure*}

\begin{figure*}[!t]
\centering
\begin{tcolorbox}[colback=gray!5!white, colframe=gray!75!black, title={Example: Queries and Retrieved Passages}, fonttitle=\bfseries]
\small
\textbf{\textsc{Support} $q_{\mathrm{sup}}$:} Beta-thalassaemia major presentation at birth: severe microcytic hypochromic anaemia, ineffective erythropoiesis, and transfusion dependence in infants

\textit{[3] Pediatrics\_Nelson\_3190:} ``Thalassemia intermedia Compound heterozygous $\beta^0$and $\beta^+$-thalassemia Moderate hemolysis, splenomegaly, moderately severe anemia, but not transfusion-dependent [\ldots]''

\textit{[5] Pediatrics\_Nelson\_3207:} ``Because alpha chains are needed for fetal erythropoiesis and production of hemoglobin F ($\alpha_2\gamma_2$), alpha chain hemoglobinopathies are present in utero. [\ldots] Deletion of all four genes leads to hydrops fetalis, severe intrauterine anemia, and death, unless intrauterine transfusions are administered.''

\vspace{2mm}
\textbf{\textsc{Distinction} $q_{\mathrm{dist}}$:} Distinguishing features between beta-thalassaemia major and alpha-thalassaemia in infants: onset of severe anaemia, haemoglobin patterns, and clinical presentation at birth

\textit{[12] InternalMed\_Harrison\_8039:} ``The $\alpha$-globin hemoglobinopathies are symptomatic in utero and after birth because normal function of the $\alpha$-globin gene is required throughout gestation and adult life. In contrast, infants with $\beta$-globin hemoglobinopathies tend to be asymptomatic until 3--9 months of age, when HbA has largely replaced HbF.''

\textit{[13] Pediatrics\_Nelson\_3209:} ``[\ldots] Disorders of the beta chain usually manifest themselves clinically between 4 and 12 months of age, unless they have been detected prenatally or by cord blood screening.''

\vspace{2mm}
\textbf{\textsc{Feature} $q_{\mathrm{feat}}$:} Key clinical features of beta-thalassaemia major in newborns: hepatosplenomegaly, jaundice, growth retardation, and Coombs-negative anaemia

\textit{[9] Obstentrics\_Williams\_7757:} ``[\ldots] he deletion of all four a-globin chain genes (--/--) characterizes homozygous a-thalassemia. Because a-chains are contained in fetal hemoglobin, the fetus is afected.''
\end{tcolorbox}
\vspace{-2mm}
\caption{Queries and selected retrieved passages for the example. Numbers in brackets are positions in the fused context. Each passage shown was retrieved only by the query it is listed under. Passages are reproduced verbatim, including errors in the source corpus, except for omissions marked [\ldots].}
\label{fig:ex757_queries}
\end{figure*}

\begin{figure*}[!t]
\centering
\begin{tcolorbox}[colback=gray!5!white, colframe=gray!75!black, title={Example: Generator Output}, fonttitle=\bfseries]
\small
Step 2: Review the documents for conditions presenting with severe anaemia at birth.

From Document [5]: `Deletion of all four genes leads to hydrops fetalis, severe intrauterine anemia, and death, unless intrauterine transfusions are administered.' This directly describes severe anaemia at birth (in utero).

Document [10]: `When none of the four genes are expressed, no a-globin chains are produced, and instead hemoglobin Bart ($\alpha$4) and hemoglobin H ($\gamma$4) are formed... The deletion of all four a-globin chain genes (--/--) characterizes homozygous a-thalassemia. Because a-chains are contained in fetal hemoglobin, the fetus is affected.' [\ldots]

Document [2] and [13]: Beta-thalassemia typically presents after birth, usually between 4--12 months of age, with severe anemia in $\beta$-thalassemia major (Cooley anemia), but not at birth. [\ldots]

Therefore, the only condition explicitly linked to severe anaemia at birth (in utero) is $\alpha$-thalassemia, specifically the homozygous form (deletion of all four $\alpha$-globin genes), which leads to hydrops fetalis.

\textbf{answer\_choice: A}
\end{tcolorbox}
\vspace{-2mm}
\caption{Excerpt of the generator output for the example. The generator answers from the retrieved evidence, with the provisional answer hypothesis withheld. Text is reproduced verbatim except for omissions marked [\ldots].}
\label{fig:ex757_output}
\end{figure*}

\begin{figure*}[!t]
\centering
\begin{tcolorbox}[colback=gray!5!white, colframe=gray!75!black, title={Example Question (MedQA)}, fonttitle=\bfseries]
\small

\textbf{Question:}

A 3-week-old male newborn is brought to the physician because of an inward turning of his left forefoot. 
He was born at 38 weeks' gestation by cesarean section because of breech presentation. 
The pregnancy was complicated by oligohydramnios. Examination shows concavity of the medial border 
of the left foot with a skin crease just below the ball of the great toe. The lateral border of the 
left foot is convex. The heel is in neutral position. Tickling the lateral border of the foot leads 
to correction of the deformity. The remainder of the examination shows no abnormalities. 
X-ray of the left foot shows an increased angle between the 1st and 2nd metatarsal bones.

Which of the following is the most appropriate next step in the management of this patient?

\textbf{Options}

A: Foot abduction brace \\
B: Arthrodesis of the forefoot \\
C: Reassurance \\
D: Tarsometatarsal capsulotomy

\end{tcolorbox}

\vspace{-2mm}
\caption{Representative MedQA example used to illustrate the end-to-end HYPER workflow.}
\label{fig:HYPER_example_question}
\end{figure*}

\begin{figure*}[!t]
\centering
\begin{tcolorbox}[colback=gray!5!white, colframe=gray!75!black, title={Hypothesis Formulation}, fonttitle=\bfseries]
\small

\textbf{Discriminating Features:}

The clinical presentation includes a medial concavity of the foot with a skin crease and correction 
with tickling of the lateral border—classic signs of \textit{talipes equinovarus} (congenital clubfoot) 
in the early stages.

\vspace{2mm}

\textbf{Confirming Evidence:}

The patient is 3 weeks old, which falls within the optimal age range for early intervention in 
congenital clubfoot.

\vspace{2mm}

\textbf{Reasoning:}

The clinical findings are consistent with congenital clubfoot. In infants younger than 6 months 
with a flexible deformity, the appropriate initial management is typically non-surgical correction 
using a foot abduction brace.

\vspace{2mm}

\textbf{Provisional Answer Hypothesis:}

Foot abduction brace

\end{tcolorbox}

\vspace{-2mm}
\caption{Output of the hypothesis formulator. The provisional answer hypothesis steers retrieval and can be confirmed or overturned by the retrieved evidence.}
\label{fig:HYPER_example_hypothesis}
\end{figure*}

\begin{figure*}[!t]
\centering
\begin{tcolorbox}[colback=gray!5!white, colframe=gray!75!black, title={Query Formulator Output}, fonttitle=\bfseries]
\small

\textbf{Support Query}

Foot abduction brace for congenital clubfoot in infants under 6 months with flexible deformity and 
response to tickling: evidence for first-line non-surgical management in early-stage talipes equinovarus

\vspace{3mm}

\textbf{Distinction Query}

Distinguishing features between foot abduction brace, arthrodesis, tarsometatarsal capsulotomy, and 
reassurance in the management of early congenital clubfoot in neonates under 3 months of age

\vspace{3mm}

\textbf{Feature Query}

Clinical and radiographic key features of congenital clubfoot in 3-week-old infants: medial concavity, 
lateral convexity, skin crease below great toe, correction with lateral tickling, and increased 
1st–2nd metatarsal angle on X-ray

\end{tcolorbox}

\vspace{-2mm}
\caption{Queries generated from the retrieval-planning state for the \textsc{Support}, \textsc{Distinction}, and \textsc{Feature} roles.}
\label{fig:HYPER_example_queries}
\end{figure*}

\begin{figure*}[!t]
\centering
\begin{tcolorbox}[colback=gray!5!white, colframe=gray!75!black, title={Retrieved Evidence Examples}, fonttitle=\bfseries]
\small

\textbf{Support / Distinction Evidence}

``Treatment of In-toeing. The mainstay of management is to identify patients who have pathologic 
reasons for in-toeing and provide reassurance and follow-up to document improvement. It can take 
until 7 to 8 years of age for correction. Fewer than 1\% of patients with in-toeing will require 
surgical intervention due to functional disability or cosmetic concerns.''

\vspace{3mm}

\textbf{Distinction / Key Feature Evidence}

``Family history is important. Clubfoot can be congenital, teratologic, or positional. Congenital 
clubfoot (approximately 75\% of cases) is typically isolated, although infants should be evaluated 
for developmental dysplasia of the hip. The infant will have hindfoot equinus and varus, forefoot 
adduction, and varying degrees of rigidity. These findings arise from abnormalities in the 
talonavicular joint.''

\end{tcolorbox}

\vspace{-2mm}
\caption{Representative passages retrieved by the generated queries. The retrieved set can contain both supporting and corrective evidence, which allows the generator to revise the provisional answer hypothesis.}
\label{fig:HYPER_example_evidence}
\end{figure*}

\begin{figure*}[!t]
\centering
\begin{tcolorbox}[colback=gray!5!white, colframe=gray!75!black, title={Generator Raw Output Examples}, fonttitle=\bfseries]
\small
The patient is a 3-week-old male with an inward turning of the left forefoot, described as concavity of the medial border of the foot with a skin crease just below the ball of the great toe and a convex lateral border. The heel is in neutral position, and tickling the lateral border leads to correction of the deformity. This indicates that the deformity is not fixed and can be passively corrected — a key feature of a positional deformity rather than a rigid congenital deformity like clubfoot.

The x-ray shows an increased angle between the 1st and 2nd metatarsal bones, which is consistent with metatarsus adductus — a condition where the forefoot is adducted (turned inward), often due to in utero positioning. This is a common, benign condition in newborns, especially in the context of a breech presentation and oligohydramnios, which may have contributed to abnormal positioning in utero.

Document [4] (Surgery\_Schwartz) states: 'Metatarsus adductus in infants will also resolve spontaneously in most cases.' Similarly, Document [8] notes that the calcaneovalgus foot (a different condition) resolves by 2 years with reassurance and passive stretching. The key point is that the deformity is not rigid and responds to passive manipulation (tickling corrects it), which supports a positional, non-pathologic origin.

Therefore, the most appropriate next step is reassurance, with monitoring and possibly passive stretching exercises, as the condition is expected to resolve spontaneously. No bracing, surgery, or arthrodesis is indicated at this stage.

Option A (foot abduction brace) is not indicated — bracing is not recommended for metatarsus adductus or intoeing in infants (Document [3]).
Option B (arthrodesis) is invasive and inappropriate for a benign, self-resolving condition.
Option D (tarsometatarsal capsulotomy) is a surgical procedure for a rigid deformity and not indicated in a condition that is passively correctable.
Thus, the best choice is C: Reassurance.\\
\textbf{answer\_choice: C}
\end{tcolorbox}
\caption{Raw generator output for the representative MedQA example. The generator answers from the retrieved evidence and the question, with the hypothesis withheld, and reaches the gold answer. The text is reproduced verbatim except for formatting.}
\label{fig:HYPER_example_generator_output}
\end{figure*}

\begin{figure*}[!t]
\centering
\begin{tcolorbox}[colback=gray!5!white, colframe=gray!75!black, title={Failure Example: Question (MedQA, Qwen3-4B)}, fonttitle=\bfseries]
\small
\textbf{Question:} Two weeks after undergoing an emergency cardiac catherization with stenting for unstable angina pectoris, a 61-year-old man has decreased urinary output and malaise. He has type 2 diabetes mellitus and osteoarthritis of the hips. Prior to admission, his medications were insulin and naproxen. He was also started on aspirin, clopidogrel, and metoprolol after the coronary intervention. His temperature is 38°C (100.4°F), pulse is 93/min, and blood pressure is 125/85 mm Hg. Examination shows mottled, reticulated purplish discoloration of the feet. Laboratory studies show: hemoglobin 14 g/dL; leukocyte count 16,400/mm\textsuperscript{3} (segmented neutrophils 56\%, eosinophils 11\%, lymphocytes 31\%, monocytes 2\%); platelet count 260,000/mm\textsuperscript{3}; erythrocyte sedimentation rate 68 mm/h; serum urea nitrogen 25 mg/dL; creatinine 4.2 mg/dL. Renal biopsy shows intravascular spindle-shaped vacuoles. Which of the following is the most likely cause of this patient's symptoms?

\vspace{2mm}
\textbf{Options:} A: Renal papillary necrosis \quad B: Cholesterol embolization \quad C: Eosinophilic granulomatosis with polyangiitis \quad D: Polyarteritis nodosa

\vspace{2mm}
\textbf{Gold answer:} B
\end{tcolorbox}
\vspace{-2mm}
\caption{Question of the failure example.}
\label{fig:HYPER_fail_question}
\end{figure*}

\begin{figure*}[!t]
\centering
\begin{tcolorbox}[colback=gray!5!white, colframe=gray!75!black, title={Failure Example: Retrieval-Planning State and Queries}, fonttitle=\bfseries]
\small
\textbf{Provisional answer hypothesis:} C. Eosinophilic granulomatosis with polyangiitis

\vspace{1mm}
\textbf{Reasoning (excerpt):} The key histological finding --- intravascular spindle-shaped vacuoles --- is most characteristic of eosinophilic vasculitis. The presence of eosinophilia (11\%), fever, malaise, and cutaneous purpura strongly point toward a condition involving eosinophils. [\ldots] In contrast, cholesterol embolization typically presents with acute renal failure and is associated with atherosclerotic disease, not eosinophilia.

\vspace{2mm}
\textbf{\textsc{Support} query:} What is the pathophysiological basis of intravascular spindle-shaped vacuoles in eosinophilic granulomatosis with polyangiitis and how do they correlate with clinical features like fever, eosinophilia, and cutaneous purpura?

\vspace{1mm}
\textbf{\textsc{Distinction} query:} How does eosinophilic granulomatosis with polyangiitis differ from polyarteritis nodosa, cholesterol embolization, and renal papillary necrosis in terms of eosinophilia, biopsy findings, and clinical presentation?

\vspace{1mm}
\textbf{\textsc{Feature} query:} What are the hallmark histological features of eosinophilic vasculitis, including intravascular spindle-shaped vacuoles, and how are they linked to systemic eosinophilia and cutaneous vasculitic manifestations?

\vspace{2mm}
\textbf{Top retrieved documents (excerpts):}
(1) Pathoma: ``Necrotizing vasculitis involving multiple organs, especially lung and kidney [\ldots]''
(2) Harrison: ``In eosinophilic granulomatosis with polyangiitis (EGPA), extravascular granulomas and necrotizing vasculitis may occur in the lungs, as well as in the heart, skin, muscle, liver, spleen, and kidneys [\ldots]''
(3) Harrison: ``The necrotizing vasculitis of eosinophilic granulomatosis with polyangiitis (Churg-Strauss) involves small and medium-sized muscular arteries [\ldots]''
\end{tcolorbox}
\vspace{-2mm}
\caption{Retrieval-planning state, queries, and retrieved documents of the failure example. All three queries carry the planner's reading of the biopsy finding.}
\label{fig:HYPER_fail_plan}
\end{figure*}

\begin{figure*}[!t]
\centering
\begin{tcolorbox}[colback=gray!5!white, colframe=gray!75!black, title={Failure Example: Generator Output (excerpt)}, fonttitle=\bfseries]
\small
B. Cholesterol embolization: This can cause acute kidney injury and may present with purpuric rash, but it is more commonly associated with atherosclerotic disease [\ldots] but not with significant eosinophilia or granulomatous features. The biopsy finding of intravascular spindle-shaped vacuoles is not characteristic of cholesterol embolization. [\ldots]

C. Eosinophilic granulomatosis with polyangiitis (Churg-Strauss syndrome): [\ldots] The patient has eosinophilia (11\%), which is notable. The biopsy shows vasculitis with intravascular changes. The presence of eosinophils and vasculitis in multiple organs (especially kidneys and skin) fits this diagnosis. [\ldots]

\textbf{answer\_choice: C}
\end{tcolorbox}
\vspace{-2mm}
\caption{Excerpt of the generator output for the failure example. The generator dismisses the correct answer using the retrieved vasculitis evidence. Text is reproduced verbatim except for omissions marked [\ldots].}
\label{fig:HYPER_fail_output}
\end{figure*}

\section{Prompt Templates}
\label{app:prompts}

This section reproduces the prompts used in the experiments. Figures~\ref{fig:prompt_cot} and~\ref{fig:prompt_generator_v1} show the answer prompts for CoT and for the retrieval-based methods. Figures~\ref{fig:prompt_rewriting}--\ref{fig:prompt_ITEM-A2} show the baseline prompts, Figure~\ref{fig:prompt_gpt_judge} the context-utility judge, and Figures~\ref{fig:prompt_HYPER_hypothesis}--\ref{fig:prompt_HYPER_formulator} the HYPER prompts, including the variant without options used in the option-exposure control. In the query formulator prompt, \{question\} contains the question text only.

\begin{figure*}[!t]
    \centering
    \begin{tcolorbox}[colback=gray!5!white, colframe=gray!75!black, title={GPT Judge Prompt}, fonttitle=\bfseries]
    \scriptsize
    \textbf{System:} \\
    You evaluate the relationship between retrieved context and the gold answer for a multiple-choice question.\\
    \\
    \textbf{User:} \\
    \#\#\# QUESTION\\
    \{question\}\\
    \\
    \#\#\# OPTIONS\\
    \{options\}\\
    \\
    \#\#\# GOLD ANSWER\\
    \{gold\_line\}\\
    \\
    \#\#\# RETRIEVED CONTEXT\\
    \{chunks\}\\
    \\
    \#\#\# TASK\\
    Given the retrieved context, evaluate its relationship to the gold answer.\\
    1. Identify the answer option most strongly supported by the retrieved context alone. Call this \'selected\_answer\'.\\
    2. Then classify how the retrieved context relates to the gold answer: whether it entails the gold answer, provides useful but incomplete support for it, or is not useful for supporting it.\\
    \\
    \#\#\# LABELS\\
    \\
    1. ENTAILED\\
    Definition:\\
    The gold answer is entailed by the retrieved context alone.\\
    The reasoning chain from the retrieved context to the gold answer is complete within the provided context.\\
    \\
    Required conditions:\\
    - The gold answer follows from the retrieved context alone.\\
    - No outside knowledge, missing fact, or unstated assumption is needed.\\
    - The conclusion is decisive, not merely plausible.\\
    - 'selected\_answer' matches the gold answer.\\
    \\
    2. USEFUL\_BUT\_NOT\_ENTAILED\\
    Definition:\\
    The retrieved context provides strong decision-relevant support for the gold answer, but the gold answer is not fully entailed by the retrieved context alone.\\
    A small additional reasoning step, minor outside knowledge, or a single missing bridge may still be needed, but the retrieved context already contains evidence that substantially narrows the decision toward the gold answer.\\
    \\
    This label should be used only when the retrieved context contains at least one of the following:\\
    - a decisive clue, rule, condition, or relation that strongly supports the gold answer, even if one final bridge is missing\\
    - evidence that rules out or strongly weakens one or more plausible alternative answers, leaving the gold answer as the clearly favored option\\
    - a partial but highly informative connection between the question and the gold answer that would make the correct answer reasonably reachable with only minimal additional knowledge\\
    \\
    Do not use this label for context that is merely relevant, broadly related, or weakly suggestive.\\
    \\
    3. NOT\_USEFUL\\
    Definition:\\
    The retrieved context does not provide strong decision-relevant support for the gold answer.\\
    It may be topically related, mention similar concepts, or provide background information, but it does not contain a decisive clue, a strong exclusion of alternatives, or a core connection that materially helps determine the gold answer.\\
    \\
    \#\#\# RULES\\
    - Evaluate only the retrieved context.\\
    - Evaluate the label relative to the gold answer.\\
    - Infer 'selected\_answer' from the retrieved context alone; do not force it to equal the gold answer.\\
    - Be conservative about ENTAILED.\\
    \\
    Output only JSON on the last line:\\
    \{\{"selected\_answer": "<A|B|C|D|UNSURE>", "label": "<ENTAILED|USEFUL\_BUT\_NOT\_ENTAILED|NOT\_USEFUL>"\}\}
    \end{tcolorbox}
    \vspace{-2mm}
    \caption{Prompt used for the context-utility judge.}
    \label{fig:prompt_gpt_judge}
\end{figure*}

\begin{figure*}[!t]
    \centering
    \begin{tcolorbox}[colback=gray!5!white, colframe=gray!75!black, title={COT Prompt}, fonttitle=\bfseries]
    \small
    \textbf{System:} \\
    You are a helpful medical expert, and your task is to answer a multi-choice medical question. Please first think step-by-step and then choose the answer from the provided options. Organize your output in a json formatted as Dict\{"step\_by\_step\_thinking": Str(explanation), "answer\_choice": Str\{A/B/C/...\}\}. Your responses will be used for research purposes only, so please have a definite answer. \\
    
    \textbf{User:} \\
    Here is the question: \\
    \{question\} \\
    
    Here are the potential choices: \\
    \{options\} \\
    
    Please think step-by-step and generate your output in json:
    \end{tcolorbox}
    \vspace{-2mm}
    \caption{Prompt used for the no-retrieval CoT baseline.}
    \label{fig:prompt_cot}

    \vspace{6mm}

    \begin{tcolorbox}[colback=gray!5!white, colframe=gray!75!black, title={LC-MQR Prompt}, fonttitle=\bfseries]
    \small
    \textbf{User:} \\
    You are an AI language model assistant. Your task is to generate exactly three different versions of the given user question to retrieve relevant documents from a vector database. By generating multiple perspectives on the user question, your goal is to help the user overcome some of the limitations of the distance-based similarity search.\\
    \\
    Original question: \{query\}\\
    \\
    Format your response in plain text as:\\
    \\
    Sub-query 1:\\
    \\
    Sub-query 2:\\
    \\
    Sub-query 3:
    \end{tcolorbox}
    \vspace{-2mm}
    \caption{The prompt used for LC-MQR baseline.}
    \label{fig:prompt_rewriting}

    \vspace{6mm}

    \begin{tcolorbox}[colback=gray!5!white, colframe=gray!75!black, title={LC-MQR with Options Prompt}, fonttitle=\bfseries]
    \small
    \textbf{User:} \\
    You are an AI language model assistant. Your task is to generate exactly three different versions of the given user question to retrieve relevant documents from a vector database. By generating multiple perspectives on the user question, your goal is to help the user overcome some of the limitations of the distance-based similarity search.\\
    \\
    Original question: \{query\}\\
    \\
    Options:\\
    \{options\}\\
    \\
    Format your response in plain text as:\\
    \\
    Sub-query 1:\\
    \\
    Sub-query 2:\\
    \\
    Sub-query 3:
    \end{tcolorbox}
    \vspace{-2mm}
    \caption{The prompt used for LC-MQR with options ablation.}
    \label{fig:prompt_rewriting_options}
\end{figure*}

\begin{figure*}[!t]
    \centering
    \begin{tcolorbox}[colback=gray!5!white, colframe=gray!75!black, title={HyDE Prompt}, fonttitle=\bfseries]
    \small
    \textbf{User:} \\
    Please write a passage to answer the question. \\
    Question: \{question\} \\
    Passage:
    \end{tcolorbox}
    \vspace{-2mm}
    \caption{Prompt used for generating hypothetical documents in the HyDE baseline.}
    \label{fig:prompt_hyde}

    \vspace{6mm} 

    \begin{tcolorbox}[colback=gray!5!white, colframe=gray!75!black, title={MAIN-RAG: Agent-1 (Predictor) Prompt}, fonttitle=\bfseries]
    \small
    \textbf{System:} \\
    You are an accurate and reliable AI assistant that can answer questions with the help of external documents. You should only provide the correct answer without repeating the question and instruction. \\
    
    \textbf{User:} \\
    Document: \\
    \{context\} \\
    
    Question: \\
    \{question\} \\
    
    Options: \\
    \{options\} \\
    
    Based ONLY on the provided document, what is the best answer to the question? Output only the answer choice.
    \end{tcolorbox}
    \vspace{-2mm}
    \caption{The prompt used for Agent-1 (Predictor) in the MAIN-RAG baseline to generate document-conditioned answers for the Document-Query-Answer triplet.}
    \label{fig:prompt_mainrag_agent1}

    \vspace{6mm} 

    \begin{tcolorbox}[colback=gray!5!white, colframe=gray!75!black, title={MAIN-RAG: Agent-2 (Judge) Prompt}, fonttitle=\bfseries]
    \small
    \textbf{System:} \\
    You are a noisy document evaluator that can judge if the external document is noisy for the query with unrelated or misleading information. Given a retrieved Document, a Question, and an Answer generated by an LLM (LLM Answer), you should judge whether both the following two conditions are reached: (1) the Document provides specific information for answering the Question; (2) the LLM Answer directly answers the question based on the retrieved Document. Please note that external documents may contain noisy or factually incorrect information. If the information in the document does not contain the answer, you should point it out with evidence. You should answer with 'Yes' or 'No' with evidence of your judgment, where 'No' means one of the conditions (1) and (2) are unreached and indicates it is a noisy document. \\
    
    \textbf{User:} \\
    Document: \\
    \{context\} \\
    
    Question with Options: \\
    \{question\} \\
    \{options\} \\
    
    LLM Answer: \\
    \{llm\_answer\} \\
    
    Does the document contain enough specific information to answer the question, and is the LLM Answer directly supported by the document? Answer with exactly one token: Yes or No.
    \end{tcolorbox}
    \vspace{-2mm}
    \caption{The prompt used for Agent-2 (Judge) in the MAIN-RAG baseline to evaluate the decision-usefulness and reliability of the retrieved document.}
    \label{fig:prompt_mainrag_agent2}
\end{figure*}

\begin{figure*}[!t]
    \centering
    \begin{tcolorbox}[colback=gray!5!white, colframe=gray!75!black, title={RRR Prompt}, fonttitle=\bfseries]
    \small
    \textbf{User:} \\
    Provide a better search query for web search engine to answer the given question, end the query with '**'. Question: \{question\} Options:\{options\}  \\
    \\
    Query:
    \end{tcolorbox}
    \vspace{-2mm}
    \caption{The prompt used for RRR baseline.}
    \label{fig:prompt_rrr}

    \vspace{6mm} 

    \begin{tcolorbox}[colback=gray!5!white, colframe=gray!75!black, title={DMQR Prompt}, fonttitle=\bfseries]
    \small
    
    \textbf{User:} \\
\#\#\# Instruction \#\#\# \\
You will receive a user’s question that requires retrieving relevant content through internet search \\
to provide an answer. There are now the following four rewriting methods, General Search \\
Rewriting, Keyword Rewriting, Pseudo-Answer Rewriting, Core Content Extraction. \\
Please apply four rewriting methods to rewrite the question. \\
\#\#\# General Search Rewriting \#\#\# \\
Rewrite the question into a general query for internet search. \\
\#\#\# Keyword Rewriting \#\#\# \\
Extract all keywords from the question and separate them with commas, preserving the amount \\
of information as in the original question. \\
\#\#\# Pseudo-Answer Rewriting \#\#\# \\
Generate an answer for the question, and use the answer to match the real answers from the search engine. \\
\#\#\# Core Content Extraction \#\#\# \\
Reduce the amount of information in the original question, only extracting the most core content. \\
The rewritten query should be more brief than Keyword Rewriting. \\
\#\#\# Example \#\#\# \\
Question: Which city was the site where the armistice agreement officially ending World War I was signed? \\
Output: \\
General Search Rewriting: City where World War I armistice agreement was signed \\
Keyword Rewriting: World War I, Armistice, Signing Location \\
Pseudo-Answer Rewriting: The armistice that ended World War I was signed in the city of Compiegne. \\
Core Content Extraction: World War I armistice signing city  \\
Begin! Only output the final result without any additional content. Do not generate any other \\
unrelated content. \\
Question: \{query\} \\
Output: 
    \end{tcolorbox}
    \vspace{-2mm}
    \caption{The prompt used for DMQR baseline.}
    \label{fig:prompt_dmqr}

\end{figure*}

\begin{figure*}[!t]
    \centering
    \begin{tcolorbox}[colback=gray!5!white, colframe=gray!75!black, title={ITEM-A: Generating pseudo answer prompt}, fonttitle=\bfseries]
    \small
    \textbf{User:} \\
You are a faithful question and answer assistant. Answer the question based on the given information with one or few sentences without the source. \\
    \textbf{assistant:} \\
Yes, i am the faithful question and answer assistant. \\
    \textbf{User:} \\
Given the information: \\
\{Documents\} \\
Answer the following question based on the given information with one or few sentences without the source.\\
Question: \{question\} \\
\\
Answer:

    \end{tcolorbox}
    \vspace{-2mm}
    \caption{The prompt used for generating pseudo answer in ITEM-A baseline.}
    \label{fig:prompt_ITEM-A1}

    \vspace{6mm} 

    \begin{tcolorbox}[colback=gray!5!white, colframe=gray!75!black, title={ITEM-A: Utility selection prompt}, fonttitle=\bfseries]
    \small
    
    \textbf{User:} \\
You are the utility judger, an intelligent assistant that can select the passages that have utility in answering the question.
    \textbf{assistant:} \\
Yes, i am the utility judger. \\
    \textbf{User:} \\
    I will provide you with \{num\} passages, each indicated by number identifier []. \\
    I will also provide you with a reference answer to the question. \\
    Select the passages that have utility in generating the reference answer to the following question from the \{num\} passages: \{query\}. \\
    \textbf{assistant:} \\
Okay, please provide the passages and the reference answer.
    \textbf{User:} \\
    {}[1] \{passage\_1\} \\
    \textbf{assistant:} \\
    Received passage [1]. \\
    \textbf{User:} \\
    {}[2] \{passage\_2\} \\
    \textbf{assistant:} \\
    Received passage [2]. \\
 ...... \\
     \textbf{User:} \\
Question: \{query\}. \\ 
Reference answer: \{pseudo answer\}. \\
\\
The requirements for judging whether a passage has utility in answering the question are: The passage has utility in answering the question, meaning that the passage not only be relevant to the question, but also be useful in generating a correct, reasonable and perfect answer to the question. \\
Directly output the passages you selected that have utility in generating the reference answer to the question. The format of the output is: 'My selection:[[i],[j],...].'. Only response the selection results, do not say any word or explain.

    \end{tcolorbox}
    \vspace{-2mm}
    \caption{The prompt used for utility selection in ITEM-A baseline.}
    \label{fig:prompt_ITEM-A2}

\end{figure*}

\begin{figure*}[!t]
    \centering
    \begin{tcolorbox}[colback=gray!5!white, colframe=gray!75!black, title={HYPER: Hypothesis Formulator Prompt}, fonttitle=\bfseries]
    \small
    \textbf{System:} \\
    You are an expert analyst taking an exam. \\
    
    \textbf{User:} \\
    Question: \{question\} \\
    
    Options: \\
    \{options\} \\
    
    Analyze this question carefully. Think step-by-step about each option, considering the information given in the question and relevant knowledge. Reason through the elimination analysis before making your final assessment. \\
    
    After your analysis, provide your final assessment in JSON: \\
    \{ \\
    \hspace*{4mm} "discriminating\_features": ["2-3 features that distinguish between options"], \\
    \hspace*{4mm} "reasoning": "brief explanation why this is the best answer", \\
    \hspace*{4mm} "confirming\_evidence": ["1-3 specific facts that would confirm this answer"], \\
    \hspace*{4mm} "best\_guess": "A/B/C/D", \\
    \hspace*{4mm} "best\_guess\_text": "<<copy the chosen option text verbatim>>" \\
    \}
    \end{tcolorbox}
    \vspace{-2mm}
    \caption{Prompt used for the HYPER hypothesis-formulation stage.}
    \label{fig:prompt_HYPER_hypothesis}

    \vspace{6mm}
    \begin{tcolorbox}[colback=gray!5!white, colframe=gray!75!black, title={HYPER: Hypothesis Formulator without Options Prompt}, fonttitle=\bfseries]
    \small
    \textbf{System:} \\
    You are an expert analyst taking an exam. \\
    
    \textbf{User:} \\
    Question: \{question\} \\
    \\
    Analyze this question carefully. Think step-by-step about each option, considering the information given in the question and relevant knowledge. Reason through the elimination analysis before making your final assessment. \\
    
    After your analysis, provide your final assessment in JSON: \\
    \{ \\
    \hspace*{4mm} "discriminating\_features": ["2-3 features that distinguish between options"], \\
    \hspace*{4mm} "reasoning": "brief explanation why this is the best answer", \\
    \hspace*{4mm} "confirming\_evidence": ["1-3 specific facts that would confirm this answer"], \\
    \hspace*{4mm} "best\_guess": "write your best guess for the answer" \\
    \}
    \end{tcolorbox}
    \vspace{-2mm}
    \caption{The prompt used for the Hypothesis Formulator (Stage 1) without options for ablation.}
    \label{fig:prompt_HYPER_hypo_no_options}
\end{figure*}

\begin{figure*}[!t]
    \centering
    \begin{tcolorbox}[colback=gray!5!white, colframe=gray!75!black, title={HYPER: Query Formulator Prompt}, fonttitle=\bfseries]
    \small
    \textbf{System:} \\
    You are a search query expert. Generate precise, targeted search queries. Output ONLY the 3 queries in the exact format requested. \\
    
    \textbf{User:} \\
    Generate 3 highly targeted search queries to find evidence for this question. \\
    
    Question: \{question\} \\
    Best Guess Answer: \{best\_guess\_text\} \\
    Reasoning: \{reasoning\} \\
    Evidence Needed: \{confirming\_evidence\} \\
    Key Features: \{discriminating\_features\} \\
    
    Generate 3 SPECIFIC queries: \\
    Query 1: Find evidence supporting \{best\_guess\_text\} - focus on the main reasoning \\
    Query 2: Find distinguishing criteria between the top candidate answers \\
    Query 3: Find specific key features or facts \\
    
    Format: \\
    Query 1: [query] \\
    Query 2: [query] \\
    Query 3: [query]
    \end{tcolorbox}
    \vspace{-2mm}
    \caption{Prompt used for the HYPER query-formulation stage.}
    \label{fig:prompt_HYPER_formulator}

    \vspace{6mm}

    \begin{tcolorbox}[colback=gray!5!white, colframe=gray!75!black, title={Final Generator Prompt}, fonttitle=\bfseries]
    \small
    \textbf{System:} \\
    You are a helpful medical expert, and your task is to answer a multi-choice medical question using the relevant documents. Please first think step-by-step and then choose the answer from the provided options. Organize your output in a json formatted as Dict\{"step\_by\_step\_thinking": Str(explanation), "answer\_choice": Str\{A/B/C/...\}\}. Your responses will be used for research purposes only, so please have a definite answer. \\
    
    \textbf{User:} \\
    Here are the relevant documents: \\
    \{context\} \\
    
    Here is the question: \\
    \{question\} \\
    
    Here are the potential choices: \\
    \{options\} \\
    
    Please think step-by-step and generate your output in json:
    \end{tcolorbox}
    \vspace{-2mm}
    \caption{The MIRAGE general prompt used by all retrieval-based methods on MedQA and MMLU-Med.}
    \label{fig:prompt_generator_v1}
\end{figure*}

\mbox{}
\clearpage

\end{document}